\pdfoutput=1

\documentclass[11pt]{article}

\usepackage[final]{acl}

\usepackage{times}
\usepackage{latexsym}
\usepackage{booktabs} 
\usepackage{algorithm}
\usepackage{algpseudocode} 
\usepackage[most]{tcolorbox}
\usepackage[dvipsnames]{xcolor}
\usepackage[table]{xcolor}
\usepackage{makecell}
\usepackage{pdfpages}
\usepackage{needspace}
\usepackage{tikz}
\usepackage{xurl}   
\usepackage{fontawesome} 
\usepackage{placeins}
\usepackage{needspace}
\usepackage{enumitem}

\usepackage[T1]{fontenc}

\usepackage[utf8]{inputenc}

\usepackage{microtype}

\usepackage{inconsolata}

\usepackage{graphicx}
\usepackage{listings}
\usepackage{multirow}
\usepackage{subcaption}
\usepackage{courier}
\usepackage{cuted}
\usepackage{array} 
\usepackage{csquotes}
\usepackage{url}
\usepackage{tabularx}
\usepackage{array}

\newcolumntype{L}{>{\raggedright\arraybackslash}X}

\UseRawInputEncoding

\title{\textsc{DEtect}:\\Determining Ease and Textual Clarity of German Text Simplifications}

\newcommand{\uzh}{\includegraphics[width=1.5em]{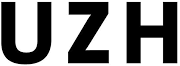}}

\newcommand{\huggingface}{
  \raisebox{-0.5ex}{\includegraphics[height=1.3em]{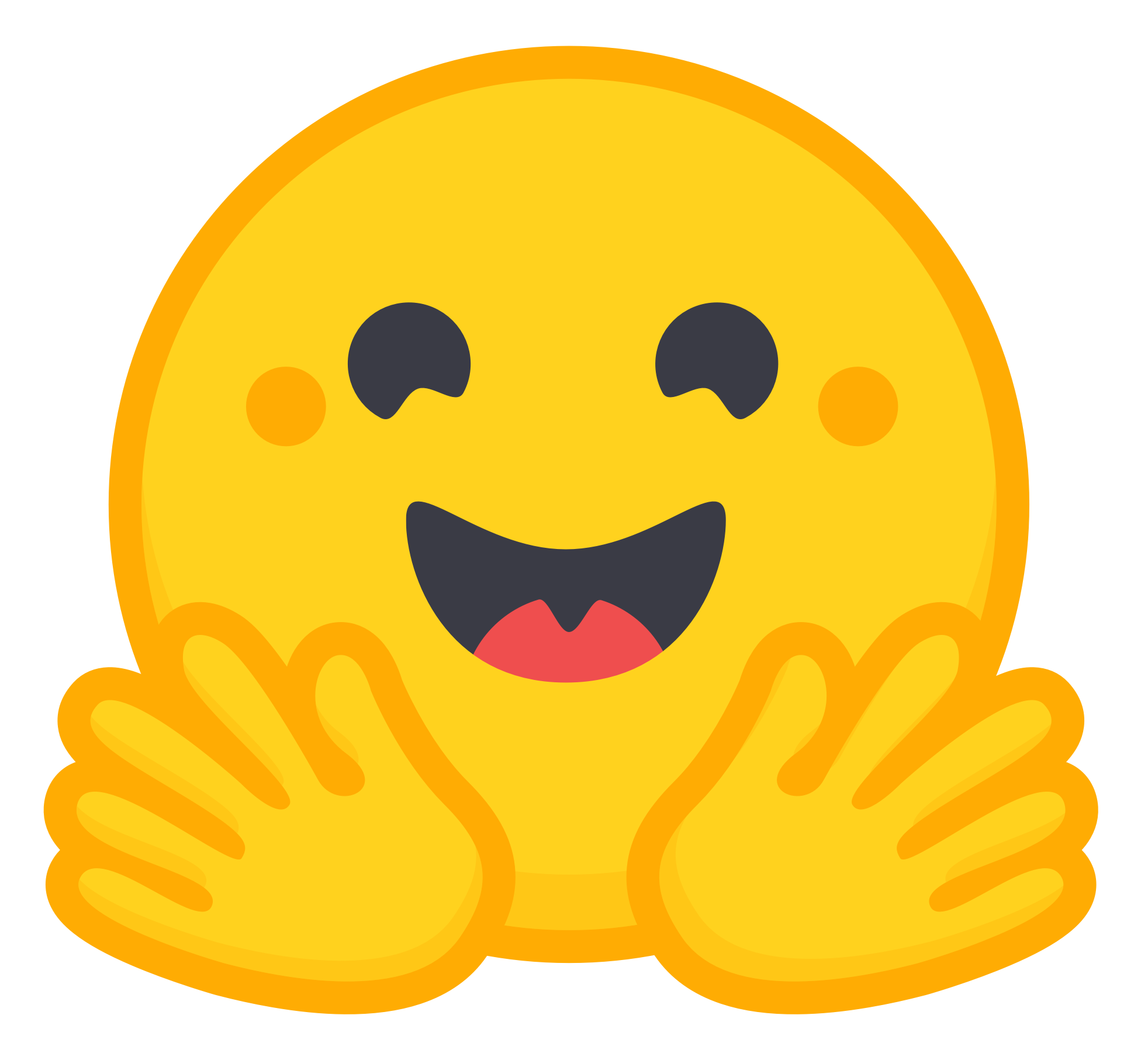}}
}

\newcommand{\human}{
  \raisebox{-0.5ex}{\includegraphics[height=1.5em]{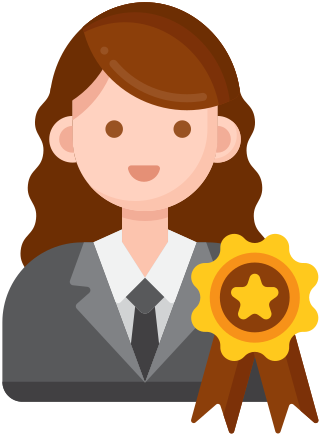}}
}

\newcommand{\robot}{
  \raisebox{-0.6ex}{\includegraphics[height=1.3em]{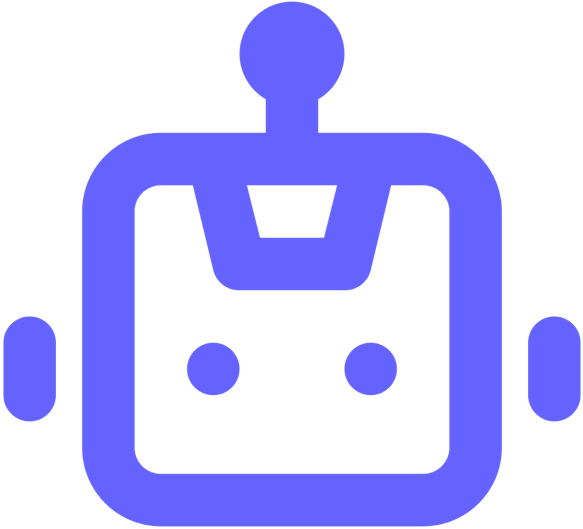}}
}

\definecolor{iconlight}{gray}{0.75} 

\author{
\hspace{0.5em}
  \textbf{Maria Korobeynikova}$^{\uzh}$ \hspace{0.2em}
  \textbf{Alessia Battisti}$^{\uzh}$ \hspace{0.2em}
  \textbf{Lukas Fischer}$^{\uzh}$
  \hspace{0.2em}
  \textbf{Yingqiang Gao\footnotemark[2]}$^{\uzh}$
  \\ \\
  \hspace{1em}
  $^{\uzh}$Department of Computational Linguistics, University of Zurich, Switzerland \\ \\
  \hspace{0.6em} 
  \texttt{\{maria.korobeynikova, yingqiang.gao\}@uzh.ch} \\
}

\definecolor{StepBlue}{HTML}{389BFF}

\begin{document}
\maketitle

\renewcommand{\thefootnote}{\fnsymbol{footnote}}

\footnotetext[2]{Corresponding author.}

\begin{abstract}
Current evaluation of German automatic text simplification (ATS) relies on general-purpose metrics such as SARI, BLEU, and BERTScore, which insufficiently capture simplification quality in terms of \textit{simplicity}, \textit{meaning preservation}, and \textit{fluency}. While specialized metrics like LENS have been developed for English, corresponding efforts for German have lagged behind due to the absence of human-annotated corpora. To close this gap, we introduce \textsc{DEtect}, the first German-specific metric that holistically evaluates ATS quality across all three dimensions of \textit{simplicity}, \textit{meaning preservation}, and \textit{fluency}, and is trained entirely on synthetic large language model (LLM) responses. Our approach adapts the LENS framework to German and extends it with (i) a pipeline for generating synthetic quality scores via LLMs, enabling dataset creation without human annotation, and (ii) an LLM-based refinement step for aligning grading criteria with simplification requirements. To the best of our knowledge, we also construct the largest German human evaluation dataset for text simplification to validate our metric directly. Experimental results show that \textsc{DEtect} achieves substantially higher correlations with human judgments than widely used ATS metrics, with particularly strong gains in meaning preservation and fluency. Beyond ATS, our findings highlight both the potential and the limitations of LLMs for automatic evaluation and provide transferable guidelines for general language accessibility tasks.

\noindent
\begin{minipage}{\columnwidth}
\centering
\raisebox{-0.1cm}{\hspace{-3em}\includegraphics[height=0.5cm]{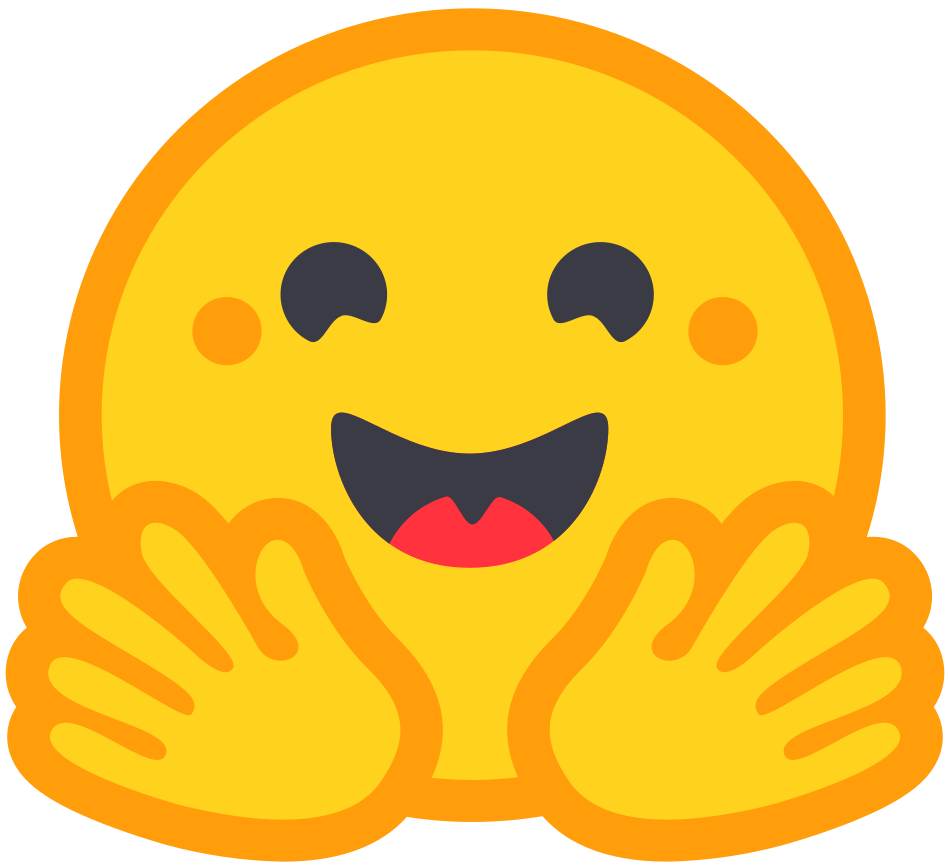}}~
\href{https://huggingface.co/datasets/ZurichNLP/SimpEvalDE}{\textcolor{gray}{Dataset}} \quad
\raisebox{-0.15cm}{\includegraphics[height=0.5cm]{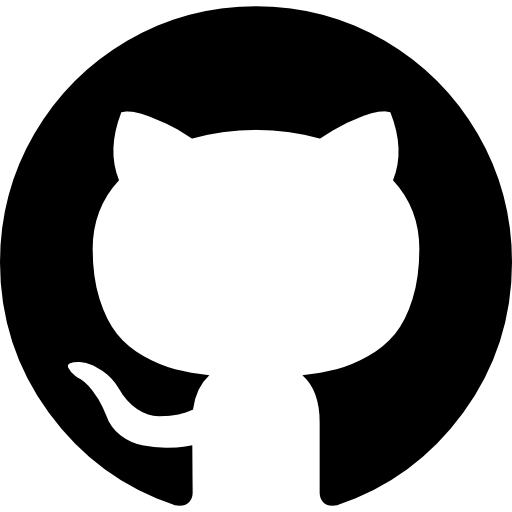}}~
\href{https://github.com/ZurichNLP/DETECT}{\textcolor{gray}{Code}}
\end{minipage}

\end{abstract}

\section{Introduction}

\begin{figure}[htb]
    \centering
    \includegraphics[width=\columnwidth]{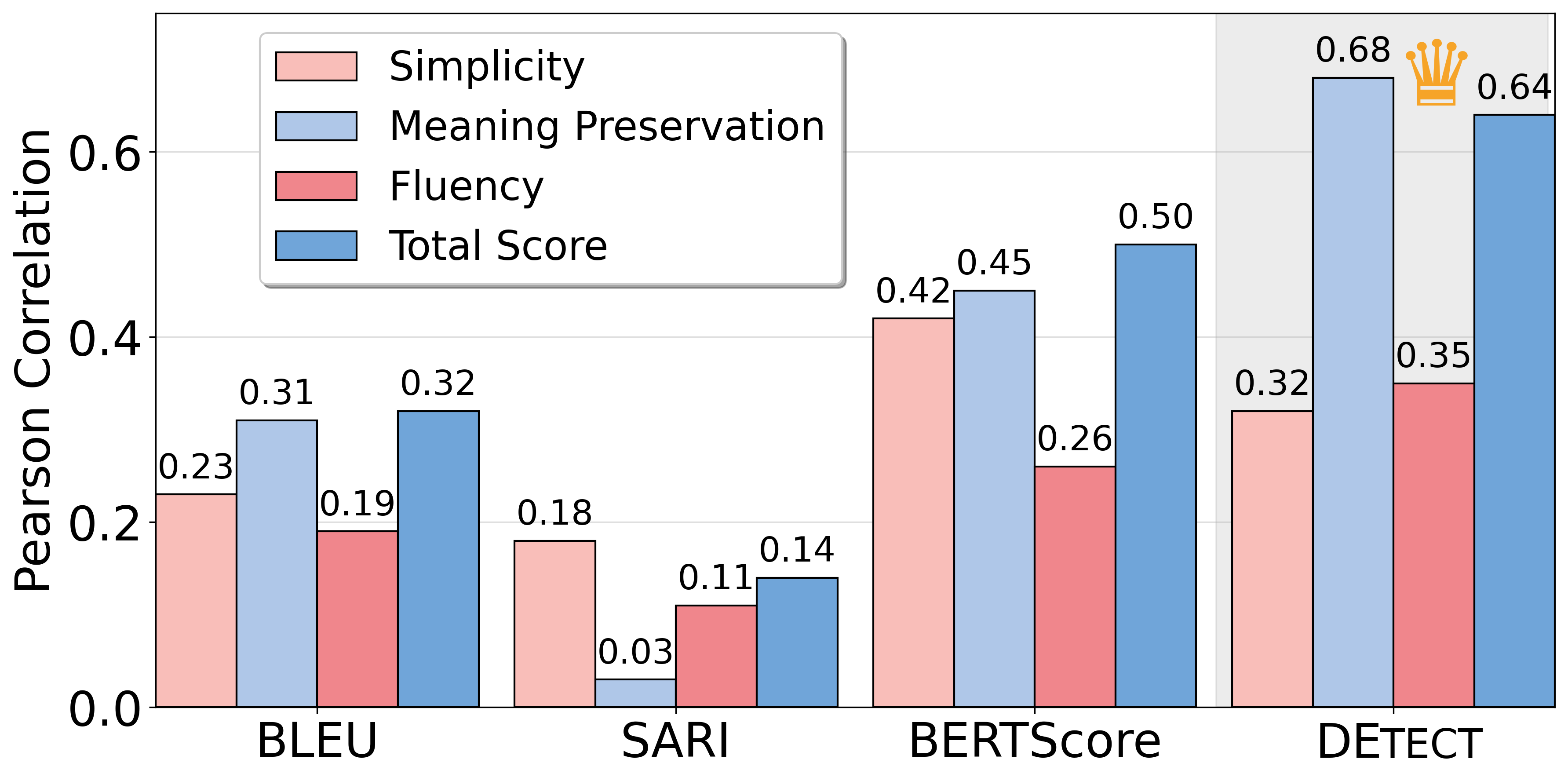}
    \caption{Correlation of standard ATS metrics and \textsc{DEtect} with human judgments across \textit{simplicity}, \textit{meaning preservation}, \textit{fluency}, and the total score. Standard ATS metrics (BLEU, SARI, BERTScore) show limited alignment, while \textsc{DEtect} achieves higher correlations, particularly for \textit{meaning preservation} and overall score.}
    \label{fig:teaser}
\end{figure}

Automatic text simplification (ATS) aims to improve language accessibility for diverse target group users, including language learners, persons with cognitive disabilities, and persons with low literacy levels \citep{gao2025evaluating}. While German ATS research has advanced with the emergence of multilingual LLMs and new parallel datasets \citep{ebling2022automatic,stodden2023deplain}, automatic evaluation relies on metrics like BLEU \citep{papineni2002bleu}, SARI \citep{xu2016sari}, and BERTScore \citep{zhang2020bertscore}, which focus on Ngram overlap or embedding similarity rather than directly measuring the core criteria of simplification quality: \textit{simplicity}, \textit{meaning preservation}, and \textit{fluency}, leading to weak correlations with human assessments~\citep{alva2021suitability}. Recent English-specific metrics like LENS~\citep{maddela2023lens} have addressed this challenge by training models on human-rated quality scores based on core criteria, achieving stronger alignment with human judgments than traditional metrics. However, no equivalent metric exists for German \citep{stodden2024easse}. 

\begin{figure*}[t!]
\centering
\includegraphics[width=\textwidth]{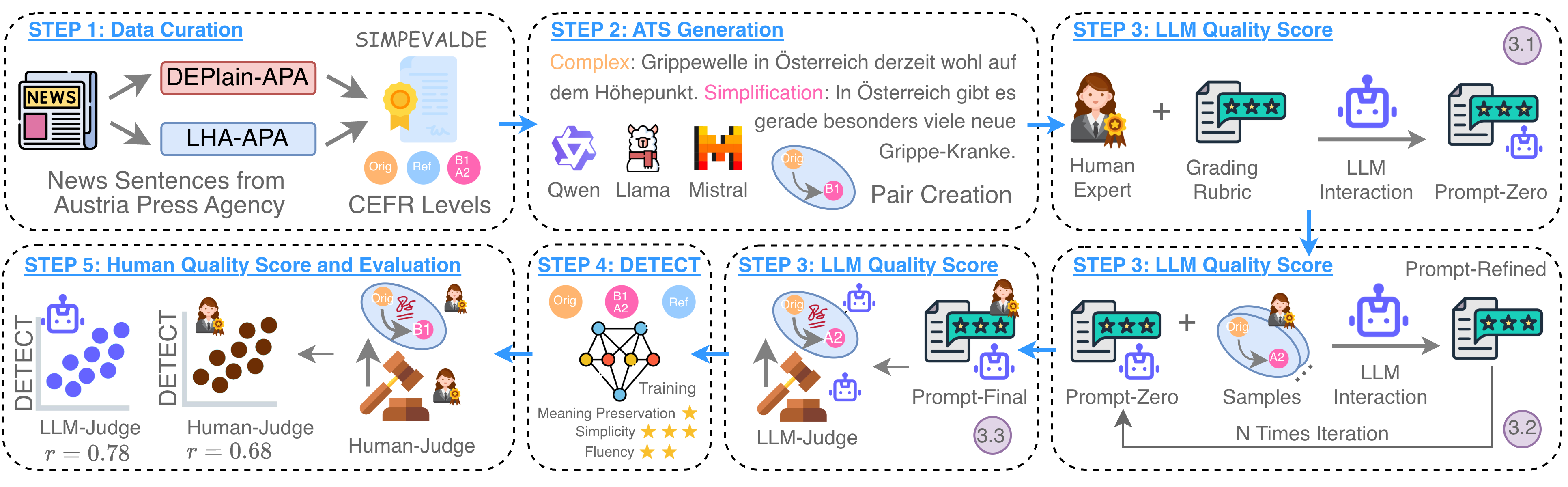}
\caption{Pipeline of \textsc{DEtect}. The framework consists of four steps: (1) \textbf{Data curation}. Construction of \textsc{SimpEvalDE} from Austrian Press Agency news sentences aligned across CEFR levels (DEPlaIn-APA and LHA-APA); (2) \textbf{ATS generation}. LLMs produce multiple simplifications for each complex sentence; (3)\textbf{ LLM quality scoring}. Open-source LLMs are prompted as judges to assign ATS quality scores using a prompt developed through iterative human expert feedback on the original LENS prompt; (4) \textbf{\textsc{DEtect}}: A feed-forward neural network is trained to predict \textit{simplicity}, \textit{meaning preservation} and \textit{fluency} from the synthetic LLM scores; (5) \textbf{Evaluation}. \textsc{DEtect} is validated against human quality judgments using a simplified \textsc{Rank \& Rate} protocol \citep{maddela2023lens} and compared to standard ATS metrics.}
\label{fig:detect-graph}
\end{figure*}

To address this research gap, we introduce \textsc{DEtect}, the first German-specific ATS evaluation metric designed to directly assess text simplification quality across the three core dimensions mentioned above. Unlike LENS, which relies on human annotations, we leverage LLMs as synthetic evaluators to generate training data for this learnable metric. We construct \textsc{SimpEvalDE}, a benchmark dataset combining existing German corpora with outputs from six ATS models, annotated with quality scores from three open-source LLMs. Additionally, we explore LLM-based refinement of grading criteria to improve evaluation consistency. Our work addresses three key research questions (RQs) in the ATS context: 
\paragraph{RQ1} Can quality-based evaluation metrics be adapted for German without human-annotated data?
\paragraph{RQ2} Can LLMs provide reliable supervision for training learnable evaluation metrics?
\paragraph{RQ3} Can LLMs systematically refine evaluation rubrics to improve judgment consistency?

With \textsc{DEtect}, we aim at proposing general guidelines for language accessibility tasks when leveraging LLMs in a Human-in-the-Loop setting, thereby contributing to a deeper understanding of how synthetic evaluation pipelines can complement human judgment and foster the development of more reliable, cross-linguistic evaluation frameworks.

\section{\textsc{DEtect}: Learnable Evaluation Metric for German Text Simplification}

The methodology of \textsc{DEtect} is designed to closely follow the LENS framework \cite{maddela2023lens}, but it replaces human-annotated data with LLM-generated synthetic data and refines the grading rubric through an iterative LLM-Human feedback loop. It constructs a synthetic supervision pipeline in four main steps (see Figure~\ref{fig:detect-graph}):
\begin{itemize}[noitemsep, topsep=0pt, partopsep=0pt, parsep=0pt, left=0pt]
\item \textcolor{StepBlue}{STEP 1}: We compile a parallel corpus of complex sentences and reference simplifications, denoted \textsc{SimpEvalDE}.
\item \textcolor{StepBlue}{STEP 2}: For each complex sentence in \textsc{SimpEvalDE}, six automatic text simplifications are generated using different LLMs.
\item \textcolor{StepBlue}{STEP 3}: Each generated simplification is scored along three dimensions: \textit{simplicity}, \textit{meaning preservation}, and \textit{fluency}, by prompting three open-source LLMs as judges. These scores are then used to train \textsc{DEtect}. Specifically:
\begin{itemize}[noitemsep, topsep=0pt, partopsep=0pt, parsep=0pt, left=2pt]
\item \textit{Prompt-Zero} is created by prompting GPT-4o \citep{openai2024gpt4o} to address identified limitations in the existing ATS instructions and is iteratively refined through a multi-round process using diagnostic examples and expert feedback to produce \textit{Prompt-Final};
\item Three distilled LLMs are prompted with \textit{Prompt-Final} to provide quality scores for \textit{simplicity}, \textit{meaning preservation} and \textit{fluency} for the entire dataset.
\end{itemize}
\item \textcolor{StepBlue}{STEP 4}: The training set is used to fine-tune \textsc{DEtect}, implemented as a RoBERTa-based \citep{liu2019roberta} feed-foward neural network (FFNN) to predict LLM-derived quality scores from the similarity between a text simplification, its complex source, and references.
\item \textcolor{StepBlue}{STEP 5}: 
The test set of \textsc{SimpEvalDE} is also manually scored by human annotators according to the same criteria, following a simplified version of the \textsc{Rank \& Rate} protocol introduced in LENS \citep{maddela2023lens}.Then the trained \textsc{DEtect} model is evaluated on the test set by measuring its correlation with both human and LLM quality scores, as well as with established ATS metrics such as SARI \citep{xu2016sari}, BLEU \citep{papineni2002bleu}, and BERTScore \citep{zhang2020bertscore}.
\end{itemize}

\subsection{\textsc{SimpEvalDE}}

The \textsc{SimpEvalDE} dataset is designed to mirror the structure of \textsc{SimpEval}\textsubscript{\textsc{past}} (100 training sentences) and \textsc{SimpEval}\textsubscript{2022} (60 test sentences) introduced by LENS, where the target simplification level was not explicitly defined. Our manual analysis indicates that the simplifications in these datasets achieve an average Flesch Reading Ease (FRE; \citet{flesch1948yardstick}) score of 67, which has been reported to roughly correspond to CEFR level B1 \citep{council2020cefrcv, natova2021estimating}. While FRE scores cannot reliably predict CEFR levels, particularly outside of English, our analysis of A2- and B1-level sentences in the APA datasets suggests an approximate correspondence.

Building on this observation, we turn to existing German resources for ATS, which provide material at CEFR levels B1 and A2. Two variants are available: (i) LHA-APA \citep{spring2021exploring}, consisting of 1,828 automatically aligned Complex-Simple pairs at CEFR B1/A2, and (ii) the smaller \textsc{DEplain-apa} \citep{stodden2023deplain}, which is manually aligned but lacks complex sentences. Given the importance of gold-standard references for training our metric, we attempt to merge \textsc{DEplain-apa} with exact matches from LHA-APA. This procedure produces only 26 complete matches, and further attempts to expand alignments for the B1/A2 levels reveal many incorrect pairs.Manual inspection confirms that the APA corpus, originally constructed for document-level simplification, poses challenges for sentence-level alignment. Even correctly matched sentences often fail to correspond fully, with simplifications either omitting key information from the complex sentence or introducing unrelated content (see Appendix~\ref{appendix:aligned-sentences} for examples). 

To address this, we design a systematic filtering approach: adjusted BERTScore is used to identify high-similarity matches across the LHA-APA corpus, followed by a manual review of candidate pairs. While this strategy successfully eliminates low-quality pairs, i.e., 219 out of 532 randomly selected low-similarity cases prove to be entirely incorrect, it ultimately yields only 44 additional gold-standard pairs.

To reach a size of 160 rows, we therefore relax the multi-reference requirement by manually including single-reference matches (38 Complex-Simple pairs at CEFR B1 and 19 at CEFR A2) and incorporate test set pairs from \textsc{DEplain-apa}, where both sentences fully preserve the same meaning.

The final dataset is then stratified across training and test sets based on match type and simplification strategy, following the LENS categories of \textit{delete}, \textit{split}, and \textit{paraphrase} operations; the distributions can be found in Appendix~\ref{appendix:desimpeval-lexicalvar}. We use an empirical algorithm to identify the simplification strategy, shown as follows: 
\begin{algorithm}
\caption{Identify simplification strategy}
\algrenewcommand{\alglinenumber}[1]{}
\setcounter{ALG@line}{0}
\begin{algorithmic}
\If{SentenceReductionRatio $> 1$}
    \State Simplification Strategy $\leftarrow$ \textit{split}
\ElsIf{SentenceReductionRatio $\leq 1$ \textbf{and} WordReductionRatio $< 1$}
    \State Simplification Strategy $\leftarrow$  \textit{delete}
\Else
    \State Simplification Strategy $\leftarrow$  \textit{paraphrase}
\EndIf
\end{algorithmic}
\label{simplificationtypealg}
\end{algorithm}

We show two random examples for each simplification strategy identified by this algorithm in Appendix~\ref{sec:appendix:simp-types} and present its statistical validity later in Section~\ref{sec:results}. We show the overall statistics of \textsc{SimpEvalDE} and the comparison with those in the original LHA-APA and \textsc{DEplain-apa} datasets in Appendix~\ref{appendix:desimpeval-lexicalvar} . 

\subsection{Generating Text Simplifications}
\label{method:ats_gn}

To generate ATS outputs for evaluation, we follow the two-category approach of LENS by selecting six high-performing LLMs: general-purpose instruction-tuned models and task-specific fine-tuned models.

For the instruction-tuned models, we include both German-specific systems (LeoLM;  \citet{pluester2023leolm} and DiscoLlama; \citet{discoleo2024llama3}) and multilingual variants (Qwen2-7B; \citet{qwen2024qwen2} and Llama3-8B; \citet{meta2024llama3}). All instruction-tuned models are prompted using a German translation of the LENS prompt, augmented with few-shot examples balanced across three simplification strategies. This design follows prior findings that few-shot prompting outperforms zero-shot settings \citep{ryan2023revisiting}. 

For the task-specific models, we use mBART-\textsc{DEplain} and mT5-\textsc{DEplain} \citep{stodden2023deplain}, both fine-tuned on the \textsc{DEplain-apa} dataset underlying \textsc{SimpEvalDE}. This setup avoids the need for additional training.

All models are run with their default parameters, except for LeoLM, where adjustments to \texttt{temperature} and \texttt{top\_p} were required to mitigate nonsensical generations observed during preliminary trials. Full configuration details and prompts are provided in Appendix~\ref{sec:appendix-atsgen}.

\subsection{Quality Score Labeling with LLMs}

Given the resource constraints that made full human evaluation infeasible, and motivated by the rationale to test the capacity of LLMs for this task, we rely on the LLM-as-a-Judge \citep{zheng2023judging} approach to produce quality score annotations for Complex-Simple pairs. Specifically, we adapt the LENS grading rubric into an LLM-compatible prompt through a Human-in-the-Loop \citep{chang2024survey} process with a high-capacity model. This prompt is then applied to three smaller distilled models (Distil-Llama-8B, Distil-Qwen-7B, and Zephyr-7B), whose outputs are averaged to yield the final \textit{LLM-Judge} score.

The original LENS rubric employs five discrete rating levels (0, 25, 50, 75, 100) to jointly assess \textit{simplicity}, \textit{meaning preservation}, and \textit{fluency}. However, this design poses several challenges for LLM-based evaluation: (1) vague definitions of core criteria such as ``\textit{simplicity}'' and the associated simplification levels; (2) an implicit assumption of parallel progression across all dimensions, despite evidence that \textit{simplicity} and \textit{meaning preservation} often trade off against each other \citep{cripwell2024evaluating}; and (3) hidden assumptions that require explicit clarification when translated into prompts for LLMs. These limitations hinder LLMs from consistently applying the rubric and exacerbate ambiguity in edge cases. 

To overcome this, we propose evaluating each criterion separately and aggregating the results with a weighted formula for the total score, with S = \textit{simplicity}, MP = \textit{meaning preservation}, and F = \textit{fluency}:
\begin{equation*}
\resizebox{\columnwidth}{!}{$
\begin{aligned}
& \text{Total Score} \\
& = 
\begin{cases}
\min(\text{MP}, \text{S}, \text{F}), & \text{if } \min(\text{MP}, \text{S}, \text{F}) < 25; \\
0.4 \text{MP} + 0.4 \text{S} + 0.2 \text{F}, & \text{otherwise}.
\end{cases}
\end{aligned}
$}
\end{equation*}
Following prior findings that \textit{simplicity} and \textit{meaning preservation} are more failure-prone than fluency \citep{cripwell2024evaluating}, we assign double weight to the former relative to the latter. In addition, we ground our prompt optimization in concrete German guidelines for Easy Language (German: \textit{Leichte Sprache}; \citet{bredel2016leichte}), incorporating explicit assumptions, for example, allowing models to assign any score between 0 and 100 rather than restricting them to five discrete levels. Prompt refinement is carried out with the assistance of GPT-4o \citep{openai2024gpt4o} as an iterative process:
\begin{itemize}[noitemsep, topsep=0pt, partopsep=0pt, parsep=0pt, left=0pt]
\item \textbf{Initial Prompt Creation}: GPT-4o generates \textit{Prompt-Zero} based on the original rubric and its identified limitations;
\item \textbf{Iterative Refinement}: Four diagnostic examples (meaning omission, added complexity, fluent but incorrect meaning, and unchanged input/output) are used to reveal weaknesses of the current prompt, with human feedback guiding revisions and discussions for the optimized prompt;
\item \textbf{Final Prompt Generation}: LLM scores and human feedback are integrated;
\item \textbf{Few-shot Enhancement}: Six additional examples graded by GPT-4o are incorporated to distill knowledge into smaller models.
\end{itemize}

All prompts, examples used, GPT-4o and human feedback, and the final few-shot examples can be found in Appendix~\ref{appendix:grading-rubric}. Our inspection of the refined prompt and the few-shot examples confirms that our approach successfully addresses most of the identified issues, providing clearer definitions of the evaluation criteria and improving the handling of edge cases. 

To select the three models for the final prompt and compute the \textit{LLM-Judge} score, we analyzed a set of DeepSeek distilled models \citep{guo2025deepseek}, including Distil-Qwen-1.5B, Distil-Qwen-7B, Distil-Qwen-14B, Distil-Llama-8B, and Zephyr-7B \citep{tunstall2024zephyr}. The analysis focused on correlations between the scores of smaller models ($\leq$8B) and the larger 14B model in order to assess whether the smaller models could serve as reliable proxies and whether they could produce sufficiently diverse outputs. Details of this analysis are provided in Appendix~\ref{appendix:model-selection}.

Our results show that Distil-Qwen-7B achieves correlations between $r=0.51$ and $r=0.87$ with Distil-Qwen-14B across the individual evaluation criteria. From ($\leq$8B) models, Zephyr-7B exhibits lower diversity in its scores, it still reaches correlations between $r=0.41$ and $r=0.58$ with Distil-Qwen-7B. Across all models, the highest correlations are consistently observed for \textit{meaning preservation} and the lowest for \textit{fluency}.

Based on these findings, we selected Distil-Llama-8B, Distil-Qwen-7B, and Zephyr-7B to calculate the final \textit{LLM-Judge} scores, aggregated as the arithmetic mean of their individual output quality scores.

\begin{figure*}[hbt!]
    \centering
    \begin{subfigure}[b]{0.32\textwidth}
        \includegraphics[width=\textwidth, height = 5cm]{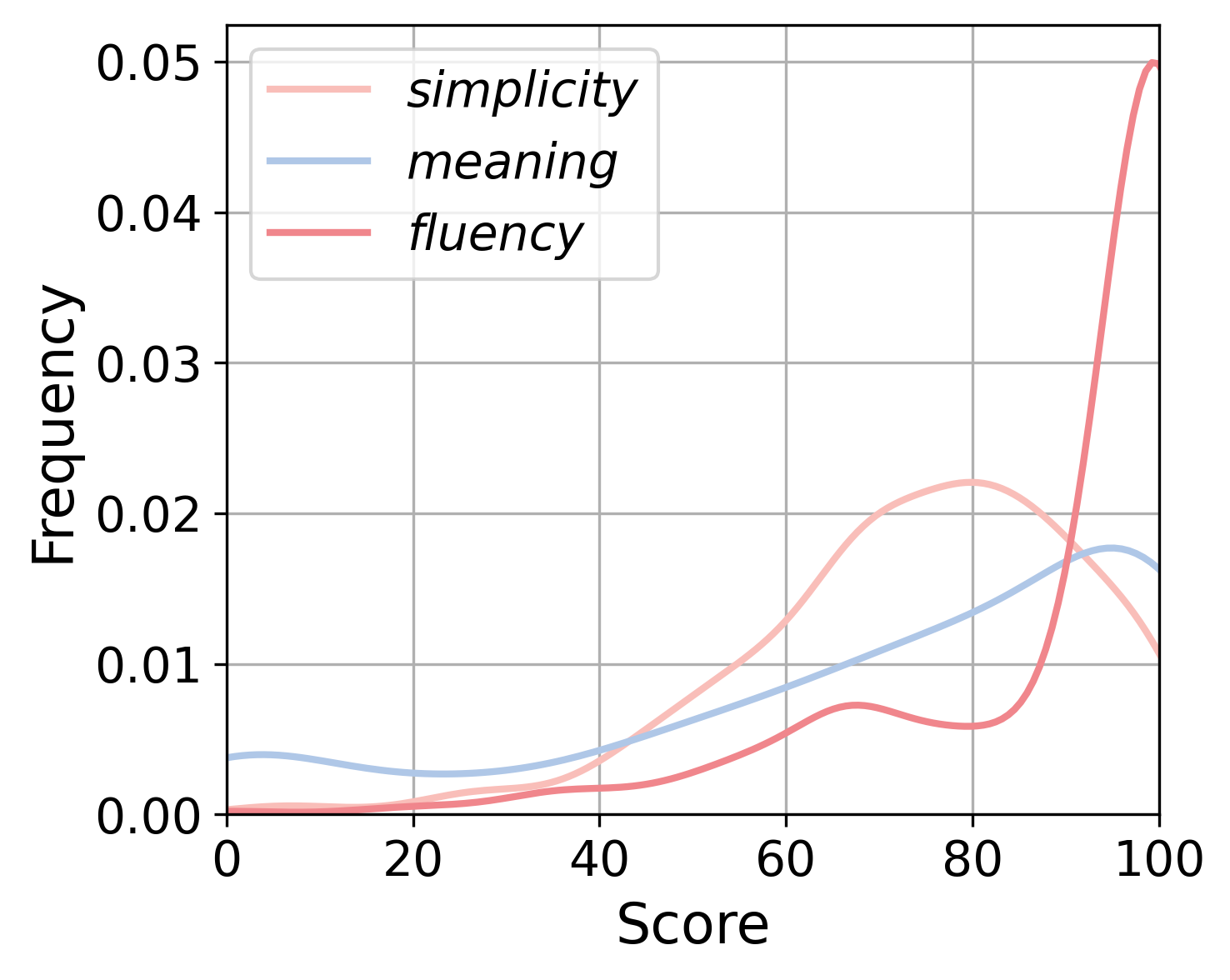}
         \caption{\textit{Human-Judge}}
    \end{subfigure}
     \begin{subfigure}[b]{0.32\textwidth}
         \includegraphics[width=\textwidth, height = 5cm]{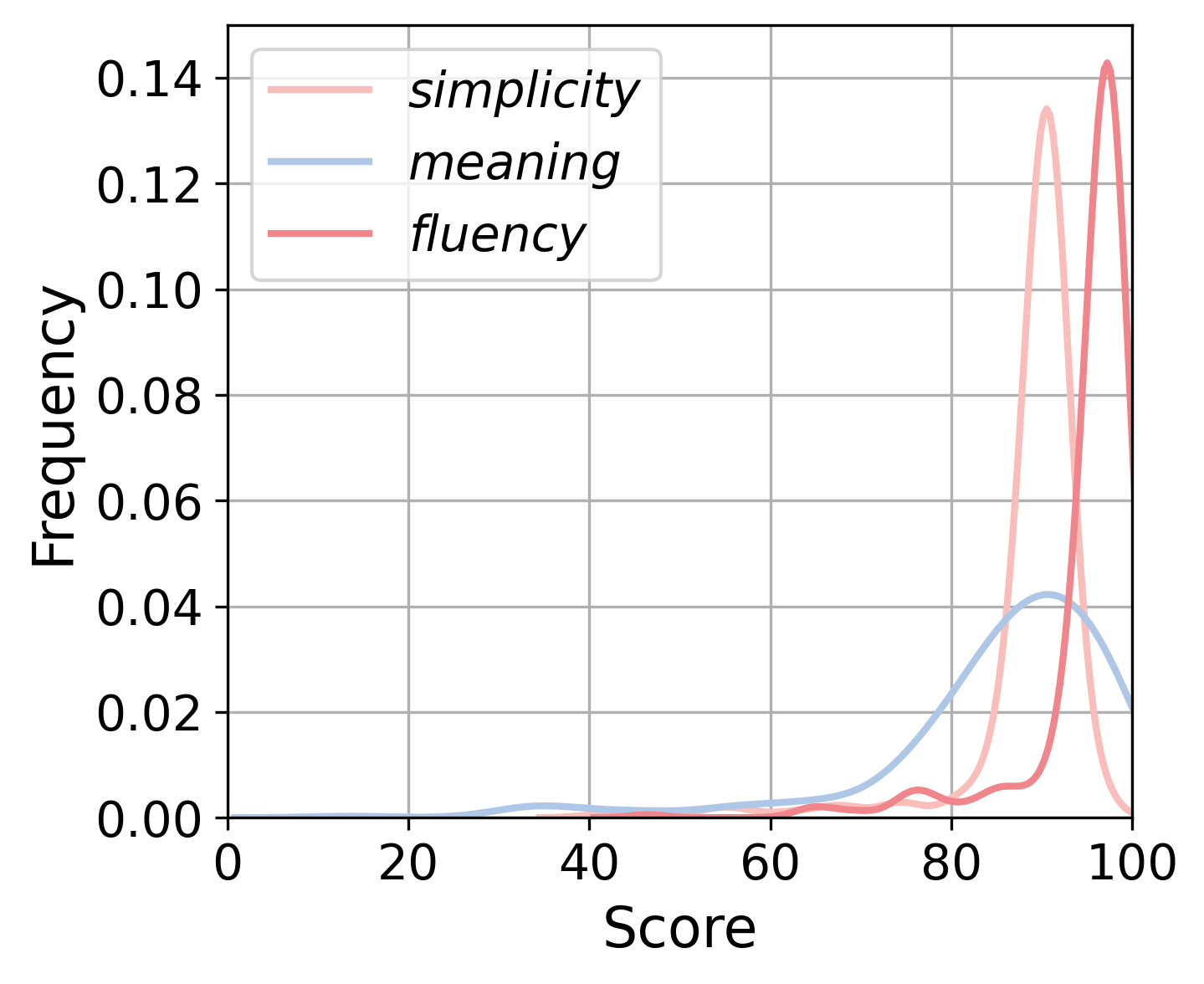}
         \caption{\textit{LLM-Judge}}
     \end{subfigure}
     \begin{subfigure}[b]{0.32\textwidth}
         \includegraphics[width=\textwidth, height = 5cm]{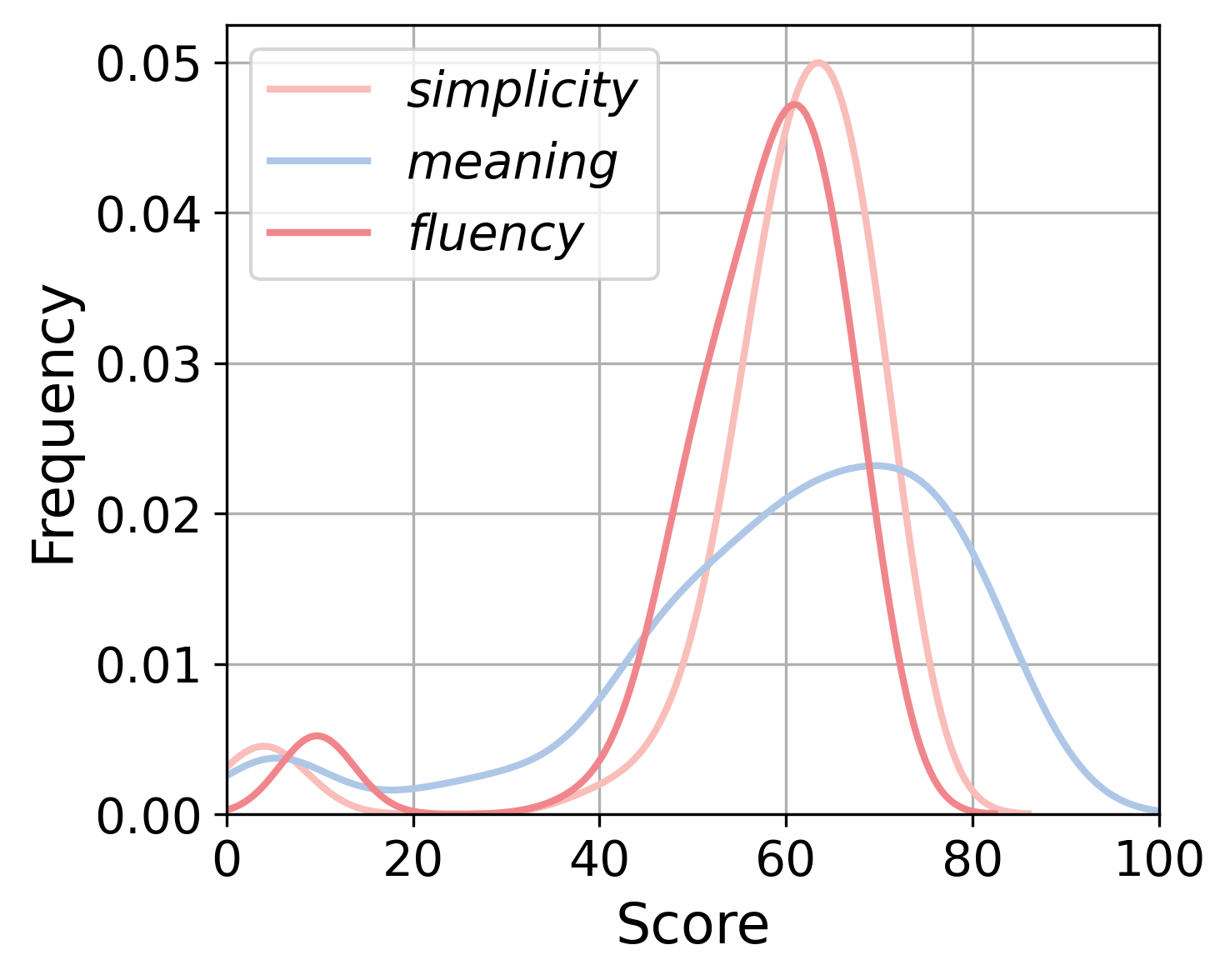}
         \caption{\textsc{DEtect}}
     \end{subfigure}
    \caption{Distributions of \textit{Human-Judge}, \textit{LLM-Judge} and \textsc{DEtect} by criterion}
    \label{fig_score_distributions}
\end{figure*}

\subsection{\textsc{DEtect} Training and Validation}
\label{method:detect-train}

\textsc{DEtect} builds on the LENS architecture, which concatenates embeddings of the complex sentence, simplified output, and references, along with their dot products and differences. These representations are then fed into a neural network. We introduce three key modifications. First, the model is adapted to predict three separate scores, i.e., \textit{simplicity}, \textit{meaning preservation}, and \textit{fluency}, instead of the single composite score used in LENS. Second, we incorporate German-specific embeddings WECHSEL \citep{minixhofer2022wechsel} in place of the multilingual RoBERTa-XML \citep{liu2019roberta} encoder used by LENS. Third, to address the considerably smaller training dataset, we tune hyperparameters such as the learning rate, dropout, and hidden layer size to reduce overfitting.

We explore several \textsc{DEtect} variants with different architectural and training configurations, with each variant building on insights from the previous run, all trained on five epochs and evaluated on the validation set using Pearson correlation and Spearman rank. The final model is selected based on the best validation performance across these metrics. To support multi-criteria prediction, the loss function is defined as the arithmetic mean of the three output scores. We report the configurations of different \textsc{DEtect} variants in Table~\ref{tab:detect-runs} in Appendix~\ref{appendix:desimpeval}.

\subsection{Human and Automatic Evaluation}
\label{method:human-eval}

We evaluate the predictions of \textsc{DEtect} on the \textsc{SimpEvalDE} test set against two reference measures: (1) averaged LLM quality scores (\textit{LLM-Judge}), which capture direct model accuracy using the same criteria on which the model was trained, and (2) human quality scores (\textit{Human-Judge}), which provide an independent benchmark based on human judgment.

Human scores were collected using a customized Google Sheets tool implementing a simplified \textsc{Rate \& Rank} procedure \citep{maddela2023lens}. Evaluators first classified each simplification by strategy (\textit{split}, \textit{paraphrase}, \textit{delete}), then rated \textit{simplicity}, \textit{meaning preservation}, and \textit{fluency} according to the same criteria used in the LLM-grading prompt. Finally, they verified score consistency via a separate ranking sheet (see Appendix~\ref{sec:app_detect_metrics} for details). Three voluntary native German-speaking experts independently graded all 360 test pairs. 

Finally, we benchmark \textsc{DEtect} against standard ATS evaluation metrics (i.e., BLEU \citep{papineni2002bleu}, SARI \citep{xu2016sari}, BERTScore Precision \citep{zhang2020bertscore}) by comparing their correlations with both \textit{LLM-Judge} and \textit{Human-Judge}.

\section{Results}
\label{sec:results}

\paragraph{\textit{Human-Judge} evaluation.}
Three expert graders annotated all 360 test-set examples for \textit{simplicity}, \textit{meaning preservation}, and \textit{fluency}. Inter-rater agreement is substantial, with Krippendorff's $\alpha$ \citep{krippendorff1970estimating} of $0.75$ for the total score, i.e., considerably higher than the $\alpha=0.32$ reported for the \textsc{SimpEval}\textsubscript{2022} grading in LENS. Agreement is strongest for \textit{meaning preservation} ($\alpha=0.78$), while \textit{simplicity} and \textit{fluency} showed lower levels ($\alpha=0.37$ and $\alpha=0.38$, respectively). For each criterion, the \textit{Human-Judge} score is the mean of three graders' ratings, from which the total score is computed.

As shown in Figure \ref{fig_score_distributions}, \textit{meaning preservation} exhibits the widest spread, with only one-third of examples receiving the maximum score and a local peak at 0, whereas \textit{fluency} is heavily right-skewed, with more than half of the examples achieving the maximum rating of 100. Correlation analysis shows that \textit{meaning preservation} aligns most strongly with the total score ($r=0.88$), followed by \textit{fluency} ($r=0.61$) and \textit{simplicity} ($r=0.56$). The weakest relationship is between \textit{simplicity} and \textit{meaning preservation} ($r=0.25$), consistent with the findings of \citet{cripwell2024evaluating}.

\begin{figure*}[hbt!]
    \centering
    \begin{subfigure}[b]{0.32\textwidth}
        \includegraphics[width=\textwidth, height = 5cm]{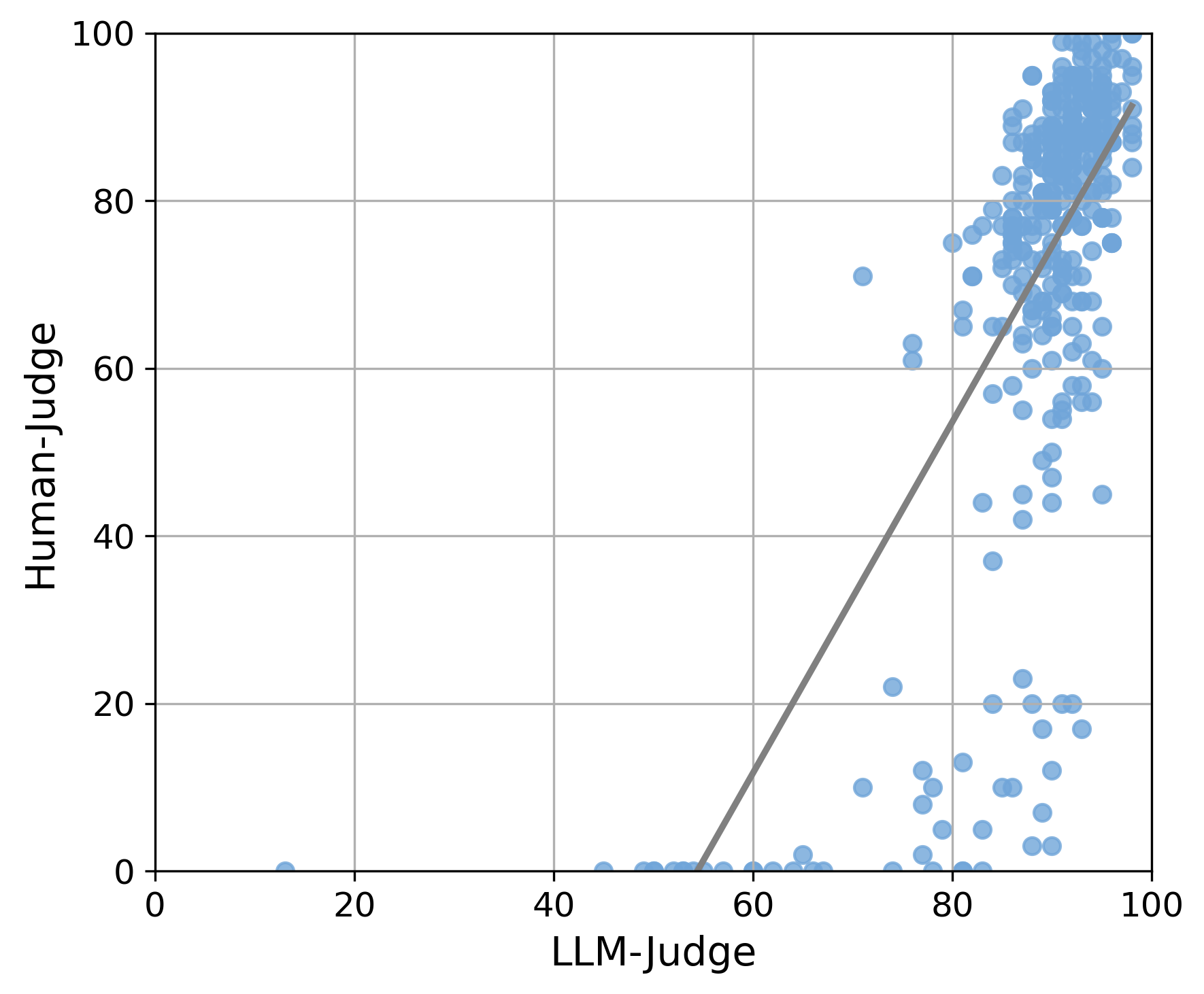}
        \caption{\textit{Human-Judge} vs. \textit{LLM-Judge}}
    \end{subfigure}
    \begin{subfigure}[b]{0.32\textwidth}
        \includegraphics[width=\textwidth, height = 5cm]{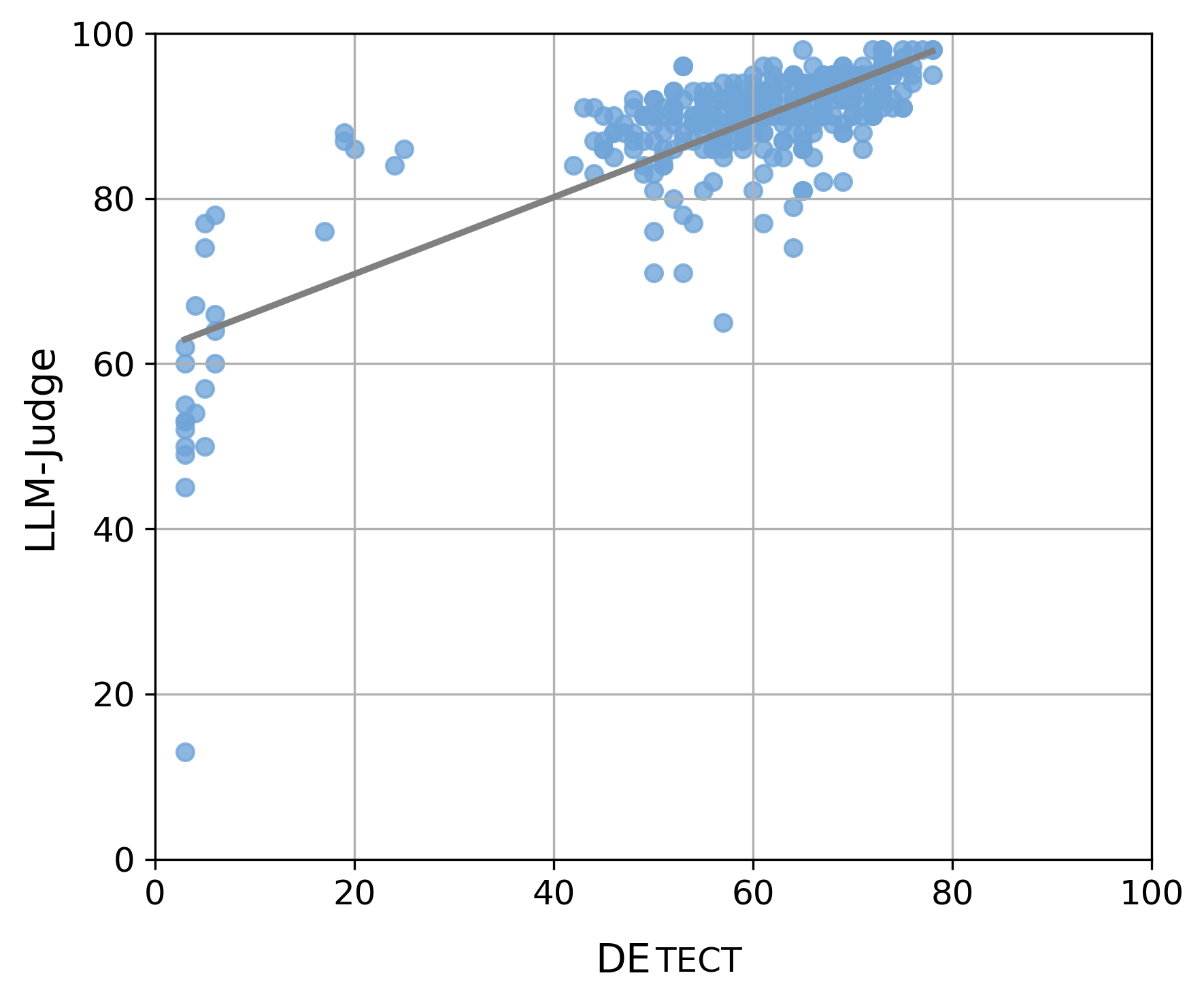}
        \caption{\textit{LLM-Judge} vs. \textsc{DEtect}}
    \end{subfigure}
    \begin{subfigure}[b]{0.32\textwidth}
        \includegraphics[width=\textwidth, height = 5cm]{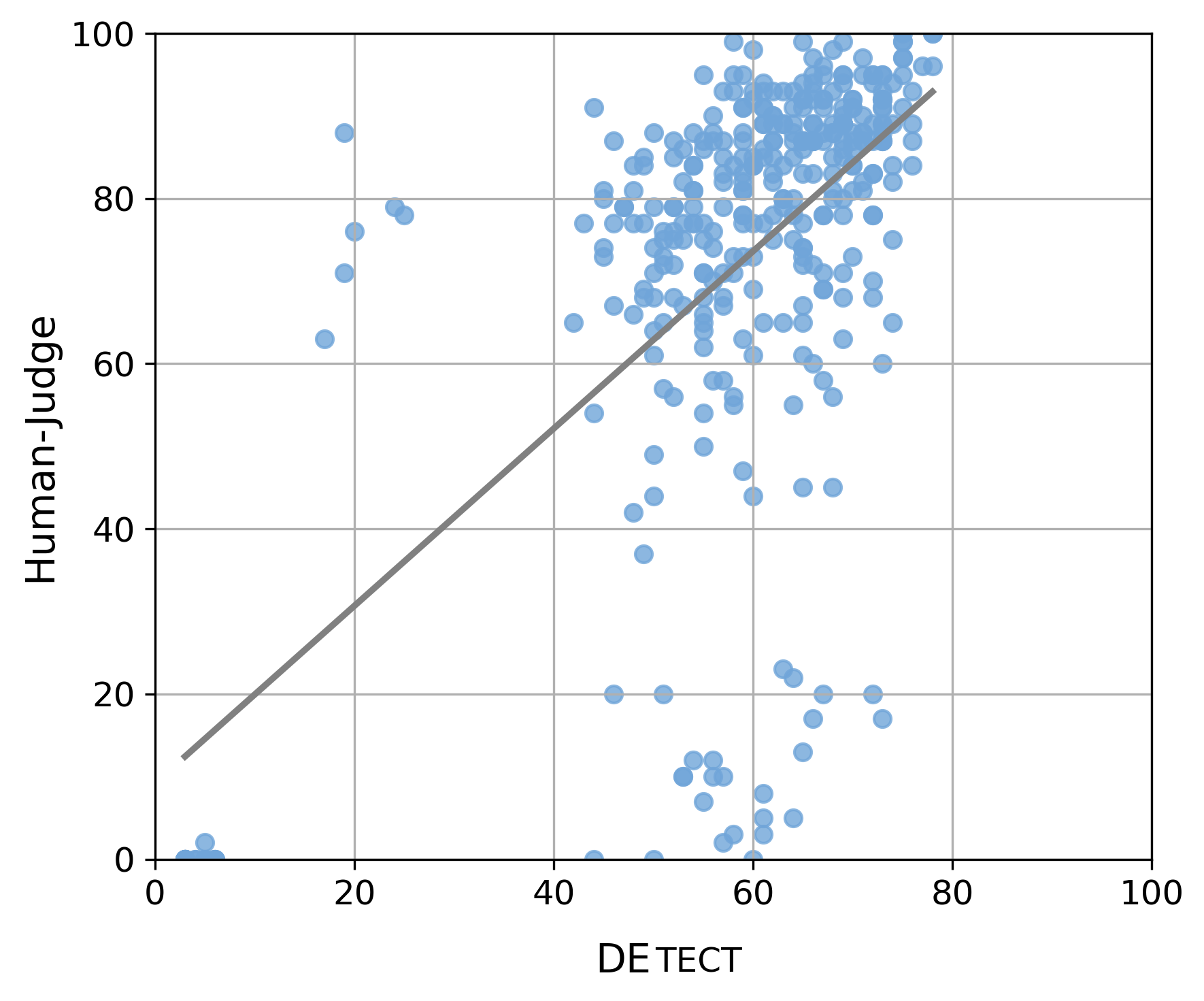}
        \caption{\textit{Human-Judge} vs. \textsc{DEtect}}
    \end{subfigure}
    \caption{Scatter plots of total \textit{Human-Judge}, \textit{LLM-Judge} and \textsc{DEtect} scores.}
    \label{scatter-total-criterions}
\end{figure*}

\paragraph{\textit{LLM-Judge} Evaluation.}

The three \textit{LLM-Judges} show lower inter-rater agreement ($\alpha=0.54$) compared to human graders ($\alpha=0.75$). As with human scoring, \textit{meaning preservation} achieves the highest agreement ($\alpha=0.60$), followed by \textit{fluency} ($\alpha=0.45$) and \textit{simplicity} ($\alpha=0.41$). \textit{LLM-Judge} score distributions shown in Figure \ref{fig_score_distributions} are narrower and more Gaussian-shaped than human ratings, with compressed interquartile ranges across all criteria, as well as higher median values for \textit{simplicity} and \textit{meaning preservation}. 

Correlation analysis shown in Table~\ref{tab:correla} shows that \textit{meaning preservation} aligns most strongly with human ratings ($r=0.77$), followed by \textit{fluency} ($r=0.45$), and \textit{simplicity} ($r=0.28$). Despite the lower two scores, the total score achieves a correlation of $r=0.72$. Figure~\ref{scatter-total-criterions} shows that while \textit{Human-Judge} and \textit{LLM-Judge} scores correlate strongly when both assign high total scores, a distinct cluster emerges where \textit{LLM-Judge} assigns scores between 40 and 100 but human ratings remain below 20. Additionally, unlike in human scores, all intra-criteria correlations for \textit{LLM-Judge} are high from $r=0.82$ between \textit{simplicity} and \textit{meaning preservation} to $r=0.94$ between \textit{simplicity} and \textit{fluency}, in contrast to the weak correlation of $r=0.25$ observed between \textit{simplicity} and \textit{meaning preservation} in the former.

\paragraph{\textsc{DEtect} training and model selection.}
Table~\ref{tab:detect-runs} in Appendix~\ref{appendix:desimpeval} reports the validation results for all \textsc{DEtect} variants, among which the \textsc{DEtect}\_multi\_wechsel\_reduced model achieves the best performance, with a Spearman rank correlation of 0.68 and a Pearson correlation of 0.87, and is therefore selected as the final model for evaluation. Following LENS, predicted scores are rescaled to the range [0, 100] by mapping them to the percentage of the area under a normal curve with a mean of 0 and a standard deviation of 1.

\begin{table}[htbp]
  \centering
  \renewcommand{\arraystretch}{1.1}
  \setlength{\tabcolsep}{2pt}
  \caption{Person correlations ($r$) between \textsc{DEtect}, \textit{Human(-Judge)} and \textit{LLM(-Judge)}.}
  \label{tab:correla}
  \resizebox{\columnwidth}{!}{
  \begin{tabular}{lcccc}
    \toprule
    \textbf{Comparison} &
    \makecell[c]{\textbf{\textit{Simplicity}}} &
    \makecell[c]{\textbf{\textit{Meaning P.}}} &
    \makecell[c]{\textbf{\textit{Fluency}}} &
    \makecell[c]{\textbf{Total}\\\textbf{Score}} \\
    \midrule
    \textit{Human} vs. \textit{LLM}    & 0.28 & 0.77 & 0.45 & 0.72 \\
    \textit{Human} vs. \textsc{DEtect}       & 0.32 & 0.68 & 0.35 & 0.64 \\
    \textsc{DEtect} vs. \textit{LLM}        & 0.82 & 0.80 & 0.78 & 0.80 \\
    \bottomrule
  \end{tabular}
  }
  
\end{table}

\paragraph{Comparison of \textsc{DEtect} with \textit{LLM-Judge}.}

As shown in Table~\ref{tab:correla}, \textsc{DEtect} exhibits strong correlations with \textit{LLM-Judge} scores (used for training the model) across all criteria: $r=0.78$ for \textit{fluency}, $r=0.80$ for \textit{meaning preservation}, $r=0.82$ for \textit{simplicity}, and $r=0.80$ overall.Visual inspection of the total-score scatter plot in Figure~\ref{scatter-total-criterions} shows distinct low- and high-score clusters of \textsc{DEtect} vs. \textit{LLM-Judge} scores, resultant of the bi-modal distribution of \textsc{DEtect} scores seen in Figure \ref{fig_score_distributions}; the fitted correlation line intersects both clusters, showing agreement between the two scorers at both ends of the scale. However, agreement differs in concentration between these regions: instances assigned scores of 20 or lower by \textsc{DEtect} correspond to a range of \textit{LLM-Judge} scores betwen 15 and 90, whereas instances assigned scores of 60 or higher by \textsc{DEtect} consistently receive \textit{LLM-Judge} scores of 70 or above, despite a larger number of observations in this region. 

\paragraph{Comparison of \textsc{DEtect} with \textit{Human-Judge}.} The correlation between \textsc{DEtect} predictions and human scores, reflecting the real performance of the model, mirrors the relationships observed between \textit{LLM-Judge} and human outputs. \textit{Meaning preservation} achieves the strongest correlation with human scores at $r=0.68$, while \textit{simplicity} shows the weakest performance at $r=0.32$, and \textit{fluency} demonstrates a similar correlation at $r=0.35$. The combined total score shows strong correlation with human judgment at $r = 0.64$. The scatter plot in Figure~\ref{scatter-total-criterions} shows overall agreement of high scores assigned by \textsc{DEtect} with human scores; however, the majority of the instances with low total scores as graded by humans are not graded as such by \textsc{DEtect} with only a few exceptions. Conversely, instances where \textsc{DEtect} assigns a low score while \textit{Human-Judge} does not are rare. Scatter plots for individual criteria (See Figure~\ref{fig:scatter-human-LLM} in Appendix~\ref{sec:app_detect_metrics}) demonstrate that \textit{meaning preservation} scores, while having the highest correlation with human judgment, exhibit the markedly highest number of instances of low human score and high \textsc{DEtect} score.


\begin{table}[htbp]
  \centering
  \renewcommand{\arraystretch}{1.1}
  \setlength{\tabcolsep}{5.5pt}
  \caption{Pearson correlations ($r$) between automatic metrics and human evaluation scores across criteria. Best values are highlighted in \textbf{bold}, second-best values are \underline{underlined}.}
  \label{tab:correlation_summary}
  \resizebox{\columnwidth}{!}{
  \begin{tabular}{lcccc}
    \toprule
    \textbf{Metric} &
    \makecell[c]{\textbf{\textit{Simplicity}}} &
    \makecell[c]{\textbf{\textit{Meaning P.}}} &
    \makecell[c]{\textbf{\textit{Fluency}}} &
    \makecell[c]{\textbf{Total}\\\textbf{Score}} \\
    \midrule
    BLEU        & 0.23 & 0.31 & 0.19 & 0.32 \\
    SARI        & 0.18 & 0.04 & 0.11 & 0.14 \\
    BERTScore   & \textbf{0.42} & \underline{0.48} & \underline{0.31} & \underline{0.55} \\
    \rowcolor{gray!20!white}
    \textbf{\textsc{DEtect}\footnotemark} & \underline{0.32} & \textbf{0.68} & \textbf{0.35} & \textbf{0.64} \\
    \bottomrule
  \end{tabular}
  }
  
\end{table}

\footnotetext{For metrics predicting separate scores for \textit{simplicity}, \textit{meaning preservation} and \textit{simplicity}, correlation is shown between each criteria and its metric, e.g., human \textit{simplicity} score vs. \textsc{DEtect} \textit{simplicity} score.}

\paragraph{Comparison with Standard ATS Metrics.}

Table~\ref{tab:correlation_summary} shows that \textsc{DEtect} substantially outperforms standard ATS metrics across nearly all evaluation dimensions, achieving the strongest correlations with human judgments in almost every category, as well as for the total score. The advantage is most pronounced for \textit{meaning preservation}, where \textsc{DEtect} reaches $r=0.68$, clearly surpassing BERTScore ($r=0.48$), BLEU ($r=0.31$), and SARI ($r=0.04$). For \textit{fluency}, \textsc{DEtect} again leads with $r=0.35$, compared to BERTScore ($r=0.31$), BLEU ($r=0.19$), and SARI ($r=0.11$). \textit{Simplicity} represents the weakest relative dimension, where \textsc{DEtect} achieves a lower correlation ($r=0.32$) than BERTScore ($r=0.42$), still outperforming BLEU ($r=0.23$) and SARI ($r=0.18$). Overall, \textsc{DEtect} achieves a correlation of $r=0.64$ with human judgments, outperforming BERTScore ($r=0.55$), BLEU ($r=0.32$), and SARI ($r=0.14$).
Further analysis (see Table~\ref{tab:detect-metrics-coorelations} in Appendix \ref{sec:app_detect_metrics}) confirms that \textsc{DEtect} achieves the highest correlations with human judgments across all simplification strategies (\textit{paraphrase}, \textit{delete}, \textit{split}) and all criteria. Among these, split-based simplifications yield the strongest correlations for every metric, including \textsc{DEtect}. The model does, however, show some degradation relative to its \textit{LLM-Judge} training data, particularly for delete- and paraphrase-based simplifications in \textit{meaning preservation}, and for delete-based \textit{fluency}.

\section{Discussion}
\label{sec:discussion}

\textsc{DEtect} exhibits strong alignment trained on LLM-generated quality scores, demonstrating effective adaptation even under the constraints of smaller datasets and fewer references compared to the original LENS framework:

\paragraph{RQ1: Metric Performance and Adaptability.}
Performance varies across evaluation criteria, with \textit{meaning preservation} achieving the highest correlation ($r=0.68$), followed by \textit{fluency} ($r=0.35$) and \textit{simplicity} ($r=0.32$). Despite an over-representation of paraphrase-based simplifications in the dataset, the metric performs particularly well on split-based cases and worst on paraphrasing, likely reflecting the greater semantic complexity of paraphrase evaluation under reduced model capacity; a more balanced dataset could further substantiate this interpretation. Overall, \textsc{DEtect} consistently outperforms standard ATS metrics for \textit{meaning preservation} and \textit{fluency} across simplification strategies. We observe that the predictions mapped against human scores tend to form high- and low-quality clusters, showing stronger agreement with human judgments for high-quality outputs, while predictions for low-quality outputs are less reliable. Moreover, because \textsc{DEtect} rarely assigns intermediate scores, it is not well suited for fine-grained ranking among similarly scored candidates. Additionally, because the outputs are rescaled for comparability, they no longer correspond directly to the discrete levels of the original rubric, which reduces interpretability. Finally, the model is trained and evaluated exclusively on news-domain data, leaving its generalization to other domains unclear. Future work should explore adapting the model on data from other domains, as well as incorporating more intermediate-quality outputs and a more balanced distribution of simplification strategies, to improve fine-grained ranking, interpretability, and generalization.

\paragraph{RQ2: LLM-Based Annotation Reliability.}
LLM-generated quality scores provide reasonably reliable supervision signals depending on criterion, showing correlations with human judgments ranging from $r=0.77$ for \textit{meaning preservation} to $r=0.28$ for \textit{simplicity}. Error analysis highlights several limitations: (1) instability across repeated evaluations of identical inputs; (2) occasional German-specific misinterpretations, likely due to the underlying models being trained primarily on English and Chinese data \citep{guo2025deepseek}; and (3) incomplete adherence to grading instructions, particularly regarding the penalization of hallucinated content. Moreover, LLMs exhibit artificially high intra-criterion correlations ($r>0.8$), in contrast to human evaluators ($r=0.25$) between \textit{simplicity} and \textit{meaning preservation}, suggesting limited sensitivity to the independence of criteria, even when providing seemingly coherent rationales. The reliance on distilled models may have amplified these issues, indicating that larger and more capable models could offer more stable and instruction-following behavior, thereby improving the reliability of synthetic data.

\paragraph{RQ3: Rubric Refinement Impact.}
The inter-rater agreement on the total score in our experiment with the reviewers using \textit{Prompt-Final}  ($\alpha=0.75$) is substantially higher than the one reported in LENS with reviewers using the original criteria that the LLM was instructed to improve upon. ($\alpha=0.32$). While these figures are not directly comparable due to the different sizes of the dataset, our observed improvement in the scoring done by LLMs during the prompt development phase, paired with observed integrations of the identified issues that the LLM was instructed to incorporate in \textit{Prompt-Final} provides indication that LLMs can play a role in enhancing rubric clarity for both human and machine evaluators. Nonetheless, the analysis of LLM prompt feedback highlights remaining ambiguities, particularly in defining what qualifies as ``important information'' in deletion-based simplifications and in determining how to assess one criterion when others fail completely, pointing to areas where even refined rubrics require further clarification. An important limitation in evaluating our approach is that to reduce the computational load, the LLMs that were used for \textit{LLM-Judge} are considerably smaller than the model used to develop \textit{Prompt-Final}. Given known prompt sensitivity across model scales, future work should assess whether rubric refinements derived from larger models consistently improve grading quality for smaller architectures. 

\section{Related Work}
\label{sec:previous-work}

ATS aims to transform complex text into simpler variants while preserving meaning and grammaticality \citep{alva2021suitability}. 
Originally rule-based, ATS progressed through statistical machine translation \citep{alva2020data} and neural sequence-to-sequence models \citep{hochreiter1997long}, with transformer architectures becoming dominant for both sentence- and document-level simplification \citep{nisioi2017exploring, zhang2017sentence, rios2021document}. Recent advances in LLMs enable zero- and few-shot simplification, which often outperforms task-specific models on English benchmarks \citep{kew2023bless, maddela2023lens}. 
In German, ATS work ranges from rule-based systems implementing \emph{Leichte Sprache} guidelines \citep{suter2016rule} to neural approaches, with mBART-based adaptations prevailing as the state-of-the-art \citep{rios2021document, anschutz2023language, carrer2024towards, stodden2024reproduction}. Emerging multilingual LLMs such as BLOOM \citep{scao2022bloom} show promise for German in zero- and few-shot regimes \citep{stodden2024reproduction}.

Sentence-level ATS requires parallel corpora of complex-simple pairs obtained through manual or automatic alignment \citep{nikolov2019large}. Such resources often involve many-to-many mappings that complicate supervision \citep{ebling2022automatic, stodden2023deplain}. A central stream of work relies on the Austrian Press Agency (APA) news with CEFR B1/A2 targets, including APA-RST, APA-LHA, and \textsc{DEplain} variants \citep{sauberli2020benchmarking, spring2021exploring, stodden2023deplain, hewett2023apa}. Other corpora include the Simple German Corpus targeting \emph{Leichte Sprache} \citep{toborek2023new}, as well as document-level collections such as \textit{20 Minuten} and \textit{Klexikon} \citep{rios2021document, aumiller2022klexikon}. Consolidation efforts like De-Lite \citep{jablotschkin2024lite} aim to unify and deduplicate sources while normalizing complexity labels. Web-derived datasets provide larger coverage but are heterogeneous, whereas professionally simplified texts are of higher quality but less accessible \citep{battisti2020corpus, ebling2022automatic}.

Evaluation of ATS typically combines human and automated assessment. Human evaluation focuses on \emph{simplicity}, \emph{meaning preservation}, and \emph{fluency} \citep{alva2020data}, though inter-rater reliability remains fragile, particularly for simplicity under absolute scales \citep{stodden2021scale, alva2021suitability}. Furthermore, the definition of ``simplicity'' remains highly audience-dependent, shaped by goals in accessibility, domain adaptation, or language learning \citep{stajner2021automatic, garimella2022text}. Automatic metrics are widely used despite their shortcomings. BLEU \citep{papineni2002bleu} and SARI \citep{xu2016sari} rely on N-gram overlap, while BERTScore \citep{zhang2020bertscore} leverages contextual embeddings but may favor conservative or ungrammatical outputs. Readability measures such as Flesch Reading Ease (FRE; \citet{flesch1948yardstick}) and Flesch-Kincaid Grade Level (FKGL; \citet{kincaid1975}) are often reported but assess \textit{simplicity} only. Structure-aware or reference-free approaches such as SAMSA \citep{sulem2018semantic}, QuestEval \citep{scialom2021questeval}, and MeaningBERT \citep{ beauchemin2023meaningbert} target structural edits or semantic preservation.  Recent learnable metrics trained on human judgments, such as LENS \citep{maddela2023lens} that models continuous quality scoring based on scores grading \textit{simplicity}, \textit{meaning preservation}, and \textit{fluency} directly. 
For German, EASSE-DE \citep{stodden2024easse} adapts tokenization and metric choices, though LENS-style learnable metrics was not applied. Parallel research has also modeled German sentence complexity directly from human labels \citep{naderi2019textcomplexity, mohtaj2022transfer, mohtaj2022overview}.

Recent work further explores LLMs for evaluation and grading. Instruction-tuning and reinforcement learning from human feedback (RLHF) align models with instructions and preferences \citep{wei2021finetuned, ouyang2022training, touvron2023llama}, while distillation transfers these capabilities to smaller, more cost-efficient models \citep{li2024survey}. Prompt engineering, including example-based prompts \citep{zhou2022large}, chain-of-thought prompting \citep{wei2022chain}, and automatic prompt or rubric induction \citep{honovich2022instruction}, can substantially improve zero- and few-shot evaluation \citep{gupta2024carmo}. Within the paradigm of LLM-as-a-Judge \citep{zheng2023judging}, structured prompting methods such as G-Eval \citep{liu2023g} increase agreement with expert judgments and scale effectively to complex, multi-criteria assessments \citep{gu2024survey}. 

\section{Conclusion}
\label{sec:conclusion}



We introduced \textsc{DEtect}, the first German-specific learnable metric for evaluating text simplification across \textit{simplicity}, \textit{meaning preservation}, and \textit{fluency}. Developed by adapting the English LENS framework with synthetic LLM generated quality scores, \textsc{DEtect} fills the research gap in German text simplification evaluation and consistently outperforms BLEU, SARI, and BERTScore, with particularly strong results in \textit{meaning preservation}. The refined grading rubric derived from LLM feedback also improved inter-rater agreement and reduced the manual effort required for supervision, showing that rubric refinement can enhance consistency for both human and automated evaluation. While some challenges remain, such as confounding factors, LLM instability, linguistic ambiguities specific to German, and residual rubric uncertainty, our study demonstrates the feasibility of creating learnable evaluation metrics through human guided synthetic supervision and provides a scalable foundation for future research on text simplification assessment for language accessibility.

\section*{Limitations}

Our work has several limitations. \textsc{DEtect} relies on synthetic supervision from LLM-generated annotations, which can be unstable across runs, conflate criteria, and occasionally misinterpret German-specific constructions. The training and the evaluation is restricted to the news domain, limiting generalizability to other genres, such as educational or medical texts. In addition, the model tends to cluster outputs into broad quality groups rather than providing fine-grained rankings, and the score rescaling procedure reduces interpretability relative to the original rubric. Finally, this metric concerns only sentence-level simplification and is not directly applicable to document-level simplification. Our future work would therefore explore more robust annotator models, broader domain coverage, and rubric refinements that better balance interpretability with predictive power.

\section*{Acknowledgments}
 This work was supported by the Swiss Innovation Agency Innosuisse, Flagship Inclusive Information and Communication Technology (IICT), funding no. PFFS-21-47. We thank Sarah Ebling, Annette Rios, Patricia Scheurer, and Anja Ryser for their valuable contributions to the study.

\bibliography{custom}

\newpage
\appendix

\section{The \textsc{SimpEvalDE} Dataset}
\label{appendix:desimpeval}

\subsection{Inappropriate Training Examples}
\label{appendix:aligned-sentences}

Table~\ref{tab:poor_simplifications} shows text simplifications we have excluded as gold-standard training samples.

\begin{table}[!htbp]
\centering
\renewcommand{\arraystretch}{1.25}
\resizebox{\columnwidth}{!}{
\normalsize
\begin{tabular}{p{4cm}p{4cm}m{3cm}}
\toprule
\textbf{Complex} & \textbf{Simplification} & \textbf{Issue} \\
\hline
Insgesamt mussten bereits mehr als 500 Menschen behandelt werden,
neun lagen auf der \textbf{Intensivstation}. & Hunderte Menschen wurden \textbf{in Sicherheit gebracht}.
Mehr als 500 Menschen mussten von \"Arzten behandelt werden. & \multirow{6}{3cm}{\begin{minipage}[c]{3cm}The simplification is unfaithful to the complex text, introducing fabricated information about people being brought to safety.\end{minipage}} \\
& \\
\textit{(In total, more than 500 people had to be treated, nine were in intensive care.)} &
\textit{(Hundreds of people were brought to safety. More than 500 people had to be treated by doctors.)} & \\
\midrule
Rund 15 Jahre zuvor hatte es in Indien nur noch etwa \textbf{1,400 Exemplare gegeben}. &
Vor rund 15 Jahren gab es in Indien nur noch \textbf{1,400 Tiger}. &
\multirow{5}{3cm}{\begin{minipage}[c]{3cm}The simplification introduces extra hallucinated information about tigers that cannot be inferred from the complex sentence.\end{minipage}} \\
& \\
\textit{(Around 15 years earlier, there had been only about 1,400 specimens in India.)} &
\textit{(Around 15 years ago, there were only 1,400 tigers left in India.)} &
\\
\bottomrule
\end{tabular}
}
\caption{Example simplifications from the APA-LHA dataset that are correctly aligned to the complex texts, but are unfaithful or add extra information.}
\label{tab:poor_simplifications}
\end{table}



\subsection{Text Simplification Strategies}
\label{sec:appendix:simp-types}

Table~\ref{tab:simplification-types} gives examples of text simplifications using different strategies (\textit{delete}, \textit{split}, \textit{paraphrase}).

\begin{table}[!htbp]
\centering
\caption{Examples of complex-simple pairs with their assigned simplification strategies, determined using the holistic Algorithm~\ref{simplificationtypealg}. In the last case, the algorithm assigns \textit{delete}, while manual verification suggests \textit{paraphrase}; both assignments are considered acceptable.}
\label{tab:simplification-types}

\resizebox{\columnwidth}{!}{%
{\fontsize{8}{10}\selectfont   
\renewcommand{\arraystretch}{1.15}
\setlength{\tabcolsep}{4pt}

\begin{tabular}{p{4.2cm} p{4.2cm} p{1.6cm}}
\toprule
\textbf{Complex} & \textbf{Simplification} & \makecell{\textbf{Strategy}} \\
\midrule

Das hat Bildungsminister Heinz Fa\ss mann (\"OVP) bei einer Pressekonferenz am Freitag bekanntgegeben. &
Das hat der Bildungs-Minister Heinz Fa\ss mann gesagt. &
\multirow{7}{1.6cm}{\centering\makecell{Delete}} \\
& & \\
\textit{(Education Minister Heinz Fa\ss mann (\"OVP) announced this at a press conference on Friday.)} &
\textit{(Education Minister Heinz Fa\ss mann said this.)} & \\
\midrule

Grippewelle in \"Osterreich derzeit wohl auf dem H\"ohepunkt. &
In \"Osterreich gibt es gerade besonders viele neue Grippe-Kranke. &
\multirow{5}{1.6cm}{\centering\makecell{Paraphrase}} \\
& & \\
\textit{(Flu wave in Austria currently probably at its peak.)} &
\textit{(In Austria there are currently particularly many new flu patients.)} & \\
\midrule

Wenn die Temperaturen hoch sind, die Sonne viel scheint und es wenig regnet produzieren die Algen in dem See besonders viel Beta Carotin - und das Wasser wird pink. &
Wenn die Sonne viel scheint und es hei\ss wird, produzieren die Algen den Farbstoff Beta Carotin. Durch diesen Farbstoff und dem vielen Salz wird das Wasser dann pink. &
\multirow{11}{1.6cm}{\centering\makecell{Split}} \\
& & \\
\textit{(When temperatures are high, the sun shines a lot and it rains little, the algae in the lake produce particularly much beta carotene - and the water turns pink.)} &
\textit{(When the sun shines a lot and it gets hot, the algae produce the pigment beta carotene. Through this pigment and the much salt, the water then turns pink.)} & \\
\midrule

Man kann nur an die Menschen appellieren, sich keine \"Ubermengen an Medikamenten verschreiben zu lassen. &
Man soll sie sich wie immer ganz normal verschreiben lassen. &
\multirow{8}{1.6cm}{\centering\makecell{Delete/\\Paraphrase}} \\
& & \\
\textit{(One can only appeal to people not to have excessive amounts of medication prescribed.)} &
\textit{(One should have them prescribed normally as always.)} & \\

\bottomrule
\end{tabular}
}}
\end{table}

\subsection{Statistics of \textsc{SimpEvalDE}}
\label{appendix:desimpeval-lexicalvar}

Table~\ref{tab:lexical-stat-simpevalde} and Table~\ref{tab:lexical-stat-simpevalde-test} present the lexical statistics of training and test set of \textsc{SimpEvalDE}, respectively. 

Table~\ref{tab:apa-lha-deplain} compares the lexical measures of APA-LHA and \textsc{DEplain-apa} dataset.

Table~\ref{tab:desimpeval-types} shows the simplification strategies in the training and test set of \textsc{SimpEvalDE}.

\begin{table*}[!ht]
\centering
\caption{Statistics of lexical measures for complex texts, CEFR B1 and A2 simplifications in the training set of \textsc{SimpEvalDE}. Q1, Q2, Q3 denote the first quartile, the median, and the third quartile, respectively.}
\label{tab:lexical-stat-simpevalde}
\resizebox{\textwidth}{!}{%
\begin{tabular}{lcccccccc}
\toprule
 \multirow{2.5}{*}{\textbf{Statistic}} & \multicolumn{2}{c}{\textbf{Complex}} & \multicolumn{3}{c}{\textbf{B1}} & \multicolumn{3}{c}{\textbf{A2}} \\
\cmidrule(lr){2-3} \cmidrule(lr){4-6} \cmidrule(lr){7-9}
 & \textbf{num\_words} & \textbf{avg\_word\_length} & \textbf{num\_words} & \textbf{avg\_word\_length} & \textbf{sentence\_count} & \textbf{num\_words} & \textbf{avg\_word\_length} & \textbf{sentence\_count} \\
\midrule
Count & 100.00 & 100.00 &  48.00 & 48.00 & 48.00 & 100.00 & 100.00 & 100.00 \\
Mean & 13.43 & 6.69 & 12.85 & 6.36 & 1.27 & 9.96 & 5.78 & 1.26 \\
Std. & 4.72 & 1.09  & 5.34 & 1.59 & 0.45 & 6.28 & 1.32 & 0.48 \\
Min. & 6.00 & 4.62  & 5.00 & 3.80 & 1.00 & 1.00 & 4.00 & 1.00 \\
Q1 & 10.00 & 6.00  & 9.00 & 5.20 & 1.00 & 6.00 & 4.67 & 1.00 \\
Q2 & 12.00 & 6.54  & 11.50 & 5.85 & 1.00 & 10.00 & 5.87 & 1.00 \\
Q3 & 17.00 & 7.07  & 16.00 & 7.21 & 2.00 & 14.00 & 6.86 & 1.00 \\
Max. & 28.00 & 10.45 & 28.00 & 12.00 & 2.00 & 27.00 & 9.50 & 3.00 \\
\bottomrule
\end{tabular}
}
\end{table*}

\begin{table*}[hbt]
\centering
\caption{Statistics of lexical measures for complex texts, CEFR B1 and A2 simplifications in the test set of \textsc{SimpEvalDE}.}
\label{tab:lexical-stat-simpevalde-test}
\resizebox{\textwidth}{!}{%
\begin{tabular}{lcccccccc}
\toprule
 \multirow{2.5}{*}{\textbf{Statistic}} & \multicolumn{2}{c}{\textbf{Complex}} & \multicolumn{3}{c}{\textbf{B1}} & \multicolumn{3}{c}{\textbf{A2}} \\
\cmidrule(lr){2-3} \cmidrule(lr){4-6} \cmidrule(lr){7-9}
 & \textbf{num\_words} & \textbf{avg\_word\_length} & \textbf{num\_words} & \textbf{avg\_word\_length} & \textbf{sentence\_count} & \textbf{num\_words} & \textbf{avg\_word\_length} & \textbf{sentence\_count} \\
\midrule
Count & 60.00 & 60.00 & 29.00 & 29.00 & 29.00 & 60.00 & 60.00 & 60.00 \\
Mean & 13.62 & 6.75 & 12.10 & 6.19 & 1.14 & 10.02 & 5.78 & 1.30 \\
Std. & 5.31 & 1.16 & 5.26 & 1.47 & 0.35 & 8.38 & 1.56 & 0.65 \\
Min. & 6.00 & 4.40 &  5.00 & 4.43 & 1.00 & 1.00 & 4.00 & 1.00 \\
Q1 & 10.00 & 5.89 &  8.00 & 5.17 & 1.00 & 2.50 & 4.00 & 1.00 \\
Q2 & 12.00 & 6.72 & 11.00 & 6.20 & 1.00 & 9.00 & 5.73 & 1.00 \\
Q3 & 17.00 & 7.38 & 15.00 & 6.60 & 1.00 & 14.00 & 6.79 & 1.00 \\
Max. & 28.00 & 10.14 & 26.00 & 10.40 & 2.00 & 40.00 & 10.33 & 4.00 \\
\bottomrule
\end{tabular}
}
\end{table*}



\begin{table*}[!ht]
\centering
\caption{Statistics of lexical measures for complex texts, CEFR B1 and A2 simplifications in the original APA-LHA dataset and \textsc{DEplain-apa} test set.}
\label{tab:apa-lha-deplain}
\begin{subtable}[t]{0.45\textwidth}
\centering
\caption{APA-LHA Dataset.}
\resizebox{\textwidth}{!}{
\begin{tabular}{lcc}
\toprule
\textbf{Statistic} & \textbf{num\_words\_compl} & \textbf{avg\_word\_length\_compl} \\
\midrule
Count   & 1828   & 1828 \\
Mean    & 17.36  & 6.39 \\
Std.     & 8.01   & 1.16 \\
Min.    & 3      & 3.50 \\
Q1    & 11     & 5.67 \\
Q2    & 17     & 6.29 \\
Q3   & 23     & 6.92 \\
Max.     & 42     & 17.20 \\
\bottomrule
\end{tabular}
}
\end{subtable}
\hfill
\begin{subtable}[t]{0.45\textwidth}
\centering
\caption{\textsc{DEplain-apa} Test Set.}
\resizebox{\textwidth}{!}{
\begin{tabular}{lcccc}
\toprule
\textbf{Statistic} & \textbf{B1\_words} & \textbf{B1\_length} & \textbf{A2\_words} & \textbf{A2\_length} \\
\midrule
Count & 1231 & 1231 & 1231 & 1231 \\
Mean  & 10.57 & 6.40 & 10.80 & 6.16 \\
Std.   & 4.68  & 1.56 & 5.04  & 1.56 \\
Min   & 2.00  & 3.71 & 1.00  & 3.47 \\
Q1  & 7.00  & 5.38 & 7.00  & 5.22 \\
Q2  & 10.00 & 6.11 & 10.00 & 5.86 \\
Q3  & 13.00 & 7.00 & 13.00 & 6.80 \\
Max.   & 45.00 & 16.67 & 48.00 & 26.00 \\
\bottomrule
\end{tabular}
}
\end{subtable}

\end{table*}

\begin{table*}[hbt!]
\centering
\caption{Count of simplification strategies across CEFR levels in the train and test sets of \textsc{SimpEvalDE}.}
\label{tab:desimpeval-types}
\resizebox{0.7\textwidth}{!}{
\begin{tabular}{l lccccc}
\hline
\textbf{Subset} & \textbf{Strategy} & \textbf{Complex-B1-A2} & \textbf{Complex-B1} & \textbf{B1-A2} & \textbf{Complex-A2} & \textbf{Total} \\
\hline
\multirow{3}{*}{Train} 
& \textit{delete}     & 5  & 5  & 3  & 5  & 18 \\
& \textit{paraphrase} & 15 & 12 & 24 & 4  & 55 \\
& \textit{split}      & 6  & 5  & 14 & 2  & 27 \\
\hline
\multirow{3}{*}{Test}  
& \textit{delete}     & 4  & 3  & 2  & 3  & 12 \\
& \textit{paraphrase} & 7  & 9  & 14 & 2  & 32 \\
& \textit{split}      & 3  & 3  & 9  & 1  & 16 \\
\hline
\end{tabular}
}
\end{table*}

\begin{table*}[!htbp]
\centering
\caption{Model configuration parameters and validation performance for \textsc{DEtect}.}
\label{tab:detect-runs}
\resizebox{\textwidth}{!}{%
\begin{tabular}{lcccccc}
\toprule
\textbf{Variant} & \textbf{Dropout} & \textbf{LR} & \textbf{Pre-trained Model} & \textbf{Hidden Sizes} & \textbf{Val. Pearson} & \textbf{Val. Spearman} \\
\midrule
\textsc{DEtect}\_multi & 0.10 & 3e-5 & xml-roberta-base & [2304, 768] & 0.86 & 0.64 \\
\textsc{DEtect}\_multi\_reg & 0.20 & 1e-5 & xml-roberta-base & [2304, 768] & 0.86 & 0.66 \\
\textsc{DEtect}\_multi\_reg\_wechsel & 0.20 & 1e-5 & roberta-base-wechsel-german & [2304, 768] & 0.87 & 0.67 \\
\textsc{DEtect}\_multi\_wechsel\_reduced & 0.10 & 1e-5 & roberta-base-wechsel-german & [128, 64] & 0.86 & 0.68 \\
\bottomrule
\end{tabular}
}
\end{table*}

\section{Configurations and Prompt Details}
\label{sec:appendix-atsgen}

We select mainstream open-source instruction-tuned and task-specific fine-tuned LLMs and their German variants as the backbone of ATS models.

\subsection{Instruction-Tuned Models}

\paragraph{LeoLM} \citet{pluester2023leolm}
\begin{itemize}[left=0pt]
    \item Model: German-specific Mistral \citep{jiang2023mistral7b} fine-tuned on diverse German instruction-following data.
    \item Configuration: \texttt{top\_p} = 0.95, \texttt{temperature} = 0.6 (adjusted due to repetition issues).
    \item HF Endpoint: \huggingface \href{https://huggingface.co/LeoLM/leo-mistral-hessianai-7b-chat}{\nolinkurl{LeoLM/leo-mistral-hessianai-7b-chat}}
\end{itemize}

\paragraph{DiscoLlama} \citet{discoleo2024llama3} 
\begin{itemize}[left=0pt]
    \item Model: German-specific instruction-tuned variant of Llama 3 \citep{grattafiori2024llama} designed for high-quality German text generation.
    \item Configuration: Default settings.
    \item HF Endpoint: \huggingface \href{https://huggingface.co/DiscoResearch/Llama3-DiscoLeo-Instruct-8B-v0.1}{\nolinkurl{DiscoResearch/Llama3-DiscoLeo-Instruct-8B-v0.1}}
\end{itemize}

\paragraph{Qwen-7B} \citet{qwen2024qwen2}  
\begin{itemize}[left=0pt]
    \item Model: A multilingual instruction-tuned Qwen model.
    \item Configuration: Default settings.
    \item HF Endpoint:~\huggingface \href{https://huggingface.co/Qwen/Qwen2-7B-Instruct}{\nolinkurl{Qwen/Qwen2-7B-Instruct}}
\end{itemize}

\paragraph{Llama-8B} \citet{meta2024llama3} 
\begin{itemize}[left=0pt]
    \item Model: A multilingual instruction-tuned Llama model.
    \item Configuration: Default settings.
    \item HF Endpoint: \huggingface \href{https://huggingface.co/meta-llama/Meta-Llama-3-8B-Instruct}{\nolinkurl{meta-Llama/Meta-Llama-3-8B-Instruct}} 
\end{itemize}

\subsection{Task-Specific Fine-Tuned Models}

\paragraph{mBART-\textsc{DEplain}} \citet{stodden2023deplain} 
\begin{itemize}[left=0pt]
    \item Model: Based on mBART \citep{chipman2022mbart} for multilingual text-to-text generation. Fine-tuned on \textsc{DEplain-apa} dataset targeting CEFR A2 level of text simplification. Previously used in German ATS research by \citet{rios2021document}, \citet{carrer2024towards}, and \citet{stodden2024reproduction}.
    \item Configuration: Default settings.
    \item HF Endpoint: \huggingface \href{https://huggingface.co/DEplain/trimmed_mbart_sents_apa_web}{\nolinkurl{DEplain/trimmed\_mbart\_sents\_apa}} 
\end{itemize}

\paragraph{mT5-DEplain} \citet{stodden2023deplain}  
\begin{itemize}[left=0pt]
    \item Model: Based on mT5  \citep{xue2021mt5} architecture for multilingual text-to-text generation. Fine-tuned on \textsc{DEplain-apa} dataset targeting CEFR A2 level of text simplification. Previously used by \citet{stodden2024reproduction} for reproduction studies.
    \item Configuration: Default settings.
    \item HF Endpoint: \huggingface \href{https://huggingface.co/DEplain/mt5-DEplain-APA}{DEplain/mt5-DEplain-APA}
\end{itemize}

\subsection{ATS Generation Prompt}

We used the following prompt to generate text simplifications. We omitted here the English translation of this prompt for less verbosity, however, we have written all LLM-as-a-Judge prompts in English to facilitate the Human-in-the-Loop process.

\begin{tcolorbox}[width=\columnwidth,
  colback=gray!5!white,
  colframe=gray!75!black,
  fontupper=\small,
  breakable,
  enhanced jigsaw
]

Bitte formulieren Sie den folgenden komplexen Satz um. Sie k\"onnen dies tun, indem Sie komplexe W\"orter durch einfachere Synonyme ersetzen (d. h. paraphrasieren), unwichtige Informationen l\"oschen (d.h. zusammenfassen) und/oder einen langen komplexen Satz in mehrere einfachere S\"atze aufteilen. Der/Die endg\"ultige(n) vereinfachte(n) Satz(e) muss/m\"ussen grammatikalisch korrekt und fl\"ussig sein und die Hauptgedanken des urspr\"unglichen Satzes beibehalten, ohne dessen Bedeutung zu ver\"andern. 

\medskip 

Ein einfacher Satz im Deutschen sollte:

\begin{itemize}[left=0pt]
    \item Einfache W\"orter und gebr\"auchliche W\"orter verwenden (seltene oder lange Substantive durch ein Synonym ersetzen oder zusammengesetzte W\"orter mit einem Bindestrich trennen, z. B. Atomabkommen -> Atom-Abkommen). Schwierige Begriffe vermeiden und gegebenenfalls erl\"autern.
    \item Einfache Satzstrukturen mit m\"oglichst wenigen Nebens\"atzen, koordinierten S\"atzen und Genitivkonstruktionen verwenden.
    \item Informationen in Klammern (z. B. Angela Merkel (die ehemalige Bundeskanzlerin)) oder in verschachtelten Kommas (z. B. Angela Merkel, die ehemalige Bundeskanzlerin, $\ldots$) vermeiden.
    \item Aktiv statt Passiv verwenden.
    \item Plusquamperfekt statt Plusquamperfekt verwenden
    \item Metaphern, Initialen/Abk\"urzungen, Sonderzeichen, Prozents\"atze oder gro\ss e Zahlen vermeiden
    \item Zahlen (vier) durch Ziffern (4) ersetzen
    \item Fremdw\"orter vermeiden, au\ss er sie sind sehr bekannt
    \item Positive statt negative S\"atze verwenden und gegebenenfalls direkte Rede mit W\"ortern wie \enquote{du} oder \enquote{sie} verwenden.
\end{itemize}

\{\texttt{five\_shot}\}

\medskip

Nur die vereinfachte Version ausgeben, ohne Einleitung oder Erkl\"arung.

\medskip

Eingabe: \{\texttt{text}\}

\medskip
Ausgabe:
\end{tcolorbox}

\section{LLM Grading Rubric Development}

\label{appendix:grading-rubric}

\subsection{Instruction to Generate \textit{Prompt-Zero}}

We used the following prompt to generate the initial Prompt-Zero for LLM-as-a-Judge evaluation.



\begin{tcolorbox}[width=\columnwidth,
  enhanced jigsaw,
  breakable,
  colback=gray!5!white,
  colframe=gray!75!black,
  fontupper=\small,
  width=\columnwidth
]

I need you to rewrite the following grading rubric so that it can be used as a prompt for a large language model to evaluate the quality of a simplified sentence(s) compared to its original version and fixes the issues outlined below.

\medskip
A complex sentence can be simplified by replacing complex words with simpler synonyms (i.e. paraphrasing), deleting unimportant information (i.e. condensing) and/or splitting a long complex sentence into several simpler sentences. The final simplified sentence(s) must be grammatically correct, flow smoothly and retain the main ideas of its original counterpart without changing its meaning.

\medskip
\textbf{Current Rubric}:

\medskip
Rate the simplified sentence(s) based on how well it simplifies the original sentence using the following rubric: 
\begin{itemize}[left=0pt]
    \item 100 - Only when the sentence is fully simplified, entirely fluent, and preserves the core meaning of the original sentence.
    \item 75 - The sentence is somewhat simpler, mostly fluent, and the meaning is close to the original sentence. 
    \item 50 - The sentence is simpler, somewhat fluent, and the meaning is similar to the original sentence. 
    \item 25 - The sentence is equivalently simple, still has some fluency, but the meaning is lost. 
    \item 0 - The sentence is completely unreadable. 
\end{itemize}

\textbf{Issues with the Current Rubric}:
\begin{enumerate}[left=0pt]
    \item Lack of clear definitions - The rubric does not explicitly define what makes a sentence simple, fluent, or meaningfully preserved.
    \item Vague language: Words like ``somewhat'' and ``similar'' lack precise meaning and could lead to differing interpretations by different evaluators. A good example of a grading criteria level could be: ``75 - Mostly simple but may retain minor complexity, such as one or two long words or slightly complex phrasing.''
    \item No separation of criteria - The rubric does not evaluate simplicity, meaning preservation, and fluency independently.
    \item Limited scoring flexibility - The model should be able to assign any score between 0 and 100, not just fixed levels. 
    \item Simplicity should not be about how well it is simplified compared to the original but how simple it is in general (i.e. if the original is already simple, it does not need to be simplified). 
    \item Meaning preservation does not account for situations where additional information is introduced not present in the original. 
    \item It is not clearly stated that the simplification can consist of two sentences if the splitting of a complex sentence simplifies reading.
\end{enumerate}

\textbf{Definitions of Key Criteria}: 
\begin{enumerate}[left=0pt]
    \item Simplicity: Measures how easy to understand the simplified sentence is, irrespective of the complexity of the original. A simple sentence in German should:
    \begin{itemize}[left=0pt]
        \item Use simple and common words (replace rare or long nouns with a synonym or split compound words using a hyphen, e.g. Atomabkommen -> Atom-Abkommen). Avoid difficult terms and explain if necessary.
        \item Use simple sentence structures with minimal subordinate clauses, coordinated clauses, and genitive constructions.
        \item Avoid including information in parentheses (e.g. Angela Merkel (die ehemalige Bundeskanzlerin)) or in nested commas (for example Angela Merkel, die ehemalige Bundeskanzlerin, ...).
        \item Use active as opposed to passive voice.
        \item Use past perfect instead of past simple.
        \item Avoid metaphors, initials/abbreviations, special characters, percentages, or big numbers.
        \item Replace numbers written as words (vier) with digits (4).
        \item Avoid foreign terms unless they are very well known.
        \item Use positive rather than negative sentences and direct speech with words like \enquote{du} or \enquote{Sie} where appropriate.
    \end{itemize}
    \item Meaning Preservation: Measures whether the simplified output adequately expresses the meaning of the original sentence, perhaps omitting the least important information. 
    \item Fluency: Measures grammatical correctness and fluency of the output. 
\end{enumerate}

\textbf{How to Rewrite the Rubric}: 
\begin{enumerate}[left=0pt]
    \item Each criterion (Simplicity, Meaning Preservation, Fluency) should be scored separately from 0 to 100. 
    \item Before providing the grading levels for each criterion, the criterion should be defined in detail, especially simplicity, and then grading levels provided.
    \item The grading rubric should have clearly defined levels (100, 75, 50, 25, 0), with concise and objective descriptions at each level that specify what makes a sentence highly simplified, moderately simplified, barely simplified, etc. Ensure that two different evaluators would not interpret the criteria differently. 
    \item The model should be instructed to assign a continuous score between 0-100. 
\end{enumerate}
The output format should be structured as follows:

\medskip
Feedback: {Explain strengths and weaknesses for each criterion, referencing {original} and {simplified}.} 

\medskip
Score: Simplicity: \{X\}, Meaning Preservation: \{Y\}, Fluency: \{Z\}.

\medskip
Now, generate the revised grading instruction prompt for a large language model following these guidelines and addressing the issues.
\end{tcolorbox}

\subsection{\textit{Prompt-Zero}}

We present the initial \textit{Prompt-Zero} as follows:

\begin{tcolorbox}[width=\columnwidth,
  enhanced jigsaw,
  breakable,
  colback=gray!5!white,
  colframe=gray!75!black,
  fontupper=\small,
  width=\columnwidth
]

\textbf{Task Overview}:

\medskip
Evaluate the quality of a simplified sentence or set of sentences by comparing it to its original version. The evaluation should focus on three main criteria: Simplicity, Meaning Preservation, and Fluency. Assign a separate score for each criterion on a scale from 0 to 100, and provide feedback highlighting strengths and weaknesses. The simplified version can consist of one or more sentences if splitting enhances readability.

\medskip
\textbf{Criteria Definitions and Grading Levels}:
\begin{enumerate}[left=0pt]
    \item Simplicity
    \begin{itemize}[left=0pt]
        \item Definition: Simplicity measures the ease of understanding the simplified sentence, irrespective of the original sentence's complexity. Key factors influencing simplicity include vocabulary, sentence structure, and clarity. 
        \item Guidelines for Simplicity: 
        \begin{itemize}[left=0pt]
            \item Use straightforward and common words, replacing rare or compound words with simpler synonyms.
            \item Maintain simple sentence structures with few subordinate clauses and minimal use of coordination and genitive forms.
            \item Present information without parentheses or nested commas.
            \item Favor active voice and past perfect tense over passive voice and simple past tense.
            \item Avoid metaphors, initials, special characters, percentages, big numbers, and complex terms.
            \item Replace numbers with digits and use positive sentences with direct speech (using \enquote{du} or \enquote{Sie} appropriately).
        \end{itemize}
        \item Grading Levels:
        \begin{itemize}[left=0pt]
            \item 100: Exceptionally simple language, very easy to understand, adheres strictly to simplicity guidelines.
            \item 75: Mostly simple with minor complexities, such as infrequent longer words or slightly complex structures.
            \item 50: Moderate simplicity, noticeable complexity in vocabulary or structure, room for improvement.
            \item 25: Complexity levels similar to the original, little effort towards simplification.
            \item 0: Highly complex and difficult to understand, fails to simplify the original sentence.
        \end{itemize}
    \end{itemize}
    \item Meaning Preservation
    \begin{itemize}[left=0pt]
        \item Definition: Meaning preservation evaluates whether the simplified version accurately conveys the core meaning of the original sentence, omitting only the least important details without introducing additional, unrelated information.
        \item Grading Levels:
        \begin{itemize}[left=0pt]
            \item 100: Fully maintains the original meaning, with no significant omissions or extraneous additions.
            \item 75: Mostly preserves the core message, with some minor omissions or slight additions.
            \item 50: Meaning is generally preserved but with notable omissions or the introduction of some unrelated information.
            \item 25: Basic meaning is lost, with significant omissions or many added details not present in the original.
            \item 0: Fails to convey the original meaning, with major discrepancies or irrelevant information.
        \end{itemize}
    \end{itemize}
    \item Fluency
    \begin{itemize}[left=0pt]
        \item Definition: Fluency assesses grammatical correctness and the natural flow of the simplified sentence. A fluent sentence reads smoothly, without awkward phrasing or grammatical errors.
        \item Grading Levels:
        \begin{itemize}[left=0pt]
            \item 100: Completely grammatically correct, flows naturally and reads smoothly.
            \item 75: Mostly fluent with minor grammatical errors or slight awkwardness.
            \item 50: Moderately fluent, with noticeable errors impacting reading flow.
            \item 25: Fluency is compromised, with frequent grammatical mistakes or clumsy phrasing.
            \item 0: Completely lacks fluency, with critical grammatical issues making the text unreadable.
        \end{itemize}
    \end{itemize}
\end{enumerate}

\textbf{Evaluation Output Format}:
\begin{itemize}[left=0pt]
    \item Feedback: Describe the strengths and weaknesses of the {simplified} sentence in comparison to the {original}. Use specific examples from the text to support the assessment within each criterion.
    \item Score:
    \begin{itemize}[left=0pt]
        \item Simplicity: \{X\}
        \item Meaning Preservation: \{Y\}
        \item Fluency: \{Z\}
    \end{itemize}
\end{itemize}

\textbf{Example Output}:
\begin{itemize}[left=0pt]
    \item Feedback:
    \begin{itemize}[left=0pt]
        \item Simplicity: The \texttt{\{simplified\_sentence\}} sentence uses approachable vocabulary, but includes one complex compound word. 
        \item Meaning Preservation: The main idea is preserved in \texttt{\{simplified\_sentence\}}, but some minor details were omitted.
        \item Fluency: \texttt{\{simplified\_sentence\}} reads smoothly except for one awkward phrase in the middle.
        \end{itemize}
        \item Score:
        \begin{itemize}[left=0pt]
            \item Simplicity: 85
            \item Meaning Preservation: 70
            \item Fluency: 80
        \end{itemize}
\end{itemize}

This comprehensive rubric and evaluation format ensures clear and objective assessment of sentence simplification, allowing for varied interpretations without losing consistency across different evaluators.
\end{tcolorbox}

\subsection{Iterative Prompt Refinement}
\label{appendix:gpt-40-feedback}
\textcolor{StepBlue}{Example 1} 

\medskip
\noindent
\textbf{Complex:}
Wenn es gelinge, die Todeszahl durch Eind\"ammungsma\ss nahmen auf 100.000 zu begrenzen, dann haben wir alle zusammen einen guten Job gemacht, sagte Trump am Sonntagabend im Wei\ss en Haus. \textit{(If we manage to limit the number of deaths to 100,000 through containment measures, then we have all done a good job together, Trump said on Sunday evening in the White House.)}

\medskip
\noindent
\textbf{Simplified:}
Trump sagte auch: Wenn nicht mehr als 100.000 Menschen sterben, dann haben wir einen guten Job gemacht. \textit{(Trump also said: If no more than 100,000 people die, then we have done a good job.)}

\medskip
\noindent
\robot \textbf{\textit{LLM-Judge} (GPT-4o):}
\begin{tcolorbox}[width=\columnwidth,
  enhanced jigsaw,
  breakable,
  colback=gray!5!white,
  colframe=gray!75!black,
  fontupper=\small,
  width=\columnwidth
]

    Evaluation of Sentence Simplification
    \begin{itemize}[left=0pt]
        \item Feedback:
        \begin{itemize}[left=0pt]
            \item Simplicity: The \texttt{\{simplified\_sentence\}} sentence achieves a relatively high level of simplicity by removing some of the more complex phrasing found in the \texttt{\{complex\_sentence\}}. For instance, it uses straightforward vocabulary, such as replacing ``Eind\"ammungsma\ss nahmen'' and ``im Wei\ss en Haus'' with simpler phrasing. However, it could further simplify by using more everyday language for ``sterben'' such as ``umkommen'' if the context allows.
            \item Meaning Preservation: The core meaning of the \texttt{\{complex\_sentence\}} is largely preserved in the \texttt{\{simplified\_sentence\}}. The simplification retains the primary message of Trump's statement regarding the death toll limit as an indicator of success. However, some contextual details, like the venue ``im Wei\ss en Haus'' and the specific timing ``am Sonntagabend'', are omitted. While these details may not be crucial, their removal somewhat changes the context.
            \item Fluency: The \texttt{\{simplified\_sentence\}} sentence is mostly fluent, with a clear grammatical structure and a smooth flow. There are no significant grammatical errors, and the sentence reads naturally. The use of direct speech is correctly integrated, maintaining coherence and readability.
        \end{itemize}
        \item Score:
        \begin{itemize}[left=0pt]
            \item Simplicity: 90
            \item Meaning Preservation: 80
            \item Fluency: 95
        \end{itemize}
    \end{itemize}
    
    The {simplified} sentence excels in simplicity and fluency, making it accessible and easy to read. While it mostly preserves the essential meaning, it could better maintain the context by considering the inclusion of key details present in the {original}. Overall, it is an effective simplification.
\end{tcolorbox}

\noindent
\human \textbf{\textit{Human-Judge}:}

\medskip
\noindent
\textbf{Simplicity:} The sentence is simpler indeed, although the sentence is somewhat long. I would grade 80.

\medskip
\noindent
\textbf{Meaning Preservation:} I agree with your feedback that there is some loss of information. Notably the omission of ``containment measures'' erases the information that people are dying due to COVID-19 and not another reason, which does somewhat alter the understanding. I would grade 60.

\medskip
\noindent
\textbf{Fluency:} I agree with your feedback but ``sagte auch'' is a bit confusing. I would give about 90.

\medskip
\medskip
\medskip
\noindent
\textcolor{StepBlue}{Example 2}

\medskip
\noindent
\textbf{Complex:}
 Liverpool steht wie im Vorjahr im Finale der Fu\ss ball-Champions-League. \textit{(Liverpool is in the final of the Champions League, just like last year.)}

\medskip
\noindent
\textbf{Simplified:}
Durch den \"uberraschend hohen Sieg steht Liverpool jetzt aber im Finale der Champions League.
\textit{(Thanks to the surprisingly high victory, Liverpool is now in the final of the Champions League.)}

\medskip
\noindent
\robot \textbf{\textit{LLM-Judge} (GPT-4o):}


\begin{tcolorbox}[width=\columnwidth,
  enhanced jigsaw,
  breakable,
  colback=gray!5!white,
  colframe=gray!75!black,
  fontupper=\small,
  width=\columnwidth
]
    Evaluation of Sentence Simplification
    \begin{itemize}[left=0pt]
        \item Feedback:
        \begin{itemize}[left=0pt]
            \item Simplicity: The \texttt{\{simplified\_sentence\}} sentence introduces slightly more complexity compared to the \texttt{\{complex\_sentence\}}. The \texttt{\{complex\_sentence\}} is already quite simple, with succinct phrasing. The inclusion of the phrase ``durch den \"uberraschend hohen Sieg'' in the \texttt{\{simplified\_sentence\}} introduces additional information, potentially increasing complexity without simplifying the original context. There is a shift to a more elaborate construction instead of the straightforward statement in the \texttt{\{complex\_sentence\}}.
            \item Meaning Preservation: The \texttt{\{simplified\_sentence\}} sentence retains the general idea that Liverpool is in the final of the Champions League. However, it introduces new information about an ``\"uberraschend hohen Sieg'', which was not present in the \texttt{\{complex\_sentence\}}. This addition could mislead readers about the context or suggest a narrative not provided in the \texttt{\{complex\_sentence\}}. The simplicity of the original indication of repetition from ``wie im Vorjahr'' is lost, altering the sequence or context regarding the team's progression to the final.
            \item Fluency: The \texttt{\{simplified\_sentence\}} sentence maintains fluency with grammatically correct structure and clear communication. It reads smoothly, but the introduction of the phrase ``jetzt aber'' could slightly disrupt the directness found in the \texttt{\{complex\_sentence\}}. The sentence still flows naturally, although it becomes slightly less straightforward due to the additional context.
        \end{itemize}
        \item Score: 
        \begin{itemize}[left=0pt]
            \item Simplicity: 60
            \item Meaning Preservation: 50
            \item Fluency: 85
        \end{itemize}
    \end{itemize}
    
    The \texttt{\{simplified\_sentence\}} sentence struggles to improve simplicity due to the introduction of extra information, which may divert from the original's straightforwardness. While it maintains overall fluency, the added context can obscure the precise meaning and relevance intended in the \texttt{\{complex\_sentence\}} sentence. Further simplification should aim at preserving the essence without unnecessary embellishments, focusing more directly on the primary message from the \texttt{\{complex\_sentence\}} without additional context unless essential.
\end{tcolorbox}

\noindent
\human \textbf{\textit{Human-Judge}:}

\medskip
\noindent
\textbf{Simplicity:} While the sentence is lexically simpler than the original, it is a syntactically more complicated structure. Adding information not present in the original also does it make it more complicated. TO INCORPORATE IN THE FINAL RUBRIC: Even if the final sentence is not extremely complex, due to it being more complex than the original, it should be below 25.

\medskip
\noindent
\textbf{Meaning Preservation:} Hallucinating information not present in the original should be highly penalized. While it does not change the sentence meaningfully as the main point is Liverpool advancing into the finals, the grading needs to score sentences that could mislead the reader low. TO INCORPORATE IN THE FINAL RUBRIC: Therefore, introduction of any information not present in the original should make the score low - below 50.

\medskip
\noindent
\textbf{Fluency:} While it is fluent, ``aber'' is not necessary and makes it slightly confusing. I would score also around 85-90. 

\medskip
\medskip
\medskip
\noindent
\textcolor{StepBlue}{Example 3}

\medskip
\noindent
\textbf{Complex:}
Die finanzielle Lage der Haushalte hat sich in der Coronakrise deutlich versch\"arft, geht aus einer Umfrage ( Sample 1.200 Personen ) der Tarifvergleichsplattform durchblicker hervor. \textit{(The financial situation of households has changed significantly during the Corona crisis, according to a survey (sample 1,200 people) by the tariff comparison platform durchblicker.)}

\medskip
\noindent
\textbf{Simplified:}
Rund 38 Prozent der Haushalte verdienen weniger als vor der Corona-Krise. Rund 67 Prozent der Haushalte wollen die Fixkosten senken. \textit{(Around 38 percent of households earn less than before the Corona crisis. Around 67 percent of households want to reduce fixed costs.)}

\medskip
\noindent
\robot \textbf{\textit{LLM-Judge} (GPT-4o):}

\begin{tcolorbox}[width=\columnwidth,
  enhanced jigsaw,
  breakable,
  colback=gray!5!white,
  colframe=gray!75!black,
  fontupper=\small,
  width=\columnwidth
]
    Evaluation of Sentence Simplification
    \begin{itemize}[left=0pt]
        \item  Feedback:
        \begin{itemize}[left=0pt]
            \item Simplicity: The \texttt{\{simplified\_sentence\}} sentences break down the information into two distinct, shorter statements, making it easier for the reader to process. This approach aids clarity and aligns well with simplicity guidelines. However, the \texttt{\{complex\_sentence\}} sentence's use of parentheses and specific phrasing is removed, potentially complicating comprehension by introducing new statistics that were not in the original, which adds unnecessary complexity instead of focusing on simplification.
            \item Meaning Preservation: The simplification significantly diverges from the \texttt{\{complex\_sentence\}} by introducing new statistical information that was not present initially. The \texttt{\{complex\_sentence\}} mentions a general deterioration in the financial situation, but the \texttt{\{simplified\_sentence\}} sentences provide specific figures (``38 Prozent'' for decreased income and ``67 Prozent'' intending to reduce fixed costs) that were not stated in the \texttt{\{complex\_sentence\}}, altering the intended message and context. Thus, it fails to adequately preserve the original meaning.
            \item Fluency: The \texttt{\{simplified\_sentence\}} sentences maintain grammatical accuracy and read smoothly, with clear connections between ideas. There is no awkward phrasing or noticeable grammatical errors, ensuring that the sentences flow naturally. The clear distinction of two separate ideas contributes to enhanced readability.
        \end{itemize}
        \item Score:
        \begin{itemize}[left=0pt]
            \item Simplicity: 40
            \item Meaning Preservation: 20
            \item Fluency: 90
        \end{itemize}
    \end{itemize}
    
    While the \texttt{\{simplified\_sentence\}} sentences attempt to present the information in a clearer manner, they introduce details not found in the \texttt{\{complex\_sentence\}}, complicating the overall message and context. It retains high fluency due to its grammatical correctness and logical flow but struggles in simplicity due to added complexity. For an effective simplification, the focus should remain on conveying the essential meaning without adding or omitting critical information from the \texttt{\{complex\_sentence\}} sentence.

    \medskip
    \noindent
    Notes on Incorporation into Final Rubric:
    \begin{enumerate}
        \item Simplicity: Ensure added information does not increase complexity compared to the original, and score below 25 if it does.
        \item Meaning Preservation: Penalize heavily (below 50) for any introduction of information not present in the original, even if it maintains core thematic elements.
        \item Fluency: Maintain high scores for sentences that are grammatically correct and smoothly flowing, even when meaning preservation is not fully achieved.
    \end{enumerate} 
\end{tcolorbox}

\noindent
\human \textbf{\textit{Human-Judge}:}

\medskip
\noindent
\textbf{Simplicity:} Here, the output was made significantly simpler than the original, which was a long sentence with two clauses with compound words and information inside parentheses. TO INCORPORATE IN THE FINAL RUBRIC: whether the original content is well reflected in the simplified output should only be assessed in Meaning Preservation category. The simplification splits the clauses into separate sentences in active voice, which is encouraged. While additional information was introduced, it is definitely easier to read. I agree that the concrete figures make it somewhat more complex, but it is still significantly easier than the original. I would give this simplification about 80-90. 

\medskip
\noindent
\textbf{Meaning Preservation:} While the general meaning of the household's situation worsening was retained in the first sentences, additional information not present in the original (38\%). Furthermore, the second sentence is completely unrelated. However, the information coming from a survey is a detail removing which does not change the meaning of the sentence. I would give this around 25. 

\medskip
\noindent
\textbf{Fluency:} Splitting a complicated sentence into two is encouraged in simplification, as long as the sentences are logically connected. Here, however, the splitting of the sentences does indeed produce a non-smooth transition between ideas. I would give this about 70. 
 
\subsection{\textit{Prompt-Final}}
\label{appendix:prompt-llm-few-shot}

We present the ultimate \textit{Prompt-Final} as follows:
\begin{tcolorbox}[width=\columnwidth,
  enhanced jigsaw,
  breakable,
  colback=gray!5!white,
  colframe=gray!75!black,
  fontupper=\small,
  width=\columnwidth
]         

\textbf{Task Overview}:

\medskip
\noindent
Evaluate the quality of a simplified German sentence or set of sentences by comparing it to its original version. The evaluation should focus on three main criteria: Simplicity, Meaning Preservation, and Fluency. Assign a separate score for each criterion on a scale from 0 to 100. The simplified version can consist of one or more sentences if splitting enhances readability.

\medskip
\noindent
\textbf{Ways to Simplify}:

\medskip
\noindent
A complex sentence can be simplified through:
\begin{itemize}[left=0pt]
    \item Paraphrasing: Replacing complex words with simpler synonyms.
    \item Condensing: Deleting unimportant information.
    \item Splitting: Dividing a long, complex sentence into several simpler sentences.
\end{itemize}

\textbf{Evaluation Criteria and Definitions}:
\begin{enumerate}[left=0pt]
    \item Simplicity
    \begin{itemize}[left=0pt]
        \item Definition: Simplicity measures how easy the simplified sentence is to understand, irrespective of the original sentence's complexity. This includes vocabulary, sentence structure, and clarity. Simplicity should not be assessed by comparing directly to the original but by how simple the sentence is in general. If the original is already very simple, an unchanged sentence might still score high.
        \item Guidelines for Simplicity:
        \begin{itemize}[left=0pt]
            \item Use simple words and common words (replace rare or long nouns with a synonym or split compound words using a hyphen, e.g. Atomabkommen -> Atom-Abkommen). Avoid difficult terms and explain if necessary.
            \item Use simple sentence structures with minimal subordinate clauses, coordinated clauses and genitive constructions.
            \item Avoid including information in parentheses (e.g. Angela Merkel (die ehemalige Bundeskanzlerin)) or in nested commas (e.g. Angela Merkel, die ehemalige Bundeskanzlerin,...).
            \item  Use active as opposed to passive voice.
            \item Use past perfect instead of past simple.
            \item Avoid metaphors, initials/abbreviations, special characters, percentages or big numbers.
            \item Replace numbers (vier) with digits (4).
            \item Avoid foreign terms unless they are very well-known.
            \item Use positive rather than negative sentences and direct speech with words like \enquote{du} or \enquote{Sie} where appropriate.
        \end{itemize}
        \item Grading Levels:
        \begin{itemize}[left=0pt]
            \item 100: Exceptionally simple, very easy to grasp; adheres strictly to simplicity guidelines.
            \item 75: Mostly simple with minor complexities, such as infrequent longer words.
            \item 50: Moderate simplicity; noticeable complexity in vocabulary or structure.
            \item 25: Similar complexity level to the original; minimal simplification.
            \item 0: Overly complex, difficult to understand.
        \end{itemize}
    \end{itemize}
    \item Meaning Preservation
    \begin{itemize}[left=0pt]
        \item Definition: Meaning preservation evaluates whether the simplified version genuinely conveys the core meaning of the original sentence without introducing unrelated information along with omitting less important details.
        \item Grading Levels:
        \begin{itemize}[left=0pt]
            \item 100: Fully retains the original meaning without omissions or unrelated additions.
            \item 75: Mostly preserves the message, with minor omissions.
            \item 50: Generally preserves meaning but with notable omissions or some unrelated information.
            \item 25: Loses basic meaning with significant omissions or many irrelevant details.
            \item 0: Fails to convey the original meaning accurately.
        \end{itemize}
    \end{itemize}
    \item Fluency
    \begin{itemize}[left=0pt]
        \item Definition: Fluency assesses the grammatical correctness and natural flow of the simplified sentence.
        \item Grading Levels:
        \begin{itemize}[left=0pt]
            \item 100: Fully grammatically correct and naturally flowing.
            \item 75: Mostly fluent with minor grammatical errors.
            \item 50: Moderately fluent with noticeable errors affecting flow.
            \item 25: Many grammatical mistakes, affecting readability.
            \item 0: Lacks fluency significantly; unreadable.
        \end{itemize}
    \end{itemize}
\end{enumerate}

\textbf{Evaluation Output Format}:
\begin{itemize}[left=0pt]
    \item Feedback: Describe the strengths and weaknesses of the simplified sentence in comparison to the original. Include specific examples to support the assessment. Score cannot be less than 0 or more than 100.
    \item Score:
    \begin{itemize}[left=0pt]
        \item Simplicity: \{X\}
        \item Meaning Preservation: \{Y\}
        \item Fluency: \{Z\}
    \end{itemize}
\end{itemize}

\textbf{Example 1}:

\medskip
\noindent
Complex: Coca-Cola bezeichnete die Auszeichnung hingegen als nicht gerechtfertigt und k\"undigte an, den Goldenen Windbeutel nicht anzunehmen.

\medskip
\noindent
Simplification: Coca-Cola will den Goldenen Windbeutel nicht annehmen.

\medskip
\noindent
Feedback:
\begin{itemize}[left=0pt]
    \item Simplicity: The simplification effectively reduces the complexity of the original sentence by eliminating intricate details and focusing on the core message. It replaces the more complex structure of the original with a straightforward statement. The removal of terms like ``bezeichnet'' and ``als nicht gerechtfertigt'' reduces complexity, aligning well with simplicity guidelines.
    \item Meaning Preservation: While the simplified sentence maintains the core meaning that Coca-Cola does not intend to accept the award (Goldenen Windbeutel), it omits some contextual details from the original. Specifically, the simplification does not convey that Coca-Cola also considered the award to be unjustified. This omission reduces the score for meaning preservation.
    \item Fluency: The simplified sentence is grammatically correct and flows naturally. It adheres to standard German syntax and structure, without any grammatical errors.
\end{itemize}

Score:
\begin{itemize}[left=0pt]
    \item Simplicity: 90
    \item Meaning Preservation: 50
    \item Fluency: 100
\end{itemize}

\medskip
\noindent
\textbf{Example 2}:

\medskip
\noindent
Complex: Der Lenker soll schwer verletzt und mit einem Notarzthubschrauber in ein Krankenhaus geflogen worden sein.

\medskip
\noindent
Simplification: Der Lkw-Fahrer wurde beim Unfall schwer verletzt und in ein Krankenhaus geflogen.

\medskip
\noindent
Feedback: 
\begin{itemize}[left=0pt]
    \item Simplicity: The simplified sentence replaces ``Lenker'' with ``Lkw-Fahrer'', making the subject more specific, which can either increase or decrease simplicity depending on the assumed knowledge of the reader. The sentence structure is straightforward, eliminates the passive voice and the use of a helicopter, opting for a more direct description. It scores well in simplicity by avoiding complex constructions and by using common vocabulary. 
    \item Meaning Preservation: The complex sentence mentions the use of an emergency doctor helicopter (``Notarzthubschrauber''), which is omitted in the simplified version. This is a minor yet specific detail that contributes to the original meaning. However, the critical information about an injury and hospital transport is preserved.
    \item Fluency: The simplified sentence is grammatically correct and maintains a natural flow, comparable to the original. It uses proper sentence structure, making it easy to read and understand.
\end{itemize}

Score:
\begin{itemize}[left=0pt]
    \item Simplicity: 85
    \item Meaning Preservation: 75
    \item Fluency: 95
\end{itemize}

\textbf{Example 3}:

\medskip
\noindent
Complex: Bei einem schweren Busungl\"uck auf der portugiesischen Urlaubsinsel Madeira sind 29 Menschen ums Leben gekommen.

\medskip
\noindent
Simplification: Am Mittwoch ist es auf der Urlaubsinsel Madeira in Portugal zu eine schweren Bus-Unfall gekommen. Dabei sinde 29 Menschen gestorben.

\medskip
\noindent
Feedback: The simplified sentence effectively uses two distinct sentences to convey the information, enhancing readability. The use of common terms like ``Unfall'' (accident) instead of ``Ungl\"uck'' (misfortune) aligns well with the simplicity criterion. However, there are some grammatical issues, such as ``sinde'' instead of ``sind'', which affects fluency.
\begin{itemize}[left=0pt]
    \item Simplicity: The simplified version achieves simplicity by splitting the original sentence and using straightforward language. The sentence structure is simple enough for easy comprehension. The simplification slightly modifies the phrasing but retains simple language, fitting well within the guidelines for Simplicity.
    \item Meaning Preservation: The simplified sentence effectively preserves the essential meaning of the original, including the location and the occurrence of a bus accident resulting in fatalities. However, the simplification introduces the detail ``Am Mittwoch'', which is not present in the original, which could affect meaning preservation slightly.
    \item Fluency: The grammatical errors, such as ``sinde'' instead of ``sind'', detract from the fluency. Additionally, ``zu eine'' should be ``zu einem'' for grammatical correctness in German. These errors can disrupt the natural flow and readability of the text.
\end{itemize}

Score:
\begin{itemize}[left=0pt]
    \item Simplicity: 80
    \item Meaning Preservation: 70
    \item Fluency: 50
\end{itemize}

\textbf{Example 4}:

\medskip
\noindent
Complex: Ein Lkw mit einem Anh\"nger ist Montagnachmittag in Munderfing im Bezirk Braunau im Innviertel mit einem Zug zusammengesto\ss en und hat in der Folge ein Haus schwer besch\"adigt.

\medskip
\noindent
Simplification: In Ober\"osterreich fuhr ein Zug in einen Lkw. Am Montag kam es im Ort Munderfing im Bundesland Ober\"osterreich zu einem Unfall. Anschlie\ss end krachte der Lkw in ein Haus. Der Fahrer wurde mit einem Hubschrauber in ein Krankenhaus gebrach.

\medskip
\noindent
Feedback:  The simplified version successfully breaks down the original sentence into several shorter sentences, enhancing readability. It effectively segregates the complex information into distinct parts, making it easier to follow along for the reader. However, the addition of new information, such as ``Der Fahrer wurde mit einem Hubschrauber in ein Krankenhaus gebracht'',  introduces details not present in the original, impacting meaning preservation. 
\begin{itemize}[left=0pt]
    \item Simplicity: The approach of splitting the original sentence into smaller sentences significantly aids understanding, and the use of straightforward language makes it simple. The sentences are straightforward and devoid of complex structures. The place ``Ober\"osterreich'' is used more than once, and the phrase ``kam es zu einem Unfall'' could be simplified further. Nevertheless, the overall simplification is good.
    \item Meaning Preservation: While the main events are preserved, there is a newly introduced detail about the driver, which was not mentioned in the original sentence, and it can mislead about the severity or outcome of the collision. This reduces the score for meaning preservation.
    \item Fluency: The sentences mostly follow grammar rules well and maintain a natural flow. There are minor issues, such as redundancy, which slightly affect overall fluency. Still, the content largely maintains a logical progression.
\end{itemize}

Score:
\begin{itemize}[left=0pt]
    \item Simplicity: 85
    \item Meaning Preservation: 50
    \item Fluency: 80
\end{itemize}

\textbf{Example 5}:

\medskip
\noindent
Complex: Damit sind seit Bekanntwerden des Erregers Sars-CoV-2 Ende vergangenen Jahres weltweit mehr als 28,6 Millionen Infektionen gemeldet worden.

\medskip
\noindent
Simplification: Des Corona-Virus seit Ausbruch gab es mehr weltweit als 28,6 Millionen Infektionen.

\medskip
\noindent
Feedback: The simplified sentence ``Des Corona-Virus seit Ausbruch gab es mehr weltweit als 28,6 Millionen Infektionen'' attempts to condense the original information. However, there are several issues in terms of simplicity, meaning preservation, and fluency.
\begin{itemize}[left=0pt]
    \item Simplicity: The simplified sentence is slightly simpler in using common words. It effectively reduces the complex expression ``Erregers Sars-CoV-2'' to ``Corona-Virus''. However, the phrase ``Des Corona-Virus seit Ausbruch'' is awkward and can confuse readers due to its incorrect grammatical construction. Additionally, the ordering and structuring make it harder to read than necessary.
    \item Meaning Preservation: The simplification loses critical context from the original sentence, such as the timeframe ``Ende vergangenen Jahres'' which sets a clear historical context for when the infections began reporting. Without this piece of information, the reader may not have a clear understanding of the full context.
    \item Fluency: The sentence lacks correct grammatical structure, especially evident in ``Des Corona-Virus seit Ausbruch'', which suffers from poor syntax. This grammatical flaw detracts from the overall fluency and makes the sentence less readable. The use of ``mehr weltweit als'' is also awkward and could be expressed more naturally to improve the flow.
\end{itemize}

Score:
\begin{itemize}[left=0pt]
    \item Simplicity: 50
    \item Meaning Preservation: 25
    \item Fluency: 25
\end{itemize}

\textbf{Example 6}:

\medskip
\noindent
Complex: Weltweit wird laut Studie immer mehr Alkohol getrunken.

\medskip
\noindent
Simplification: Laut einer Studie wird auf der Welt immer mehr Alkohol getrunken.

\medskip
\noindent
Feedback: The simplified sentence, ``Laut einer Studie wird auf der Welt immer mehr Alkohol getrunken,'' demonstrates some simplification, although the original sentence was already relatively straightforward. It remains clear and easy to understand, employing a similar sentence structure but with slight changes, like switching ``weltweit'' to ``auf der Welt,'' which is a more common phrase and may be slightly easier for some readers to parse.
\begin{itemize}[left=0pt]
    \item Simplicity: The simplification retains the straightforwardness of the original sentence but doesn't significantly increase simplicity since both sentences use common vocabulary and simple structures. There are minor changes that enhance readability, but the overall complexity remains similar.
    \item Meaning Preservation: The meaning is entirely preserved in the simplified version. Both sentences convey the same information regarding the increase in alcohol consumption worldwide. There are no omissions or additions that alter the original meaning.
    \item Fluency: The sentence is grammatically correct and flows naturally, similar to the original. There are no errors affecting readability or comprehension, and it reads smoothly in German.
\end{itemize}

Score:
\begin{itemize}[left=0pt]
    \item Simplicity: 85
    \item Meaning Preservation: 100
    \item Fluency: 100
\end{itemize}

\medskip
\noindent

Now grade the following simplification:

\medskip
\noindent
Complex: \texttt{\{complex\_sentence\}}

\medskip
\noindent
Simplification: \texttt{\{simplified\_sentence\}}

\end{tcolorbox}



\section{\textit{LLM-Judge} Selection Analysis}
\label{appendix:model-selection}

\subsection{Questions and Hypothesis}
We choose the Deepseek-V3 \citep{liu2024deepseekv3} herd of distilled models and Zephyr \citep{tunstall2024zephyr} as the backbone of \textit{LLM-Judge}. We concentrate on studying the following two questions:
\begin{itemize}[left=0pt]
    \item \textbf{Model scalability:} Do smaller models, such as Distill-Qwen-1.5B or Distill-Qwen-7B, perform similarly to the larger model Distill-Qwen-14B? Answering this question ensures that the smaller models can be used to reliably generate quality scores for model training.
    \item \textbf{Inter-model performance:} To what extent do scoring behaviors differ between different model families of a similar size, e.g., Distill-Qwen, Distill-Llama and Zephyr (i.e., a Mistral variant)? This is done to validate that the three selected models exhibit similarly high grading quality.
\end{itemize}
To investigate the above two questions, we selected the following open-source LLMs for experiments:


\begin{itemize}[left=0pt]
    \item \textbf{Distill-Qwen-1.5B}: \huggingface \href{https://huggingface.co/deepseek-ai/DeepSeek-R1-Distill-Qwen-1.5B}{\nolinkurl{deepseek-ai/DeepSeek-R1-Distill-Qwen-1.5B}}
    \item \textbf{Distill-Qwen-7B}: \huggingface \href{https://huggingface.co/deepseek-ai/DeepSeek-R1-Distill-Qwen-7B}{\nolinkurl{deepseek-ai/DeepSeek-R1-Distill-Qwen-7B}}
    \item \textbf{Distill-Llama-8B}: \huggingface \href{https://huggingface.co/deepseek-ai/DeepSeek-R1-Distill-Llama-8B}{\nolinkurl{deepseek-ai/DeepSeek-R1-Distill-Llama-8B}} 
    \item \textbf{Zephyr-7B}: \huggingface \href{https://huggingface.co/HuggingFaceH4/zephyr-7b-beta}{\nolinkurl{HuggingFaceH4/zephyr-7b-beta}}
\end{itemize}

Each model was used to generate \textit{fluency}, \textit{meaning preservation}, and \textit{simplicity} scores using the \textit{Prompt-Final} in ten sampling runs per instance for the full training set. The resulting scores were then analyzed in two ways: by examining the variance in scores across the ten runs for the same instance (sampling variability), and by evaluating the distribution of the averaged scores across the ten runs (score diversity). 

Three hypotheses were tested as a measure of model quality:
\begin{itemize}[left=0pt]
    \item \textbf{Hypothesis 1 (Score diversity):} Higher-performing models will exhibit greater standard deviation in their \textit{average scores} due to an increased ability to discriminate between low-quality and high-quality texts.
     \item \textbf{Hypothesis 2 (Sampling stability):} Higher-performing models will exhibit lower \textit{per-row sampling variance}, i.e., the standard deviation of scores across the ten runs of the same data row, due to improved consistency and output stability.
    \item \textbf{Hypothesis 3 (Score correlation):} If two models are similarly capable, their scoring behavior should be positively correlated.

\end{itemize}



\subsection{Experimental Findings}

\paragraph{1. Scalability Across Model Sizes.}

\begin{itemize}[left=0pt]
\item \textit{\textbf{Hypothesis 1: Increased Model Size Increases Scoring Variability.}} 

\begin{table}[H]
\footnotesize
\caption[Standard deviation of average scores across Distill-Qwen model sizes by criterion]{Comparison of the standard deviation of average scores for Distill-Qwen-1.5B, Distill-Qwen-7B, and Distill-Qwen-14B, grouped by evaluation criterion and model pair. The $p$-values from Levene's test indicate the statistical significance of variance differences between model pairs.}
\label{qwen_avg_std}
\resizebox{\columnwidth}{!}{%
\begin{tabular}{lcc}
\toprule
\textbf{Criterion} & \textbf{Standard Deviation} & \textbf{$p$-value} \\
\hline
\rowcolor{gray!20!white}
\multicolumn{3}{c}{\textit{Distill-Qwen-1.5B vs Distill-Qwen-7B}} \\
\textit{Simplicity} & 4.47 vs 9.22 & 0.0006 \\
\textit{Meaning P.} & 10.18 vs 16.98 & $<$0.0001 \\
\textit{Fluency} & 6.60 vs 7.55 & 0.0220 \\
\hline
\rowcolor{gray!20!white}
\multicolumn{3}{c}{\textit{Distill-Qwen-1.5B vs Distill-Qwen-14B}} \\
\textit{Simplicity} & 4.47 vs 13.60 & $<$0.0001 \\
\textit{Meaning P.} & 10.18 vs 25.12 & $<$0.0001 \\
\textit{Fluency} & 6.60 vs 12.74 & \textit{0.1604} \\
\hline
\rowcolor{gray!20!white}
\multicolumn{3}{c}{\textit{Distill-Qwen-7B vs Distill-Qwen-14B}} \\
\textit{Simplicity} & 9.22 vs 13.60 & $<$0.0001 \\
\textit{Meaning P.} & 16.98 vs 25.12 & $<$0.0001 \\
\textit{Fluency} & 7.55 vs 12.74 & 0.0075 \\
\bottomrule
\end{tabular}
}
\centering
\end{table}

As shown in Figure~\ref{fig:qwen_histograms}, the distributions of ten-run average scores for Distill-Qwen-1.5B, Qwen-7B, and Distill-Qwen-14B indicate that larger models produce greater variability in \textit{simplicity} and \textit{meaning preservation}, while trends for \textit{fluency} are less consistent (see Table~\ref{qwen_avg_std}). For \textit{simplicity}, the median remains stable across models, but the minimum scores drop sharply (50 to 14, $p < 0.0001$), driving standard deviations from 4.47 to 13.60 ($p < 0.001$). For \textit{meaning preservation}, distributions shift from near-Gaussian to bimodal, with extreme values (0 and 100) increasingly frequent in larger models; standard deviations rise significantly (10.18 to 25.11, $p < 0.0001$). For \textit{fluency}, although standard deviations also increase (6.6 to 12.7, no significance), inter-quartile ranges shrink and Distill-Qwen-14B predicts 100 in over half of cases, suggesting limited effective diversity. Overall, Hypothesis~1 is well supported for \textit{simplicity} and \textit{meaning preservation}, but evidence for \textit{fluency} remains inconclusive.

\item{\textit{\textbf{Hypothesis 2: Higher-Performing Models Exhibit Lower Sampling Variability.}}} 

Table~\ref{qwen-variances} reports the average standard deviation of scores across ten sampled outputs per input for different Distill-Qwen sizes. Overall, variance decreases as model size increases, supporting Hypothesis~2. The main exception is \textit{simplicity}, where the standard deviation rises significantly from 6.5 to 8.2 ($p = 0.0001$) between Distill-Qwen-7B and Distill-Qwen-14B. For all other criteria, standard deviations decrease consistently, with the largest reduction occurring between Distill-Qwen-1.5B and Distill-Qwen-7B.

\begin{table}[H]
\centering
\footnotesize
\caption[Per-row standard deviations for Distill-Qwen-1.5B, Distill-Qwen-7B, and Distill-Qwen-14B]{Comparison of the mean per-row output standard deviation for Distill-Qwen-1.5B, Distill-Qwen-7B, and Distill-Qwen-14B, grouped by evaluation criterion and model pair. $p$-values from independent \textit{t}-tests indicate the statistical significance of differences in mean sampling variance between models.}
\label{qwen-variances}
\resizebox{\columnwidth}{!}{%
\begin{tabular}{lcc}
\toprule
\textbf{Criterion} & \textbf{Mean Std. Deviation} & \textbf{$p$-value} \\
\hline
\rowcolor{gray!20!white}
\multicolumn{3}{c}{\textit{Distill-Qwen-1.5B vs Distill-Qwen-7B}} \\
\textit{Simplicity} & 9.006 vs 6.507 & 0.0000 \\
\textit{Meaning P.} & 15.734 vs 10.047 & 0.0000 \\
\textit{Fluency} & 11.456 vs 6.198 & 0.0000 \\
\hline
\rowcolor{gray!20!white}
\multicolumn{3}{c}{\textit{Distill-Qwen-1.5B vs Distill-Qwen-14B}} \\
\textit{Simplicity} & 9.006 vs 8.203 & \textit{0.0899} \\
\textit{Meaning P.} & 15.734 vs 7.392 & 0.0000 \\
\textit{Fluency} & 11.456 vs 4.456 & 0.0000 \\
\hline
\rowcolor{gray!20!white}
\multicolumn{3}{c}{\textit{Distill-Qwen-7B vs Distill-Qwen-14B}} \\
\textit{Simplicity} & 6.507 vs 8.203 & 0.0001 \\
\textit{Meaning P.} & 10.047 vs 7.392 & 0.0000 \\
\textit{Fluency} & 6.198 vs 4.456 & 0.0000 \\
\bottomrule
\end{tabular}
}
\end{table}

\begin{figure*}[hbt!]
    \centering
    \textbf{i) Distill-Qwen-1.5B}\\[0.5em]
    \begin{subfigure}[b]{0.32\textwidth}
        \includegraphics[width=\textwidth, height = 4cm]{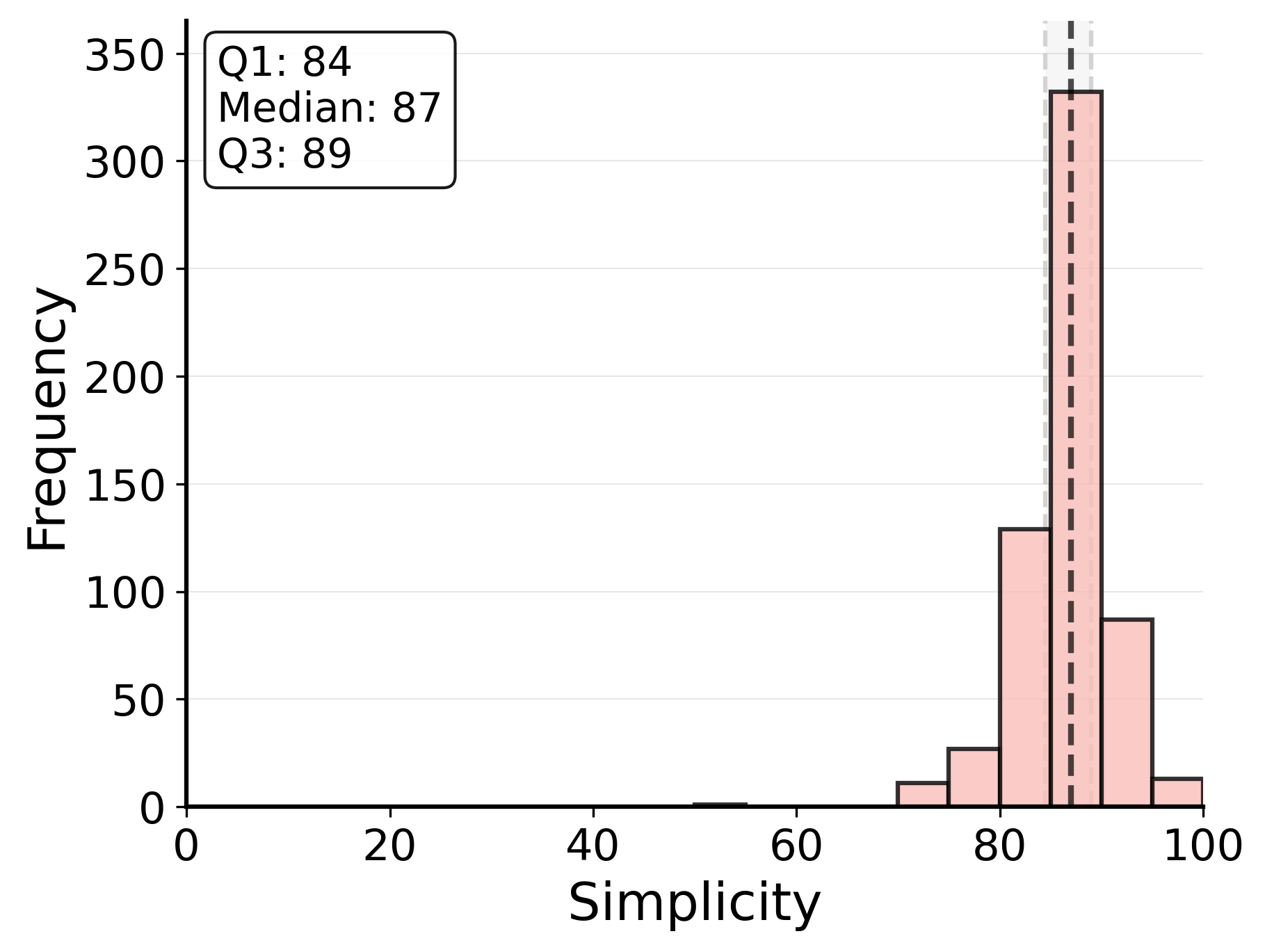}
        \caption{Simplicity}
    \end{subfigure}
    \begin{subfigure}[b]{0.32\textwidth}
        \includegraphics[width=\textwidth, height = 4cm]{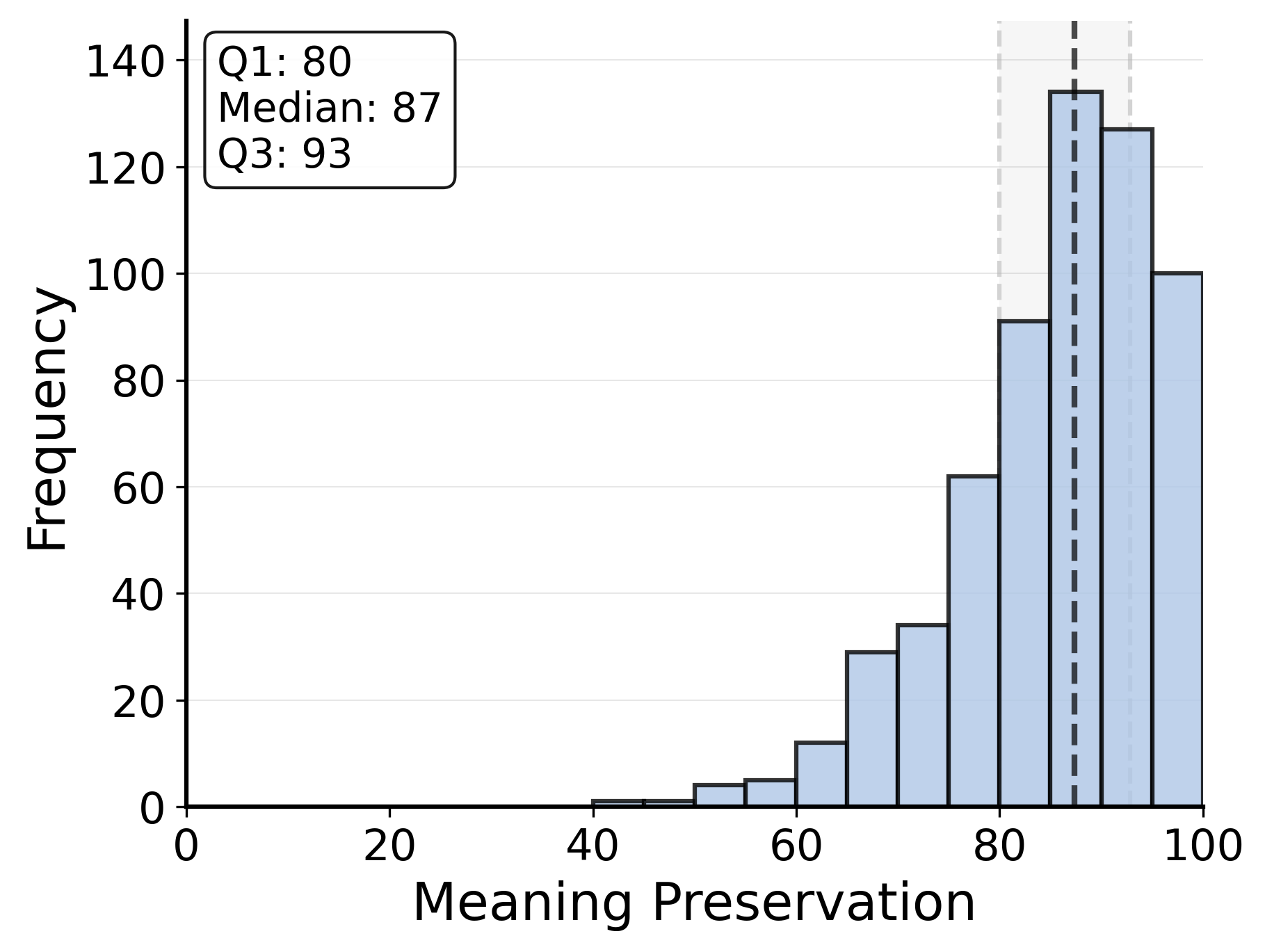}
        \caption{Meaning Preservation}
    \end{subfigure}
    \begin{subfigure}[b]{0.32\textwidth}
        \includegraphics[width=\textwidth, height = 4cm]{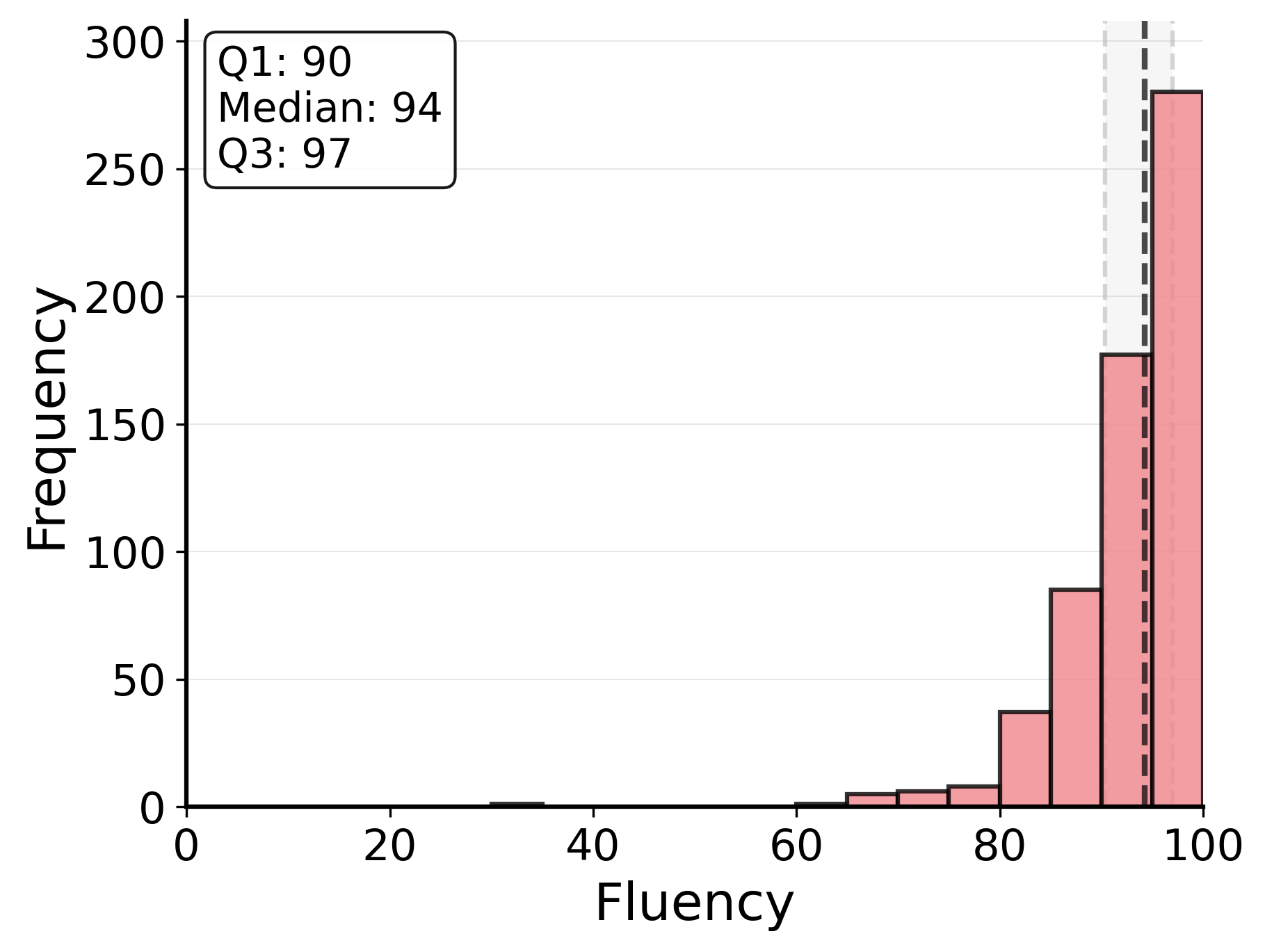}
        \caption{Fluency}
    \end{subfigure}
    
    \vspace{1em}
    \textbf{ii) Distill-Qwen-7B}\\[0.5em]
        \begin{subfigure}[b]{0.32\textwidth}
        \includegraphics[width=\textwidth, height = 4cm]{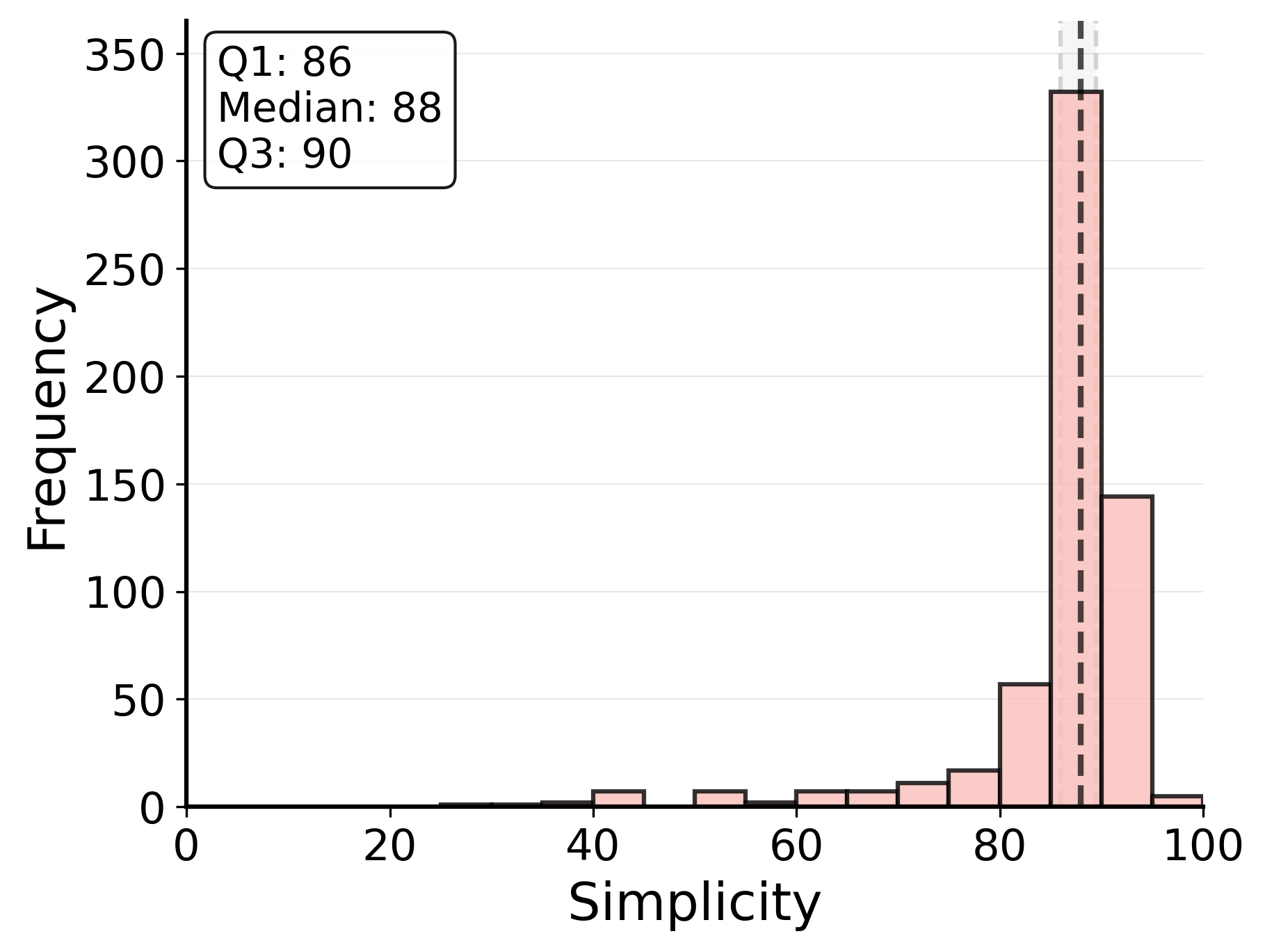}
        \caption{Simplicity}
    \end{subfigure}
    \begin{subfigure}[b]{0.32\textwidth}
        \includegraphics[width=\textwidth, height = 4cm]{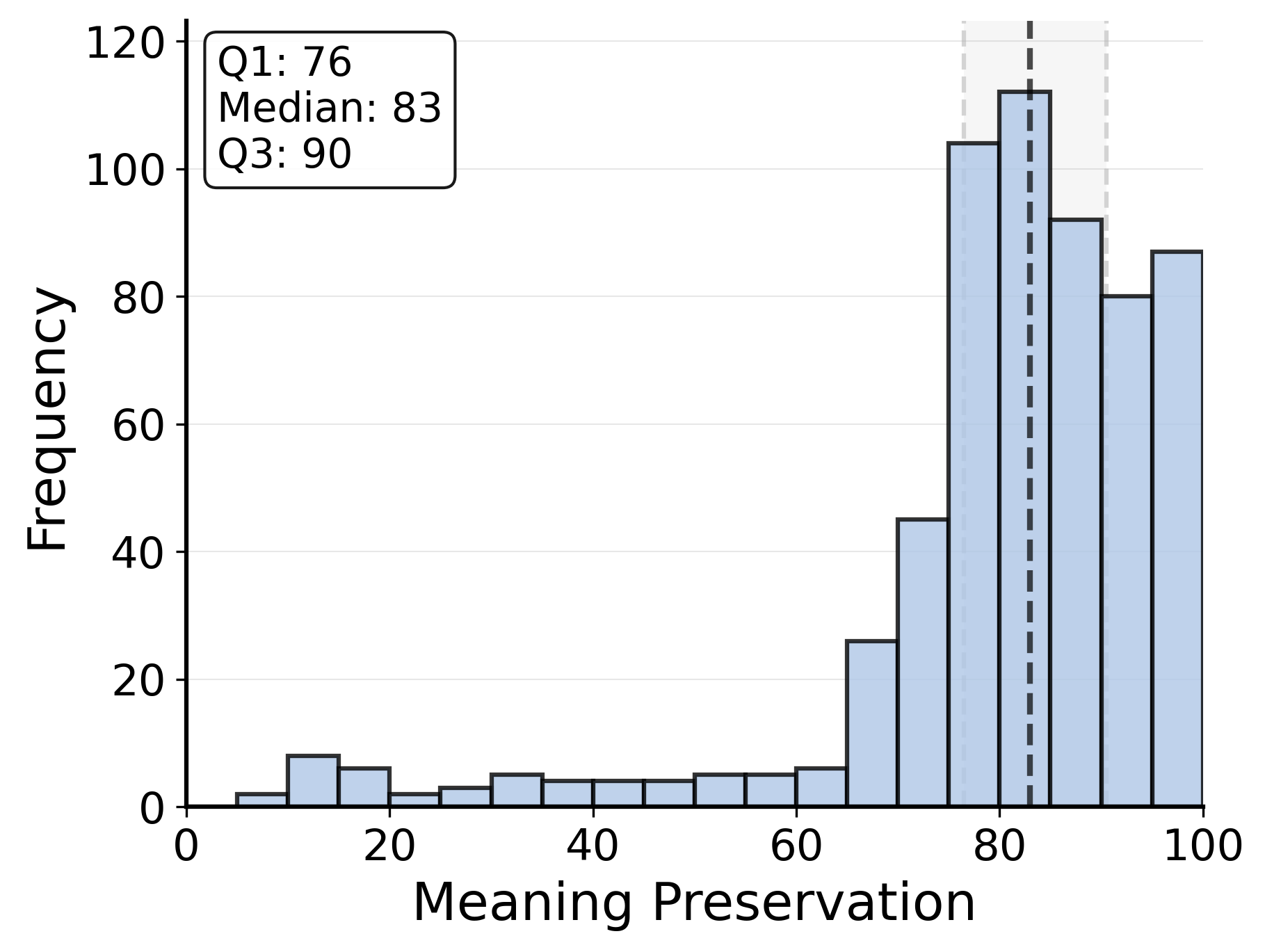}
        \caption{Meaning Preservation}
    \end{subfigure}
    \begin{subfigure}[b]{0.32\textwidth}
        \includegraphics[width=\textwidth, height = 4cm]{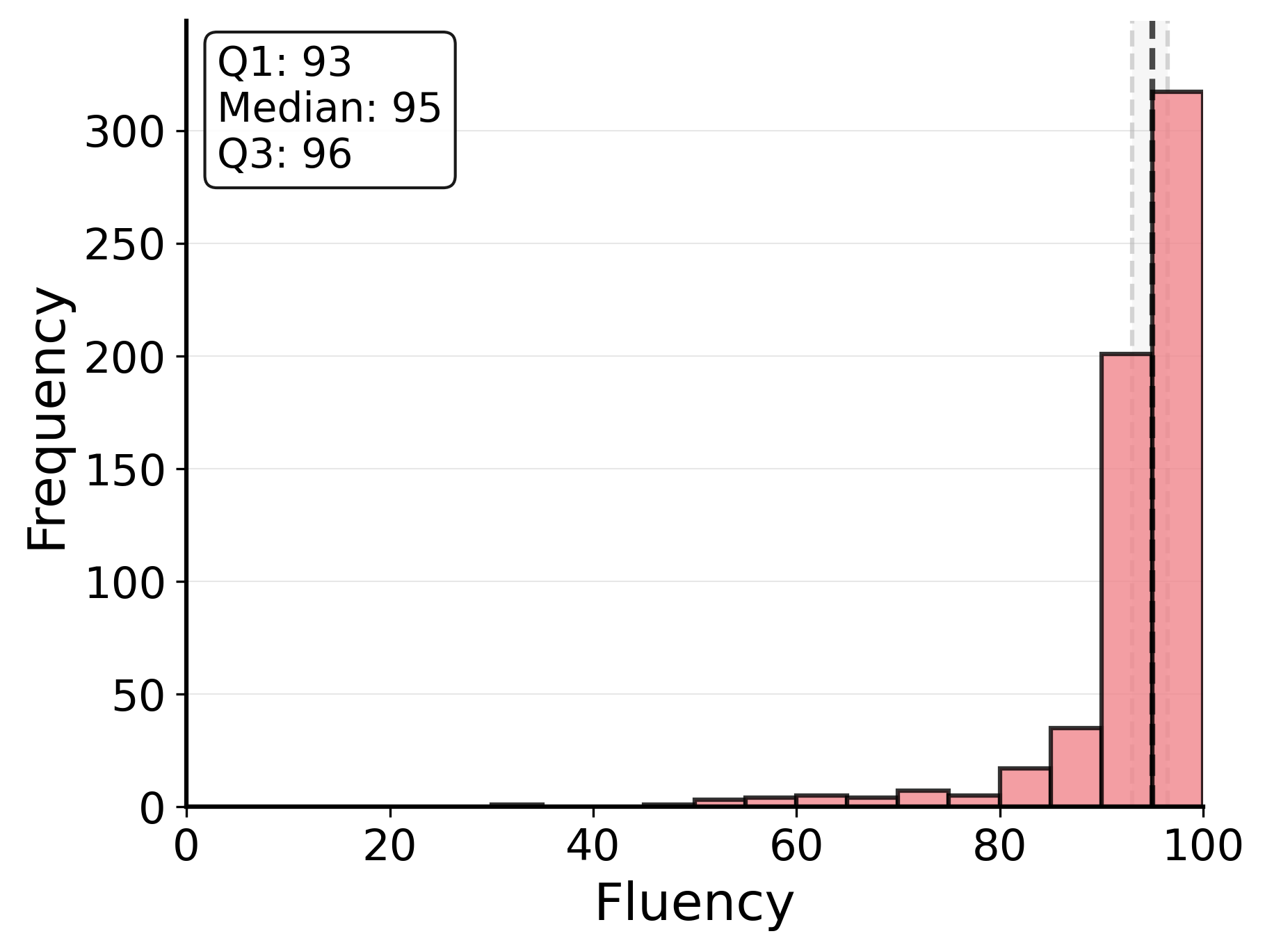}
        \caption{Fluency}
    \end{subfigure}
    
    \vspace{1em}
    \textbf{iii) Distill-Qwen-14B}\\[0.5em]
            \begin{subfigure}[b]{0.32\textwidth}
        \includegraphics[width=\textwidth, height = 4cm]{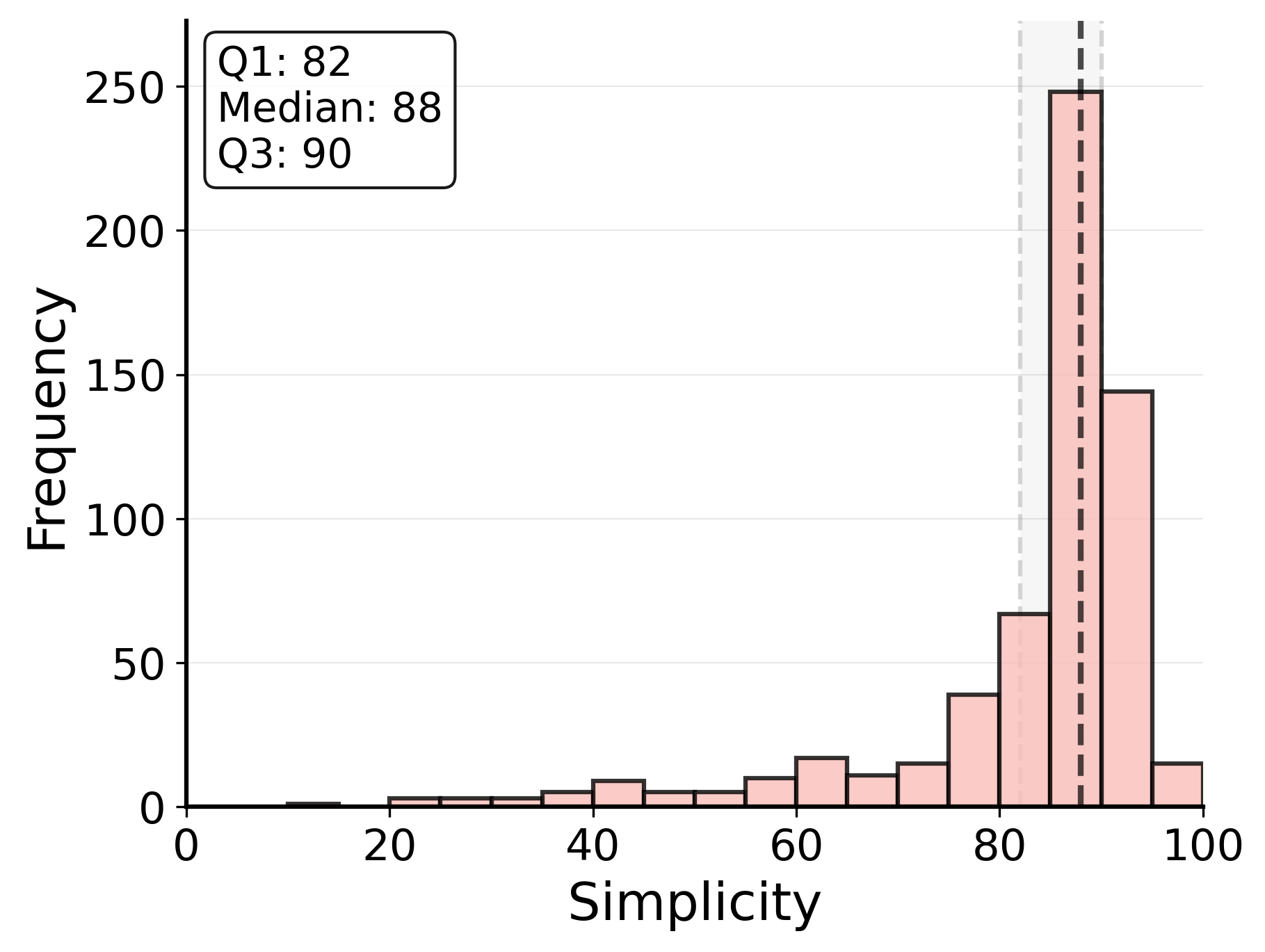}
        \caption{Simplicity}
    \end{subfigure}
    \begin{subfigure}[b]{0.32\textwidth}
        \includegraphics[width=\textwidth, height = 4cm]{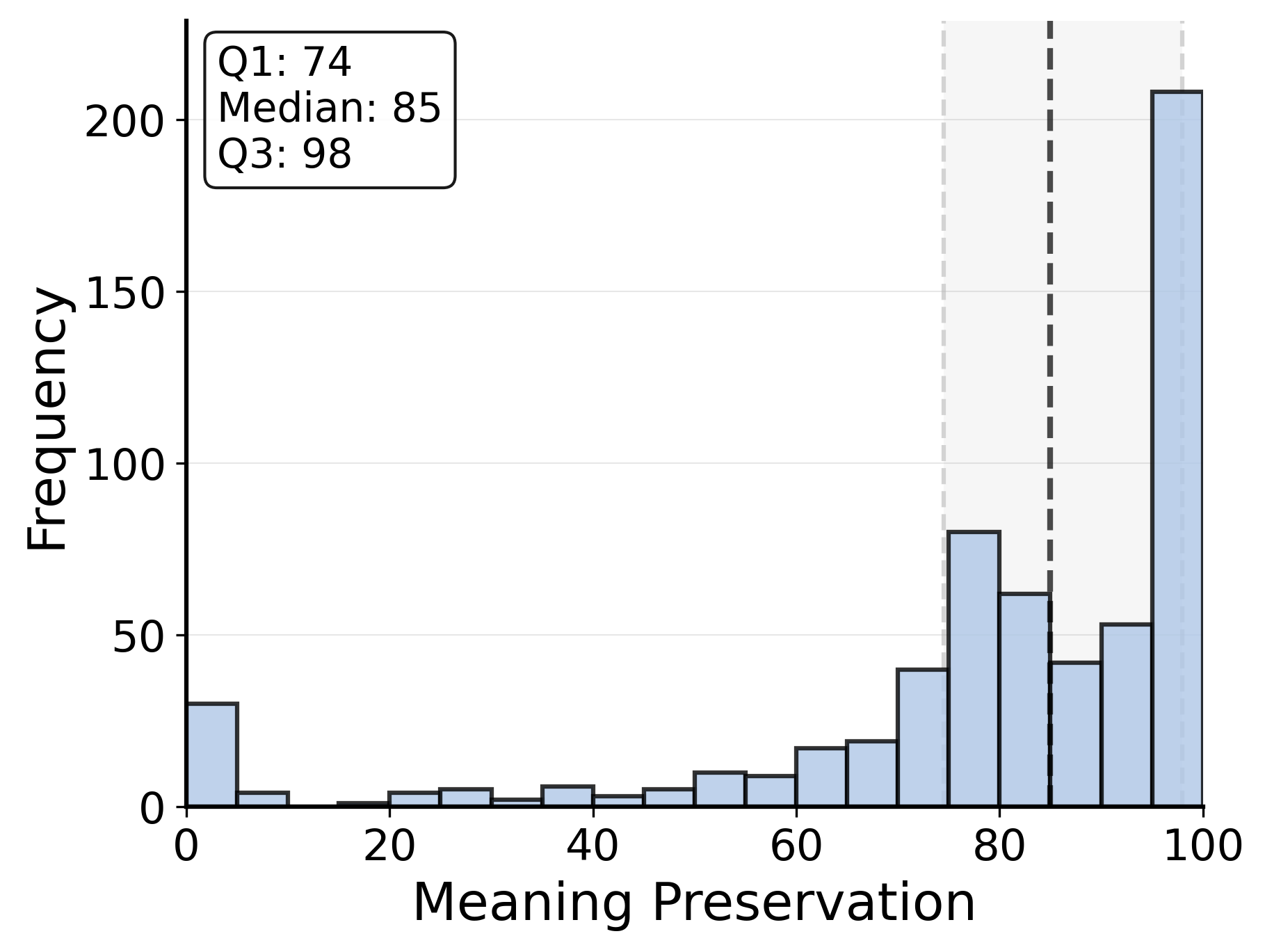}
        \caption{Meaning Preservation}
    \end{subfigure}
    \begin{subfigure}[b]{0.32\textwidth}
        \includegraphics[width=\textwidth, height = 4cm]{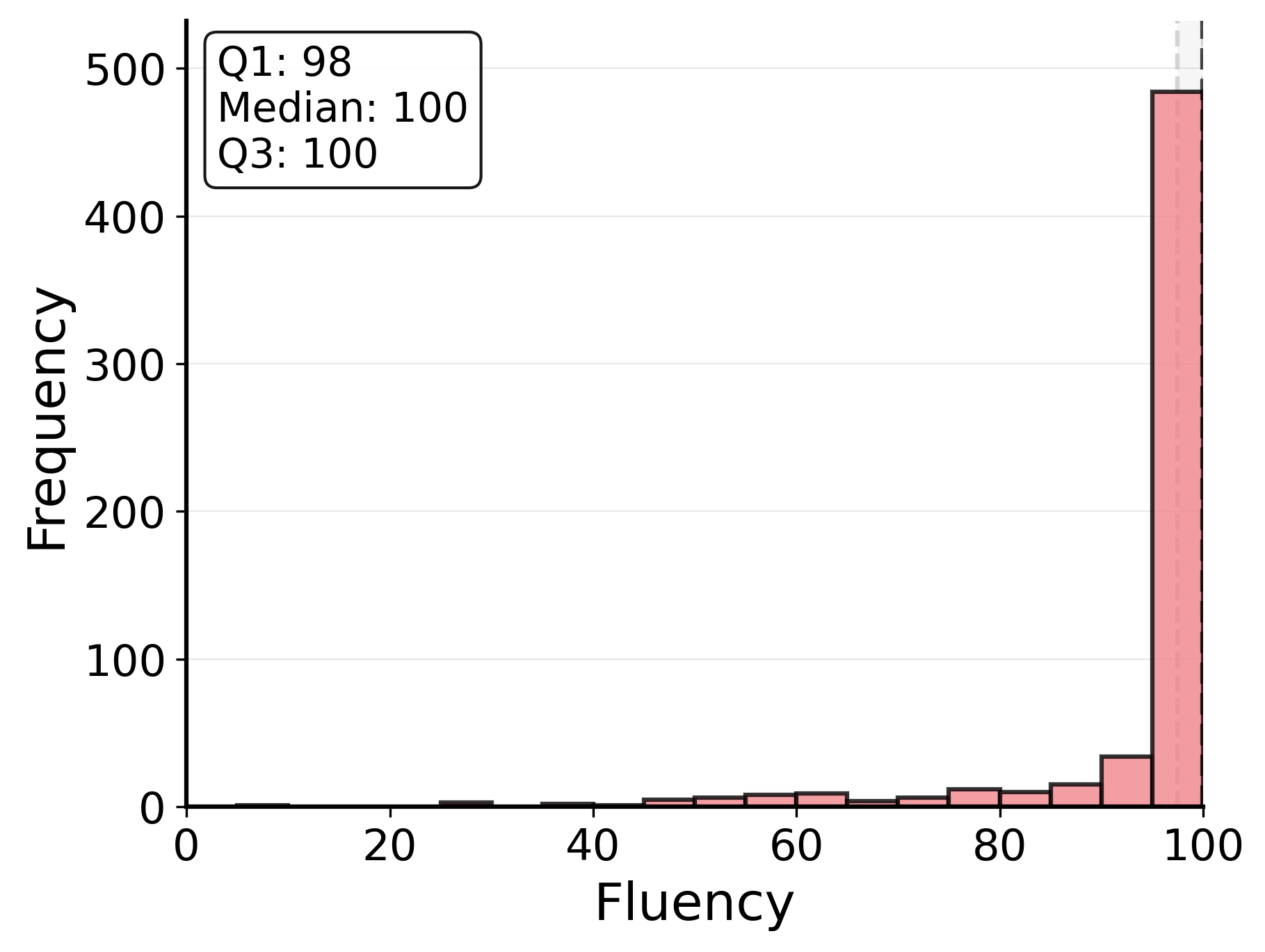}
        \caption{Fluency}
    \end{subfigure}
    \caption[Score distribution of different Qwen Modle Sizes]{Histograms of score distributions for i) Distill-Qwen-1.5B, ii) Distill-Qwen-7B, and iii) Distill-Qwen-14B, broken down by scoring criterion. The gray band indicates the IQR, while the black vertical dashed lines show the minimum, median, and maximum values. These values are also listed in the box in the upper left corner of each plot.}
    \label{fig:qwen_histograms}
\end{figure*}

\item \textit{\textbf{Hypothesis 3: Models of Increasing Size Exhibit Positive Correlation.}}

Table \ref{tab:correla_qwen} shows pairwise correlations between Distill-Qwen models by criterion, with all values significant at $p < 0.0001$. The shading highlights cases where one model predicts below 25 while the other does not. Distill-Qwen-1.5B and 7B correlate moderately (Pearson $r=0.39$ to 0.64), though \textit{meaning preservation} exhibits many cases where 7B assigns low scores that 1.5B does not. Correlations with Distill-Qwen-14B drop sharply for 1.5B (Pearson $r=0.23$ to 0.60), again reflecting more extreme low scores from the larger model. By contrast, 7B and 14B align more strongly (Pearson $r=0.51$ to 0.87), though 14B still assigns unusually low values for \textit{meaning preservation} and, to a lesser degree, \textit{fluency}. Agreement in normalized $z$-scores confirms this pattern: Krippendorff's $\alpha$ reaches 0.70/0.87/0.51 for the 7B-14B pair, compared to 0.30/0.60/0.23 for 1.5B-14B. Overall, Qwen-7B and 14B show the highest consistency across all criteria.

\end{itemize}

\begin{table}[H]
  \centering
  \caption{Pearson ($r$) and Spearman ($\rho$) correlations for (Distill-Qwen-)1.5B, (Distill-Qwen-)7B and (Distill-Qwen-)14B scores by criterion, all $p$-values are 0.}
  \label{tab:correla_qwen}
  \resizebox{\columnwidth}{!}{%
  \begin{tabular}{l@{\hskip 2pt}cccc}
    \toprule
    \multirow{2.5}{*}{\textbf{Model Comparison}} & & \multicolumn{3}{c}{\textbf{Criterion ($r$ / $\rho$)}} \\
    \cmidrule(lr){3-5}
     & & \textbf{\textit{Simplicity}} & \textbf{\textit{Meaning P.}} & \textbf{\textit{Fluency}} \\
    \midrule
    1.5B v. 7B   & & 0.39 / 0.32  & 0.64  / 0.67  & 0.42 / 0.38  \\
    1.5B v. 14B  & & 0.30  / 0.21  & 0.60  / 0.60  & 0.23  / 0.27  \\
    7B   v. 14B   & & 0.70  / 0.45  & 0.87  / 0.79  & 0.51  / 0.45  \\
    \bottomrule
  \end{tabular}
  }
\end{table}

\paragraph{2. Inter-model Comparison of Similar Sizes}

\begin{itemize}[left=0pt]
\item \textit{\textbf{Hypothesis 1: Models Differ in the Variance and Distribution of Average Scores.}}

\begin{table}[!htb]
\centering
\caption[Variance of average scores between Distill-Qwen, Distill-Llama and Zephyr by criterion]{Comparison of the standard deviation of average scores for Distill-Qwen-7B, Distill-Llama-8B, and Zephyr-7B, grouped by evaluation criterion and model pair. The $p$-values from Levene's test indicate the statistical significance of variance differences between model pairs.}
\label{tab:deviation_models}
\resizebox{\columnwidth}{!}{%
\begin{tabular}{lcc}
\toprule
\textbf{Criterion} & \textbf{Standard Deviation} & \textbf{$p$-value} \\
\hline
\rowcolor{gray!20!white}
\multicolumn{3}{c}{\textit{Distill-Qwen-7B vs Distill-Llama-8B}} \\
\textit{Simplicity} & 9.22 vs 10.95 & \textit{0.371} \\
\textit{Meaning P.} & 16.98 vs 19.27 & \textit{0.257} \\
\textit{Fluency} & 7.55 vs 9.18 & \textit{0.097} \\
\hline
\rowcolor{gray!20!white}
\multicolumn{3}{c}{\textit{Distill-Qwen-7B vs Zephyr-7B}} \\
\textit{Simplicity} & 9.22 vs 6.04 & $<$0.0001 \\
\textit{Meaning P.} & 16.98 vs 12.21 & $<$0.0001 \\
\textit{Fluency} & 7.55 vs 5.95 & $<$0.0001 \\
\hline
\rowcolor{gray!20!white}
\multicolumn{3}{c}{\textit{Distill-Llama-8B vs Zephyr-7B}} \\
\textit{Simplicity} & 10.95 vs 6.04 & $<$0.0001 \\
\textit{Meaning P.} & 19.27 vs 12.21 & $<$0.0001 \\
\textit{Fluency} & 9.18 vs 5.95 & $<$0.0001 \\
\bottomrule
\end{tabular}
}
\end{table}

\begin{figure*}[htb!]
    \centering
    \textbf{i) Distill-Qwen-7B}\\[0.5em]
    \begin{subfigure}[b]{0.325\textwidth}
        \includegraphics[width=\textwidth, height=4cm]{images/histogram_iqr_Average_Scores_across_three_graders_simplicity_avg_qwen_7_refined.png}
        \caption{Simplicity}
    \end{subfigure}
    \begin{subfigure}[b]{0.325\textwidth}
    \includegraphics[width=\textwidth,height=4cm]{images/histogram_iqr_Average_Scores_across_three_graders_meaning_avg_qwen_7_refined.png}
        \caption{Meaning Preservation}
    \end{subfigure}
    \begin{subfigure}[b]{0.325\textwidth}
        \includegraphics[width=\textwidth, height=4cm]{images/histogram_iqr_Average_Scores_across_three_graders_fluency_avg_qwen_7_refined.png}
        \caption{Fluency}
    \end{subfigure}
    
    \vspace{1em}
    \textbf{ii) Distill-Llama-8B}\\[0.5em]
        \begin{subfigure}[b]{0.325\textwidth}
        \includegraphics[width=\textwidth, height=4cm]{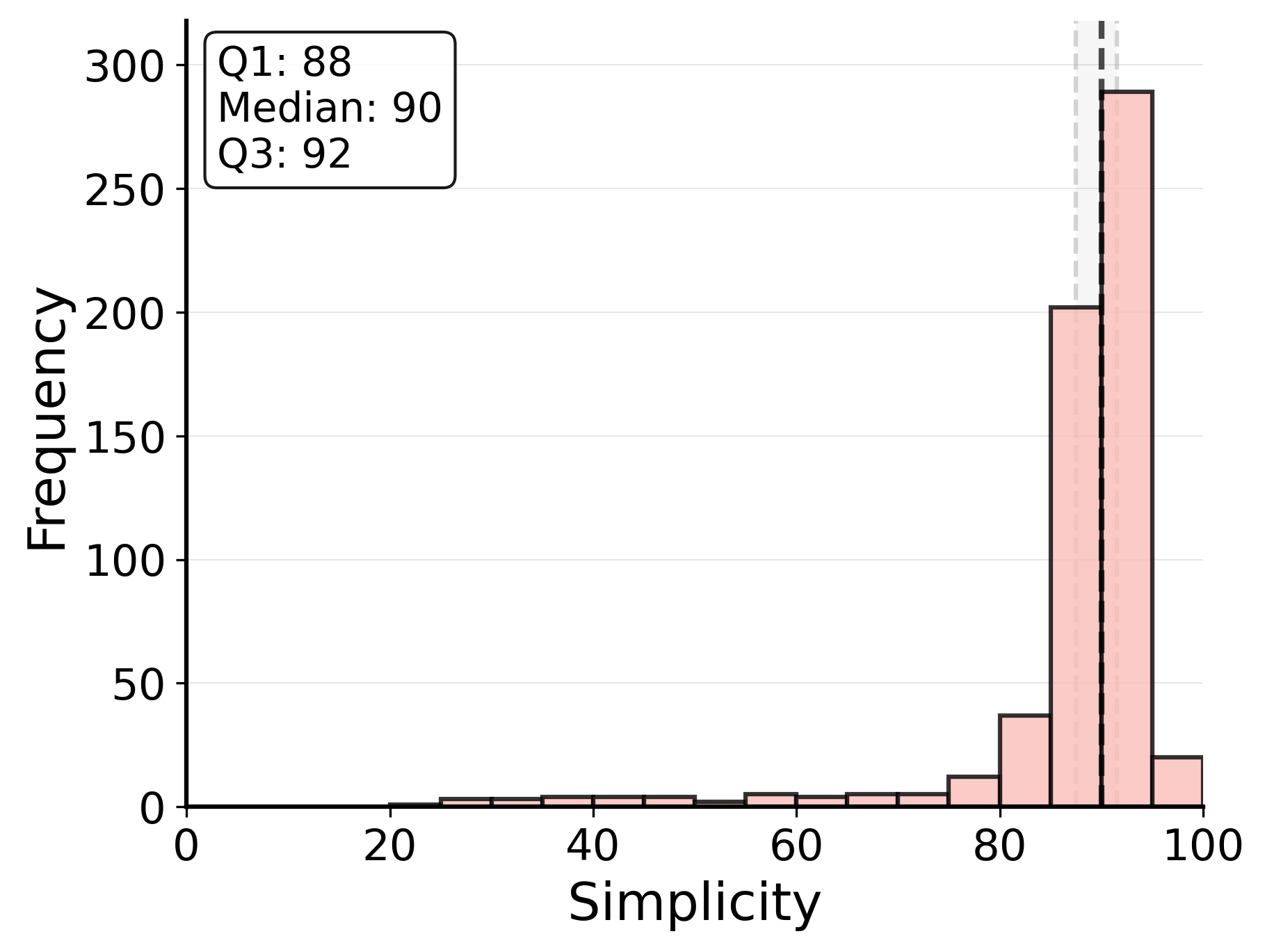}
        \caption{Simplicity}
    \end{subfigure}
    \begin{subfigure}[b]{0.325\textwidth}
        \includegraphics[width=\textwidth, height=4cm]{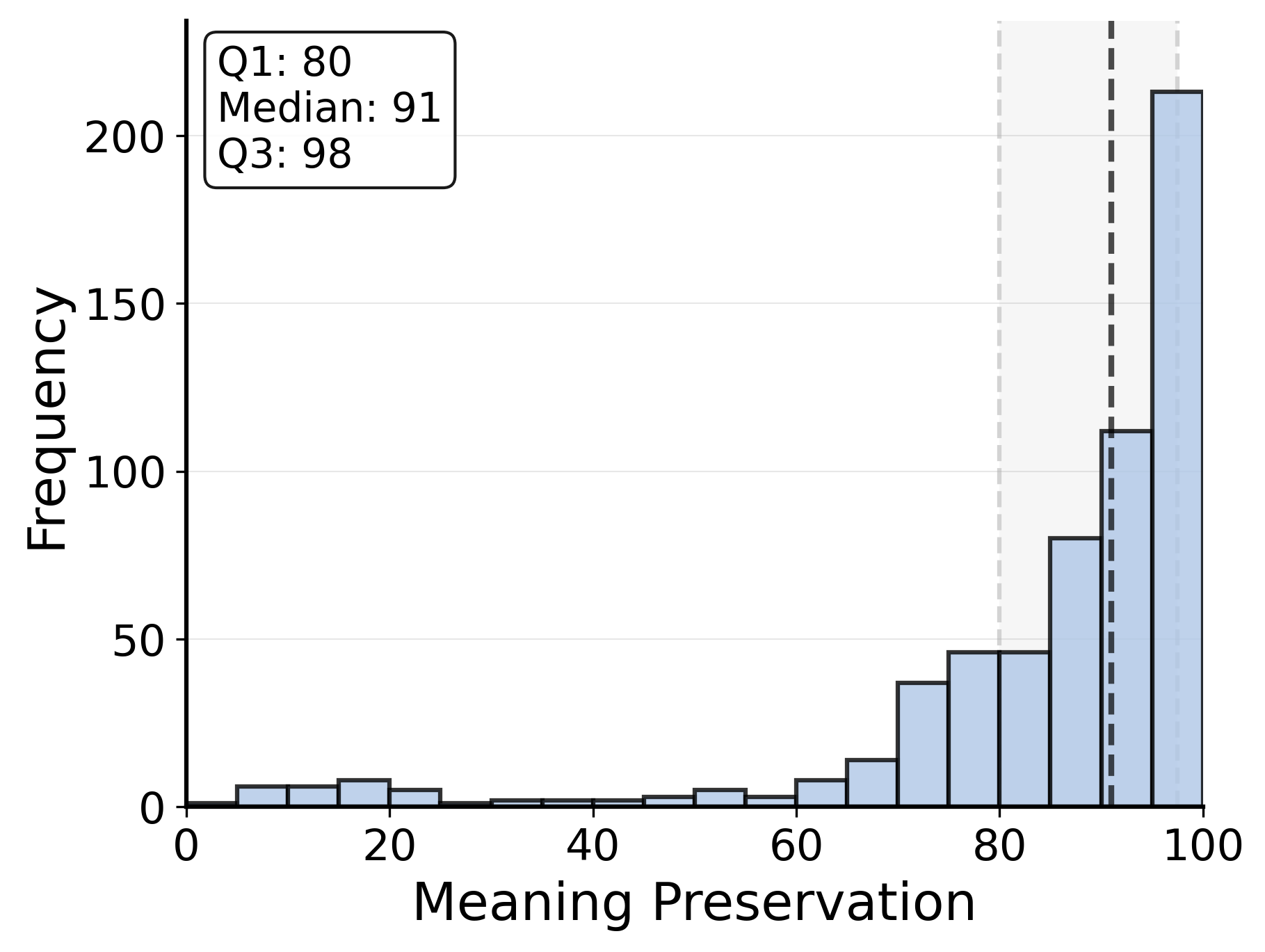}
        \caption{Meaning Preservation}
    \end{subfigure}
    \begin{subfigure}[b]{0.325\textwidth}
        \includegraphics[width=\textwidth, height=4cm]{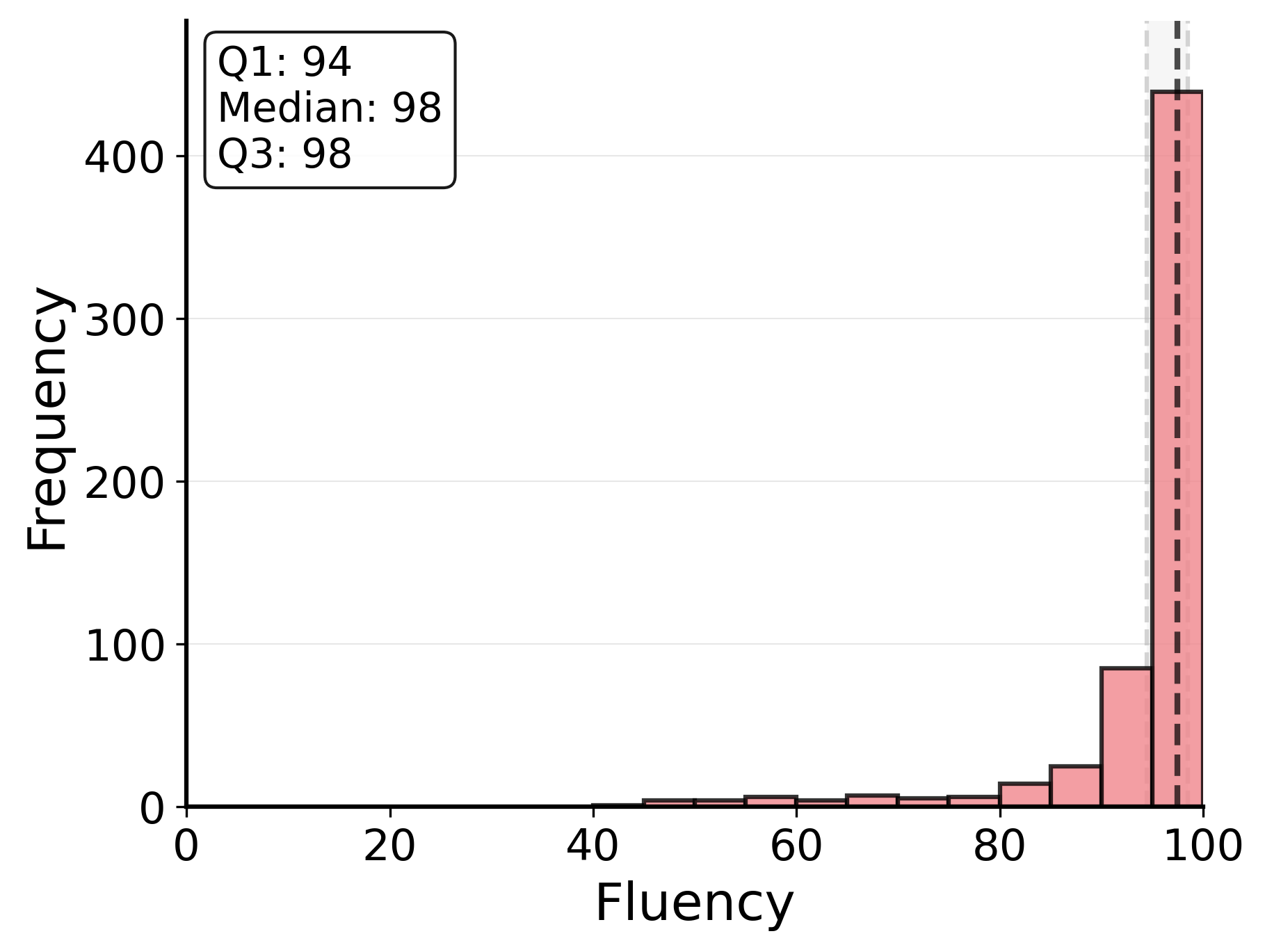}
        \caption{Fluency}
    \end{subfigure}
    
    \vspace{1em}
    \textbf{iii) Zephyr-7B}\\[0.5em]
            \begin{subfigure}[b]{0.325\textwidth}
        \includegraphics[width=\textwidth, height=4cm]{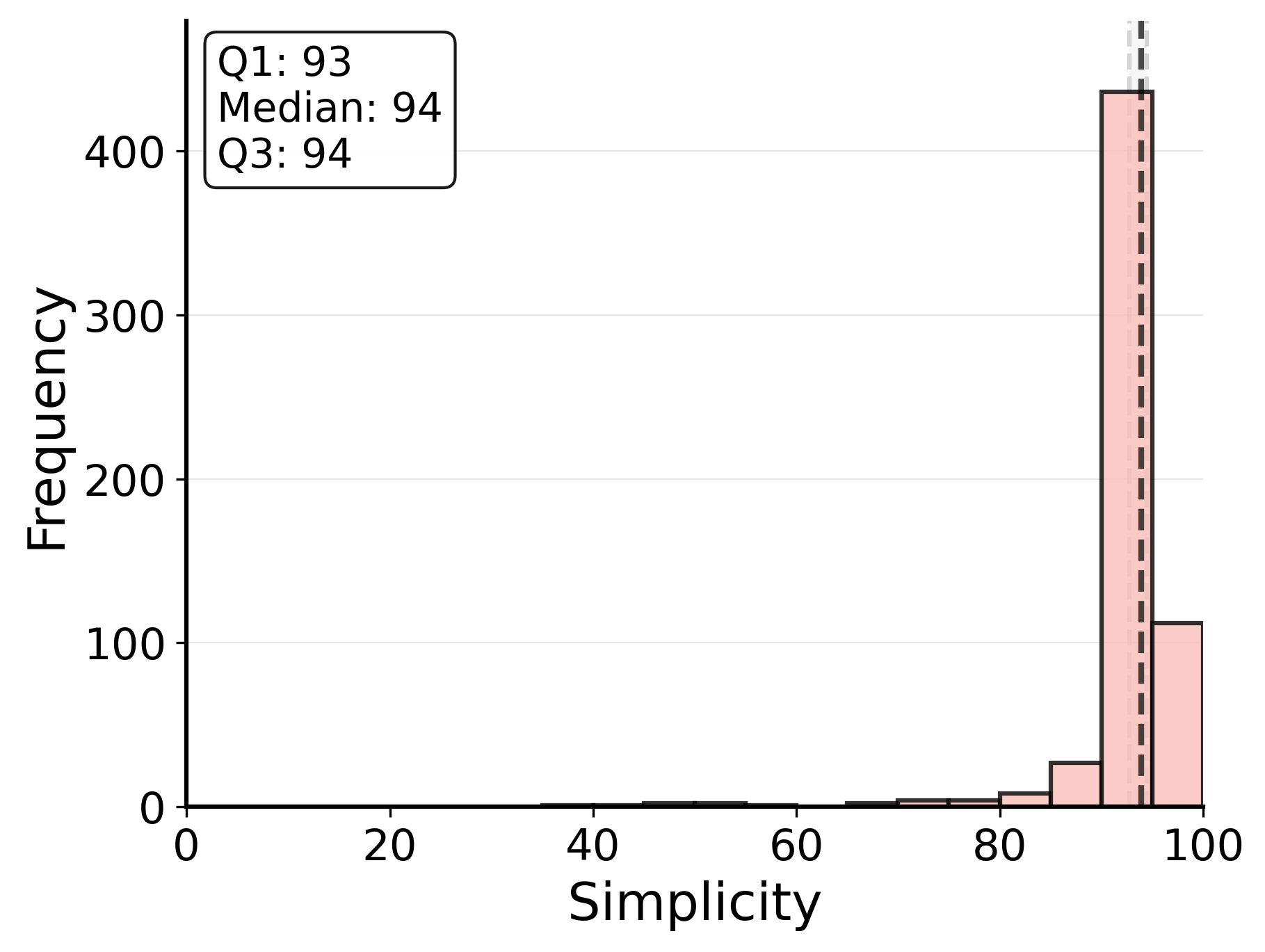}
        \caption{Simplicity}
    \end{subfigure}
    \begin{subfigure}[b]{0.325\textwidth}
        \includegraphics[width=\textwidth, height=4cm]{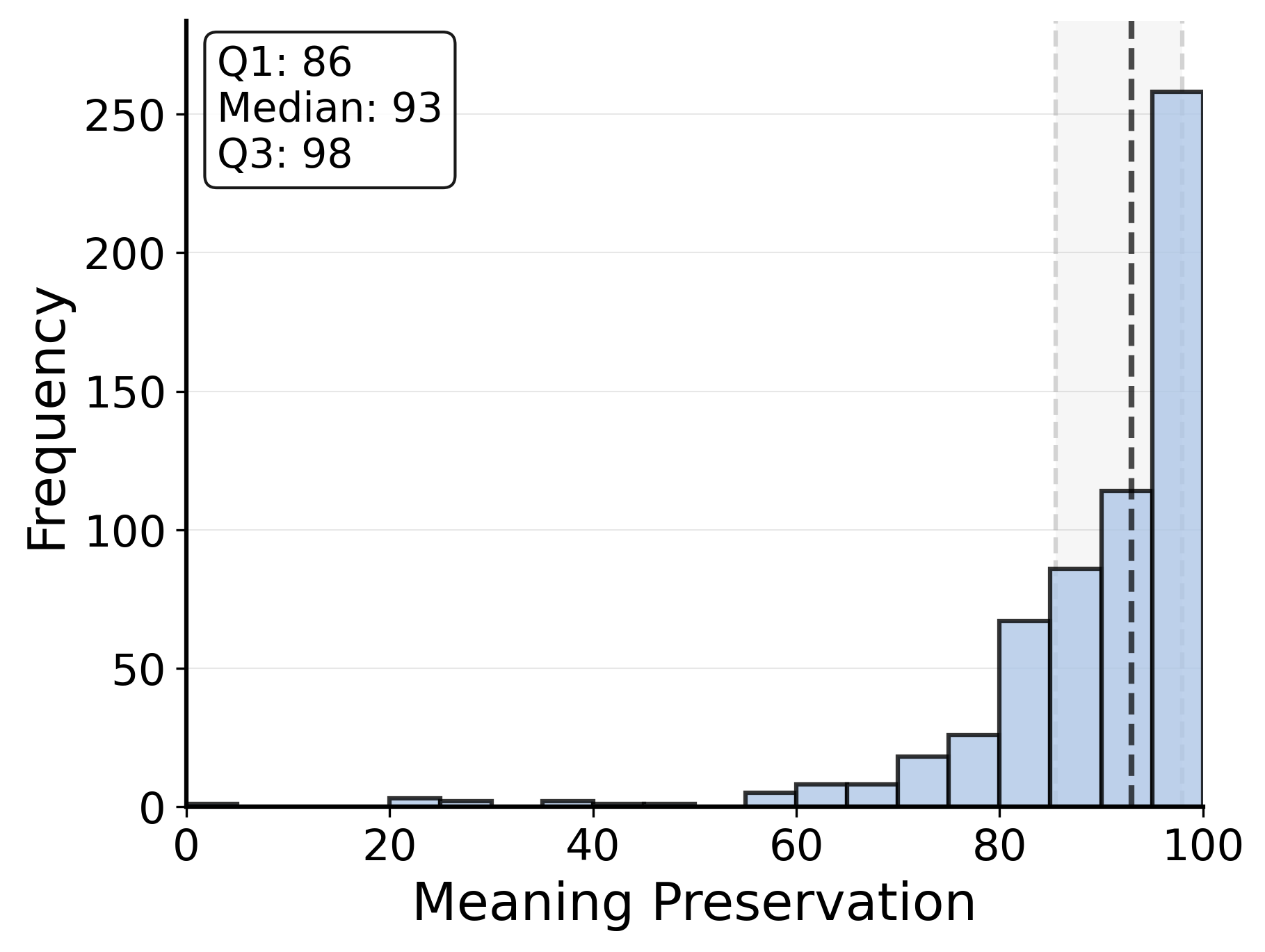}
        \caption{Meaning Preservation}
    \end{subfigure}
    \begin{subfigure}[b]{0.325\textwidth}
        \includegraphics[width=\textwidth, height=4cm]{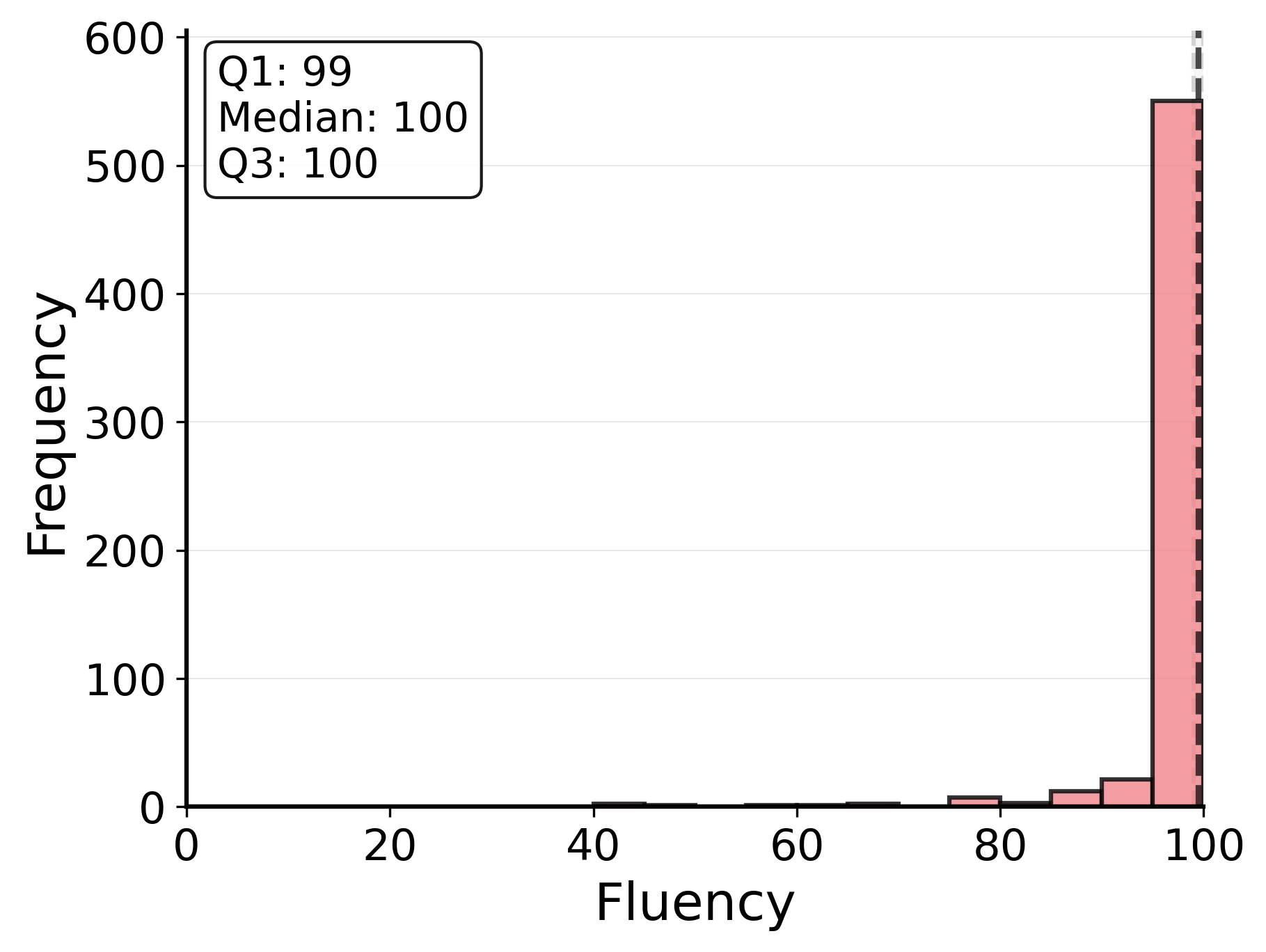}
        \caption{Fluency}
    \end{subfigure}
    \caption[Score distributions of Distill-Qwen, Distill-Llama and Zephyr by criterion] {Histograms of score distributions for i) Distill-Qwen-7B, ii) Distill-Llama-8B, and iii) Zephyr-7B, broken down by scoring criterion. The yellow band indicates the IQR, while vertical ticks show the minimum, median, and maximum values. These values are also listed in the box in the upper left corner of each plot.}
    \label{fig:model_hist_human}
\end{figure*}

Figure~\ref{fig:model_hist_human} compares the distributions of \textit{simplicity}, \textit{meaning preservation}, and \textit{fluency} for Distill-Qwen-7B, Distill-Llama-8B, and Zephyr-7B. Distill-Llama-8B shows a stronger right skew in \textit{simplicity} (median 90) and higher peaks in \textit{meaning preservation} (median 91, with values extending down to 2), though its standard deviations are not significantly different from Distill-Qwen-7B. For \textit{fluency}, the two models are visually similar, but Llama-8B has a slightly higher variance ($p < 0.1$). Zephyr-7B, by contrast, produces tighter distributions across all criteria, with higher medians (\textit{simplicity}: 94) and lower standard deviations, reflecting reduced diversity. Overall, Zephyr-7B is less variable than the other two models, while differences between Distill-Qwen-7B and Llama-8B are minor and mostly not significant (see also Table~\ref{tab:deviation_models}).

\item \textit{\textbf{Hypothesis 2: Different Models have Different Sampling Variability.}} 

Table~\ref{tab:sampling-variance} compared the variance of ten repeated outputs per simplification pair. Distill-Qwen-7B and Distill-Llama-8B show nearly identical sampling variability across all criteria (e.g., \textit{simplicity}: 6.51 vs. 6.49; \textit{meaning preservation}: 10.05 vs. 10.03), with no significant differences ($p > 0.5$). Zephyr-7B, however, exhibits significantly lower variance in \textit{simplicity} (4.01) and \textit{fluency} (3.06) than either model ($p < 0.0001$), indicating more consistent predictions. The only exception is \textit{meaning preservation}, where Zephyr-7B shows slightly higher variance (11.08 vs. 10.03), a marginal but significant difference ($p \approx 0.02$). Overall, Qwen-7B and Llama-8B display similar stability, while Zephyr-7B is notably steadier in \textit{simplicity} and \textit{fluency} but less so in \textit{meaning preservation}.

\begin{table}[!htb]
\centering
\caption[Per-row standard deviations for Distill-Qwen-7B, Distill-Llama-8B, and Zephyr-7B]{Comparison of the mean per-row output standard deviation for Distill-Qwen-7B, Distill-Llama-8B, and Zephyr-7B grouped by evaluation criterion and model pair. $p$-values from independent \textit{t}-tests indicate the statistical significance of differences in mean sampling variance between models.}
\label{tab:sampling-variance}
\resizebox{\columnwidth}{!}{%
\begin{tabular}{lcc}
\toprule
\textbf{Criterion} & \textbf{Mean Std. Deviation} & \textbf{$p$-value} \\
\hline
\rowcolor{gray!20!white}
\multicolumn{3}{c}{\textit{Distill-Qwen-7B vs Distill-Llama-8B}} \\
\textit{Simplicity} & 6.51 vs 6.49 & \textit{0.9610} \\
\textit{Meaning P.} & 10.05 vs 10.03 & \textit{0.9436} \\
\textit{Fluency} & 6.20 vs 6.41 & \textit{0.5573} \\
\hline
\rowcolor{gray!20!white}
\multicolumn{3}{c}{\textit{Distill-Qwen-7B vs Zephyr-7B}} \\
\textit{Simplicity} & 6.51 vs 4.01 & $<$0.0001 \\
\textit{Meaning P.} & 10.05 vs 11.08 & 0.0152 \\
\textit{Fluency} & 6.20 vs 3.06 & $<$0.0001 \\
\hline
\rowcolor{gray!20!white}
\multicolumn{3}{c}{\textit{Distill-Llama-8B vs Zephyr-7B}} \\
\textit{Simplicity} & 6.49 vs 4.01 & $<$0.0001 \\
\textit{Meaning P.} & 10.03 vs 11.08 & 0.0193 \\
\textit{Fluency} & 6.41 vs 3.06 & $<$0.0001 \\
\bottomrule
\end{tabular}
}
\end{table}

\item \textit{\textbf{Hypothesis 3: Models of Similar Size Exhibit Positively Correlated Scoring Behavior.}} 

Table~\ref{tab:correla_comp} shows significant positive correlations ($p < 0.0001$) across all model pairs. Distill-Qwen-7B and Distill-Llama-8B align most closely (Pearson $r$ from 0.85 to 0.89), while correlations with Zephyr-7B are more moderate (Pearson $r$ from 0.45 to 0.58). Disagreements mainly occur in cases where one model assigns scores below 25 and the other does not. Krippendorff's $\alpha$ on $z$-scores remains moderate to high (0.53 to 0.69), indicating consistent though weaker alignment with Zephyr-7B, comparable to correlations with larger Qwen models.

\begin{table}[H]
  \centering
  \caption{Pearson ($r$) and Spearman ($\rho$) correlations with $p$-values for (Distill-)Qwen(-7B), (Distill-)Llama(-8B) and Zephyr(-7B) scores by criterion. All $p$-values are 0.}
  \label{tab:correla_comp}
  \resizebox{\columnwidth}{!}{%
  \begin{tabular}{l@{\hskip 2pt}cccc}
    \toprule
    \multirow{2.5}{*}{\textbf{Model Comparison}} & & \multicolumn{3}{c}{\textbf{Criterion ($r$ / $\rho$)}} \\
    \cmidrule(lr){3-5}
    & & \textbf{\textit{Simplicity}} & \textbf{\textit{Meaning P.}} & \textbf{\textit{Fluency}} \\
    \midrule
    Qwen vs Llama     & & 0.85 / 0.51 & 0.89 / 0.77 & 0.76  / 0.45  \\
     Qwen  vs Zephyr    & & 0.49  / 0.36  & 0.58  / 0.66  & 0.42  / 0.37  \\
    Llama vs Zephyr     & & 0.45  / 0.33  & 0.58  / 0.67  & 0.41  / 0.34  \\
    \bottomrule
  \end{tabular}
  }
\end{table}

\end{itemize}

To summarize the experimental findings:
\begin{itemize}[left=0pt]
    \item \textbf{Correlation with larger judges}: smaller Qwen models show moderate to strong correlations with the 14B reference model, with the highest agreement consistently observed for meaning preservation and the lowest for fluency.
    \item \textbf{Criterion-dependent reliability}: Across all evaluated models, score reliability varies substantially by criterion; meaning preservation is the most stable dimension, while fluency exhibits compressed score ranges and lower discriminative power.
    \item \textbf{Model-specific behavior}: Distil-Qwen-7B achieves the strongest overall correlation with the 14B model, whereas Zephyr-7B produces less diverse but more conservative scores, indicating complementary strengths rather than a single optimal judge.
    \item \textbf{Implication for aggregation}: These findings motivate the use of multi-model aggregation for \textit{LLM-Judge} scores and confirm the necessity of independent human evaluation for validating downstream metrics such as \textsc{DEtect}.
\end{itemize}

\section{\textsc{DEtect} vs Other Metrics by Type}
\label{sec:app_detect_metrics}

\begin{figure*}[!hbt]
    \centering
    \textbf{i) \textit{Human-Judge} vs. \textit{LLM-Judge}}\\[0.5em]
    \begin{subfigure}[b]{0.4\textwidth}
        \includegraphics[width=\textwidth, height = 5cm]{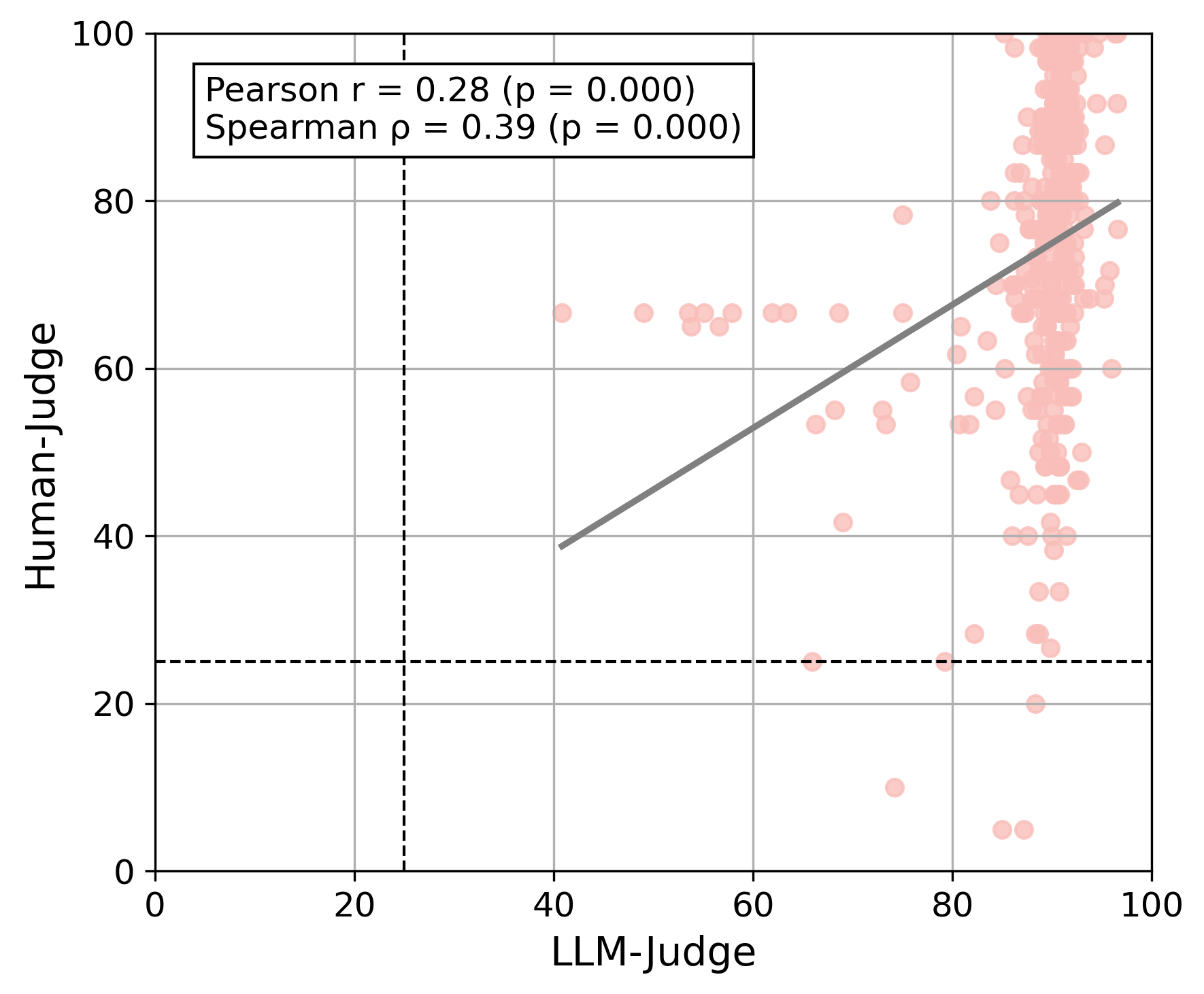}
        \caption{Simplicity Score}
    \end{subfigure}
    \begin{subfigure}[b]{0.4\textwidth}
        \includegraphics[width=\textwidth, height = 5cm]{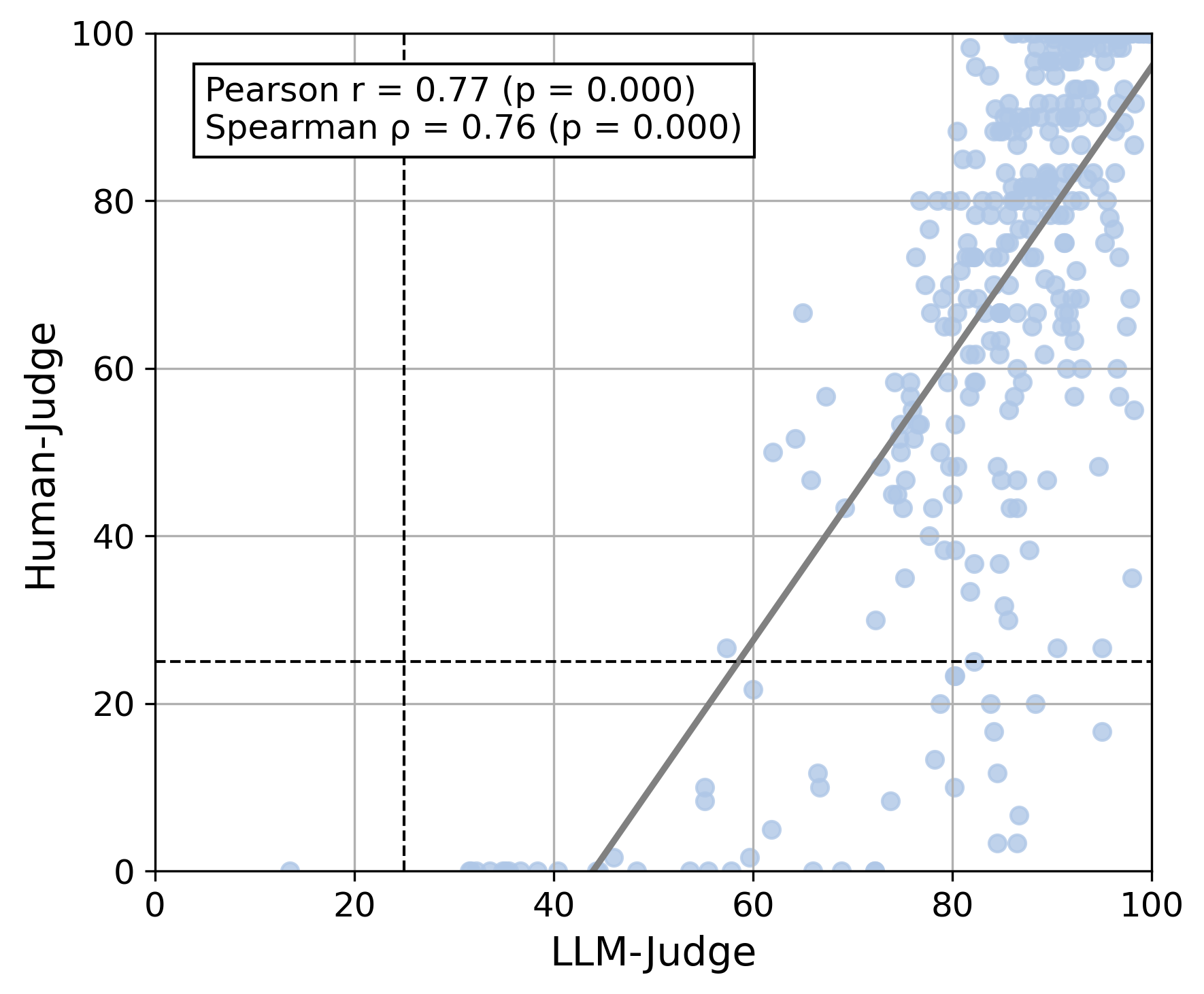}
        \caption{Meaning Preservation Score}
    \end{subfigure}
    \begin{subfigure}[b]{0.4\textwidth}
        \includegraphics[width=\textwidth, height = 5cm]{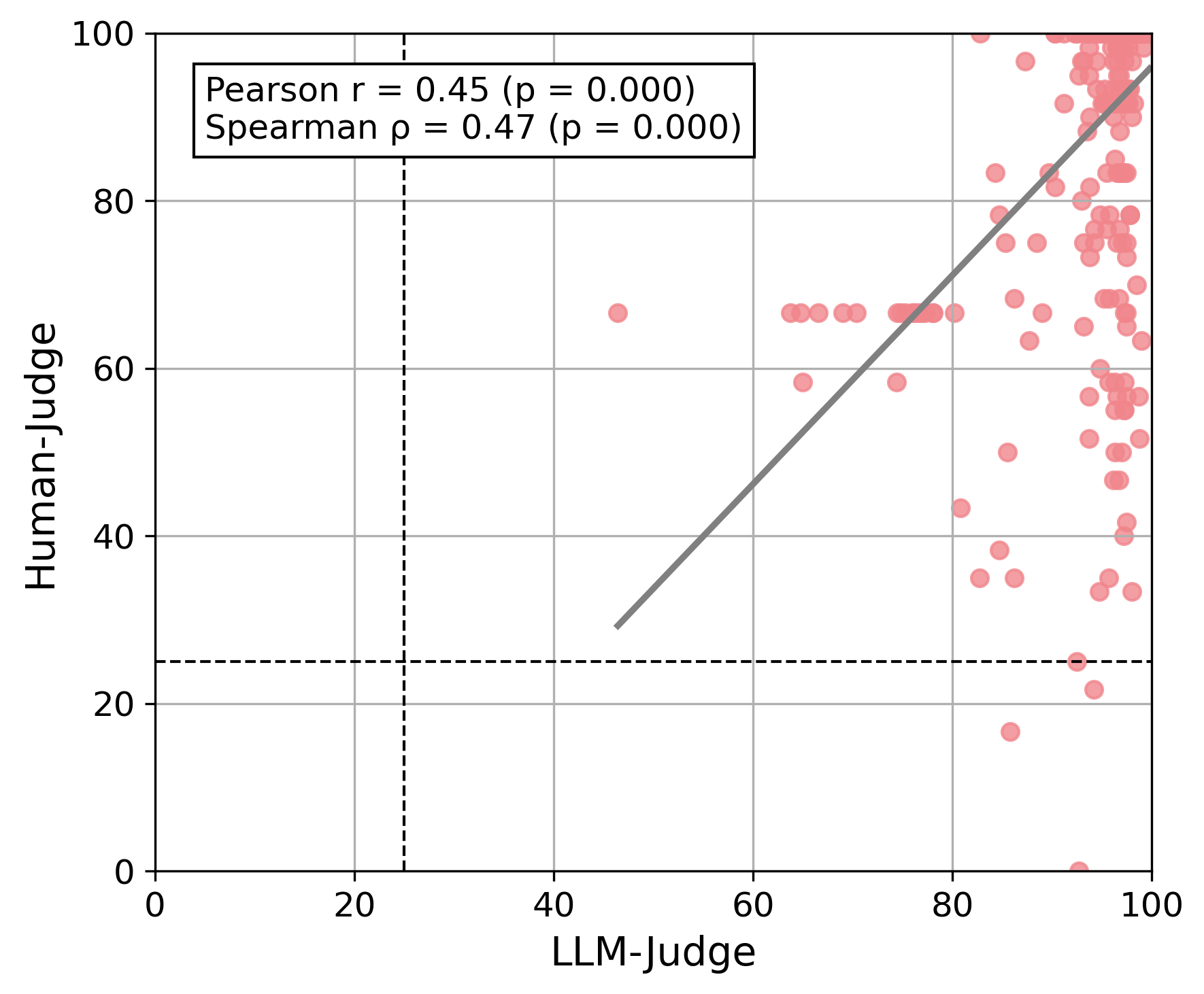}
        \caption{Fluency Score}
    \end{subfigure}
     \begin{subfigure}[b]{0.4\textwidth}
        \includegraphics[width=\textwidth, height = 5cm]{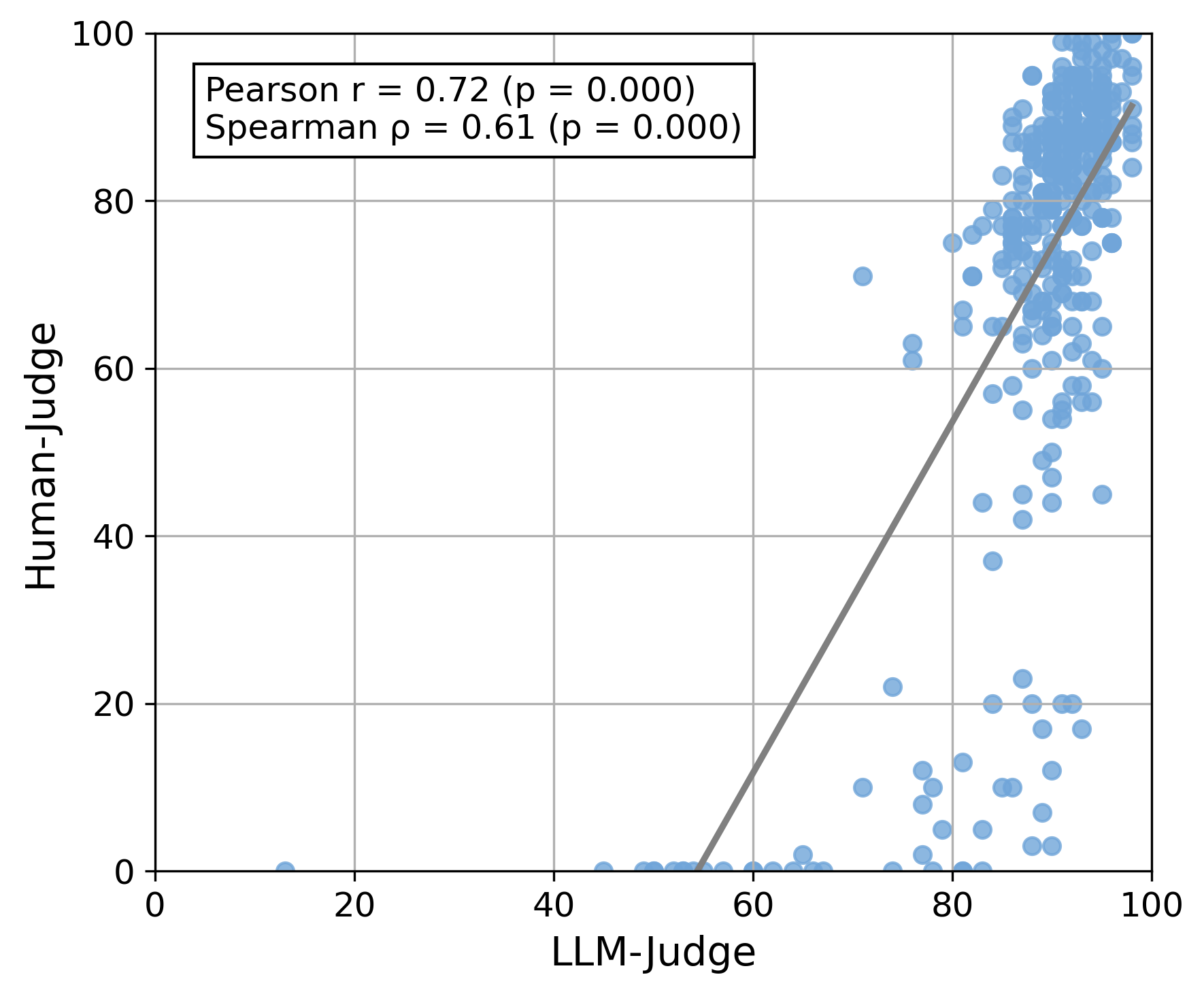}
        \caption{Total Score}
    \end{subfigure}
    \end{figure*}
    
    \begin{figure*}[!htb]
    \centering
    \textbf{ii) \textit{LLM-Judge} vs. \textsc{DEtect}}\\[0.5em]
    \begin{subfigure}[b]{0.4\textwidth}
        \includegraphics[width=\textwidth, height = 5cm]{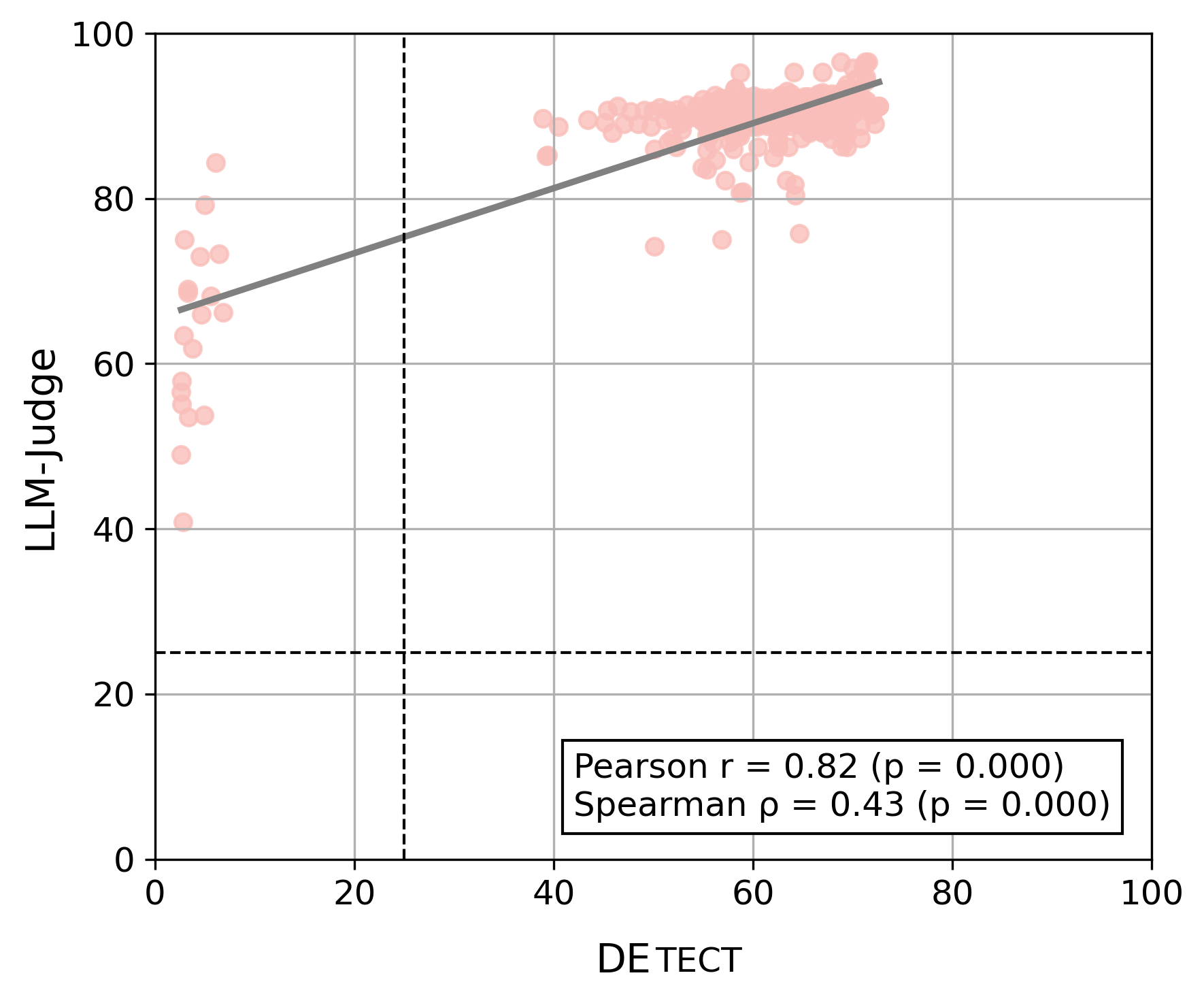}
        \caption{Simplicity Score}
    \end{subfigure}
    \begin{subfigure}[b]{0.4\textwidth}
        \includegraphics[width=\textwidth, height = 5cm]{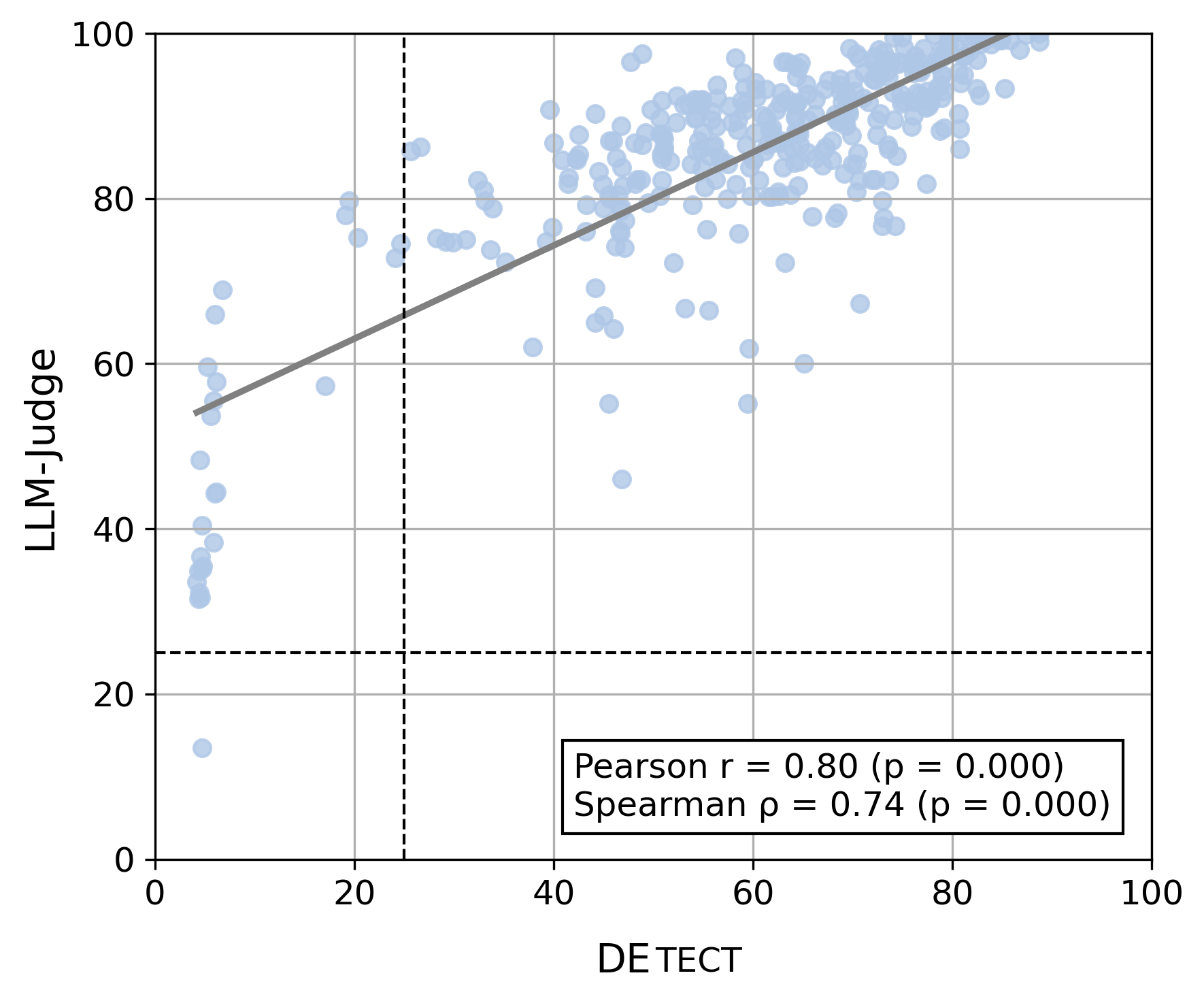}
        \caption{Meaning Preservation Score}
    \end{subfigure}
    \begin{subfigure}[b]{0.4\textwidth}
        \includegraphics[width=\textwidth, height = 5cm]{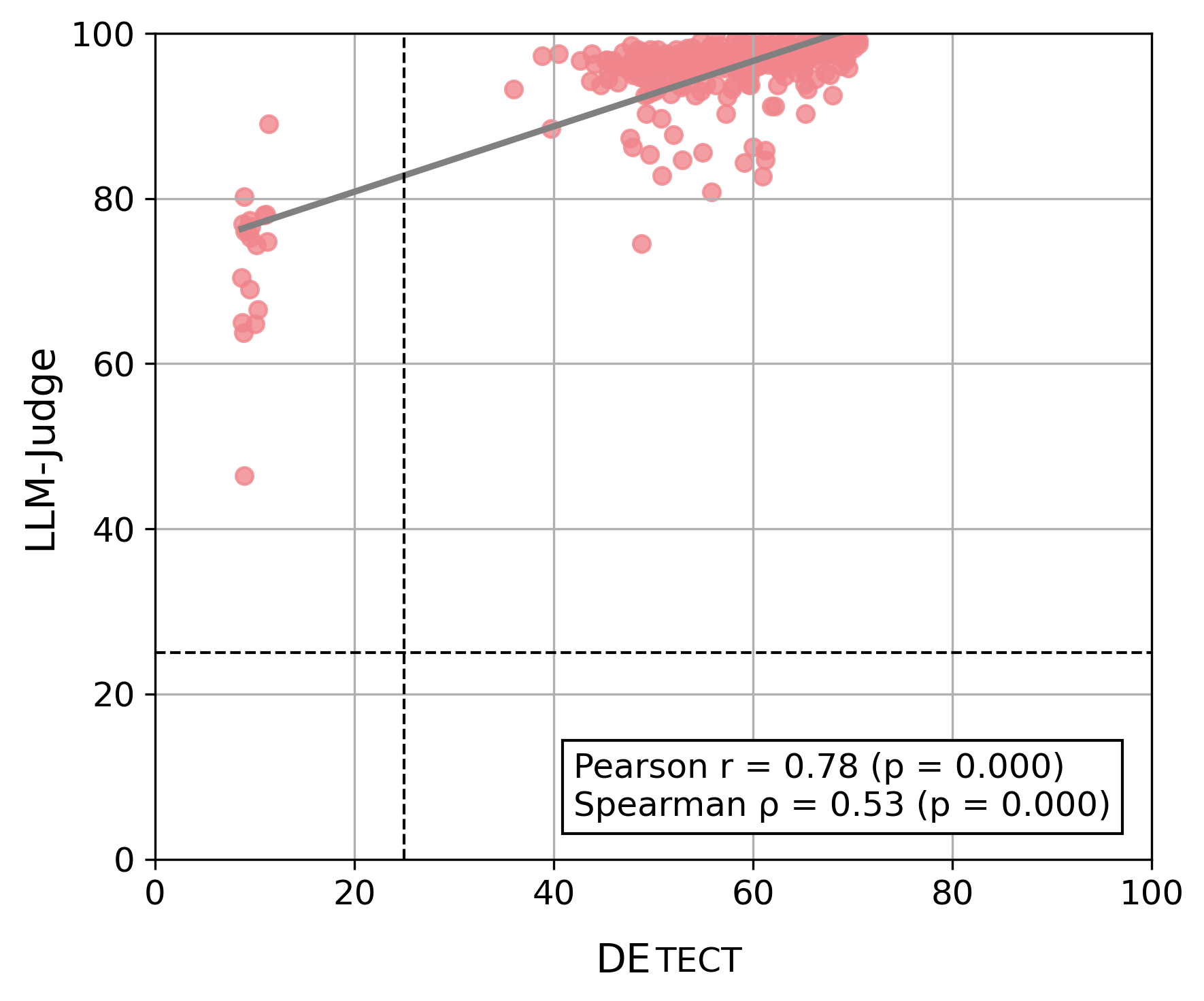}
        \caption{Fluency Score}
    \end{subfigure}
     \begin{subfigure}[b]{0.4\textwidth}
        \includegraphics[width=\textwidth, height = 5cm]{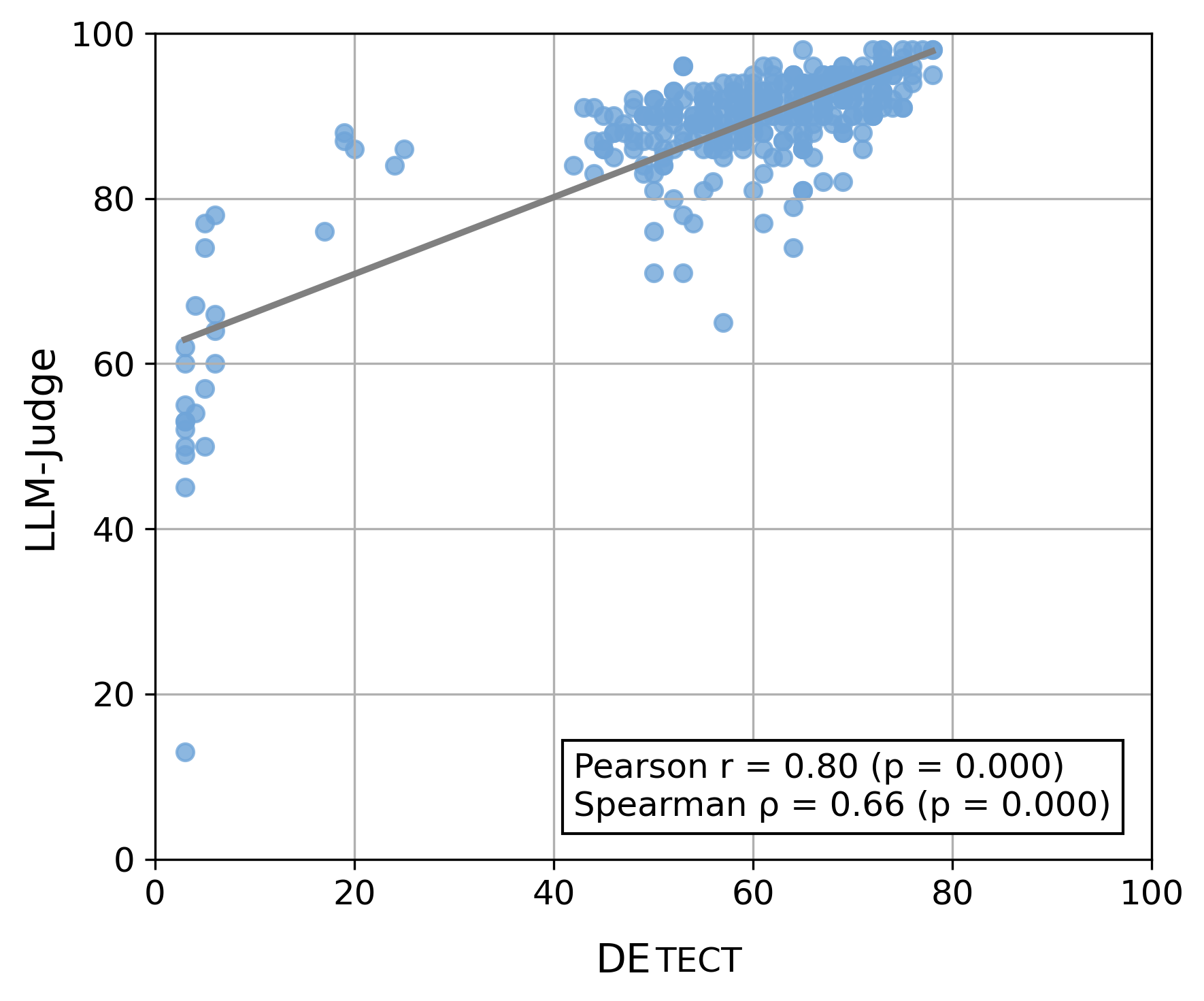}
        \caption{Total Score}
    \end{subfigure}
    \end{figure*}
    
    \begin{figure*}[!htb]
    \centering
    \textbf{iii) \textit{Human-Judge} vs.\textsc{DEtect}}\\[0.5em]
    \begin{subfigure}[b]{0.4\textwidth}
        \includegraphics[width=\textwidth, height = 5cm]{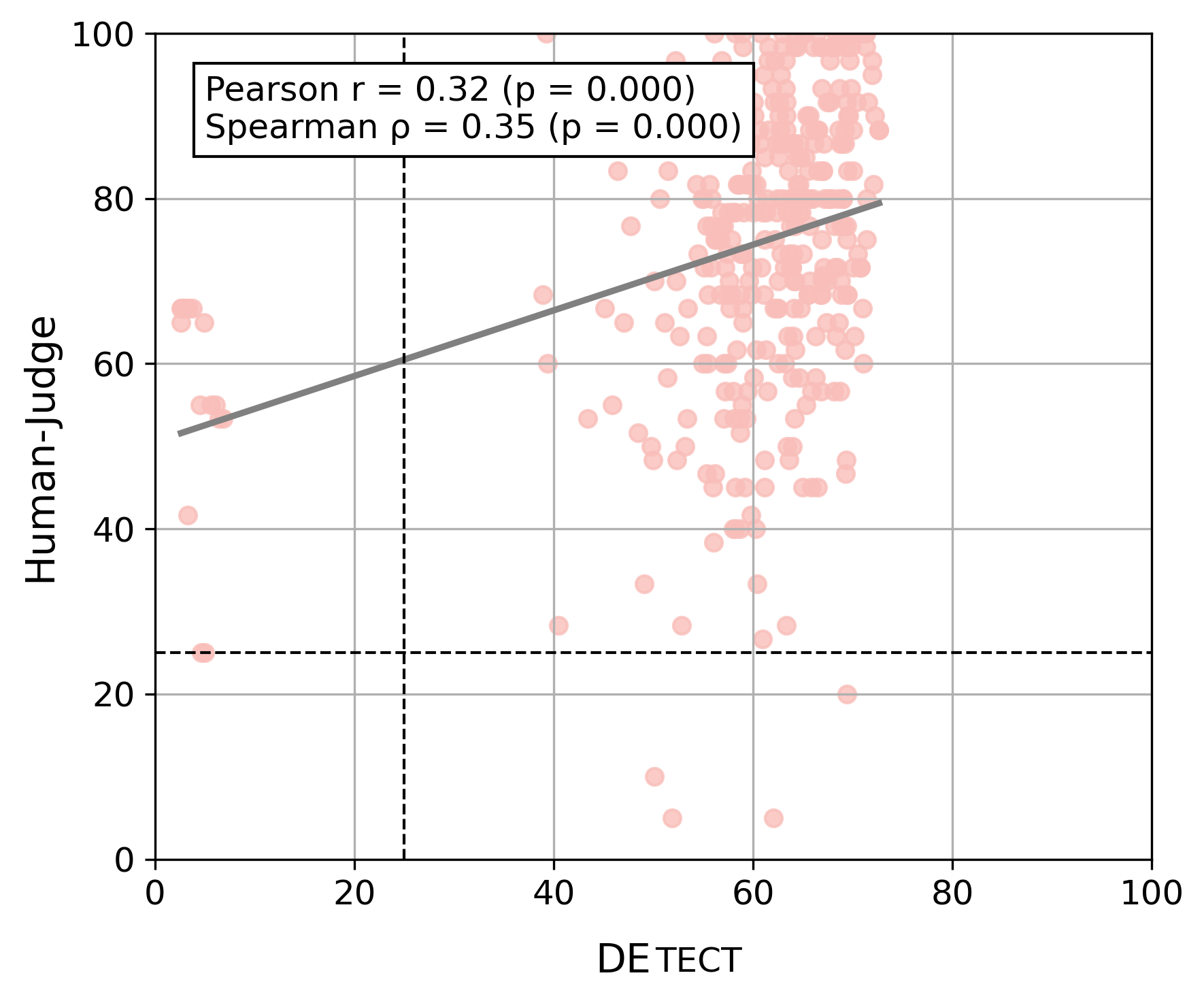}
        \caption{Simplicity Score}
    \end{subfigure}
    \begin{subfigure}[b]{0.4\textwidth}
        \includegraphics[width=\textwidth, height = 5cm]{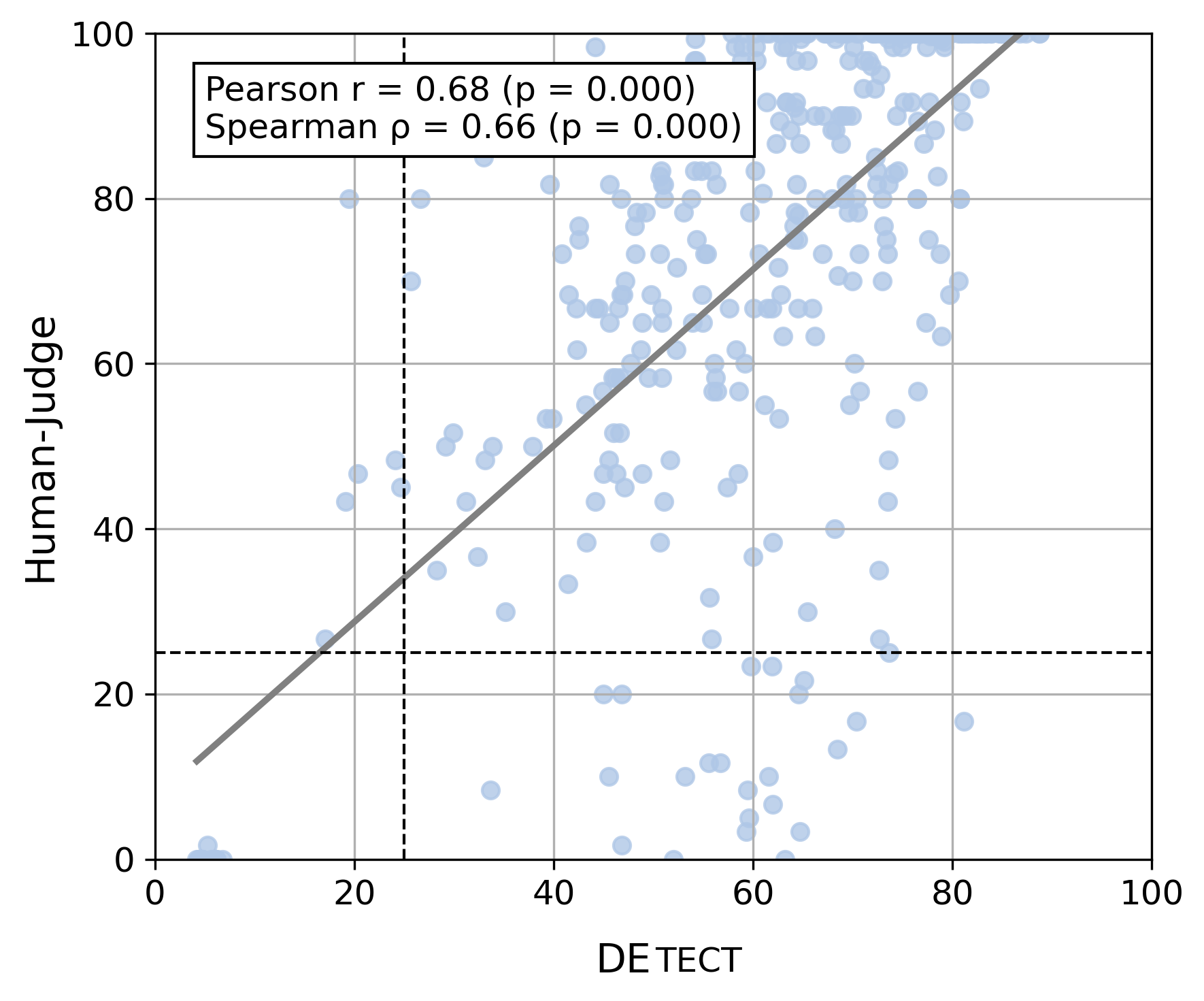}
        \caption{Meaning Preservation Score}
    \end{subfigure}
    \begin{subfigure}[b]{0.4\textwidth}
        \includegraphics[width=\textwidth, height = 5cm]{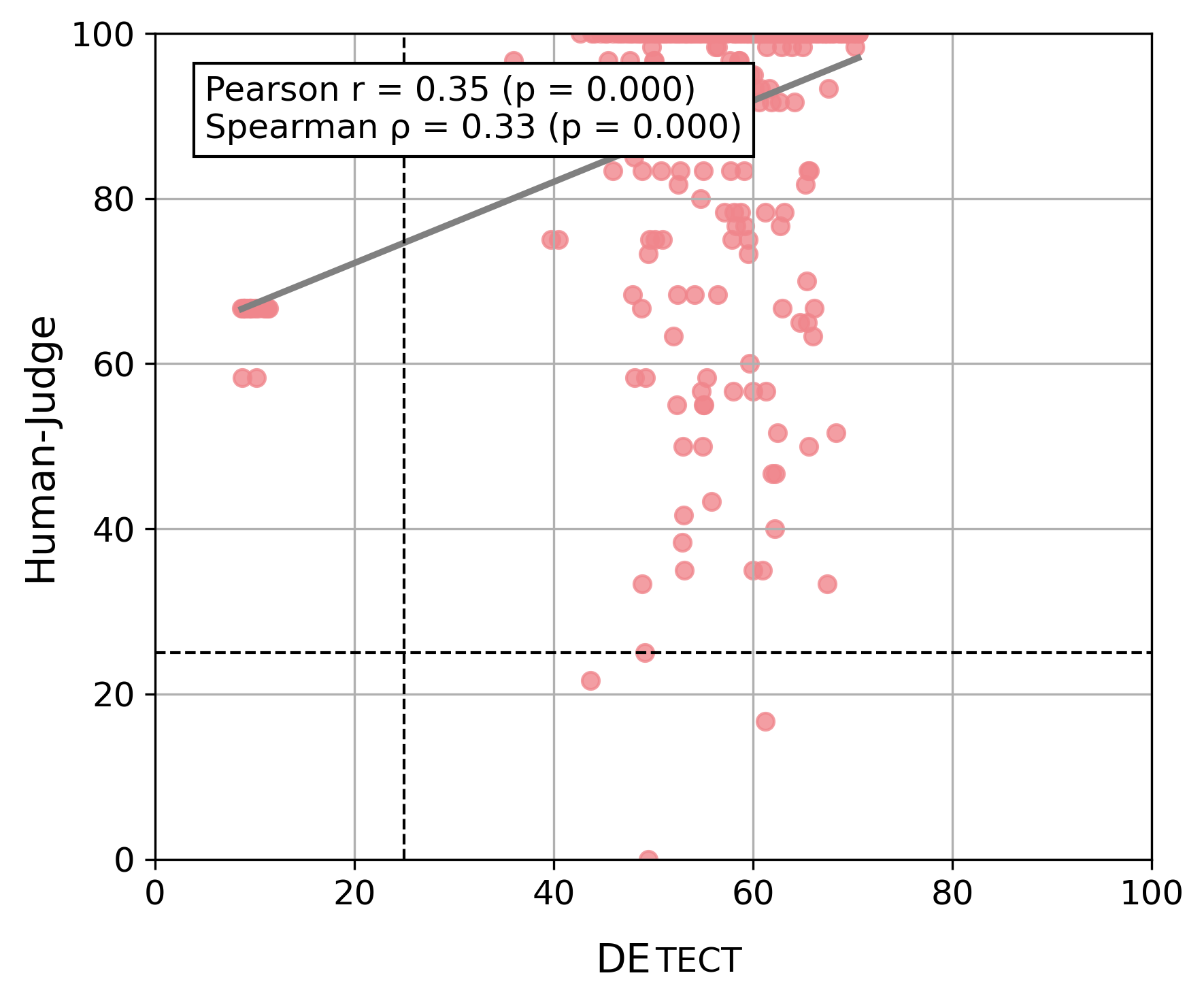}
        \caption{Fluency Score}
    \end{subfigure}
     \begin{subfigure}[b]{0.4\textwidth}
        \includegraphics[width=\textwidth, height = 5cm]{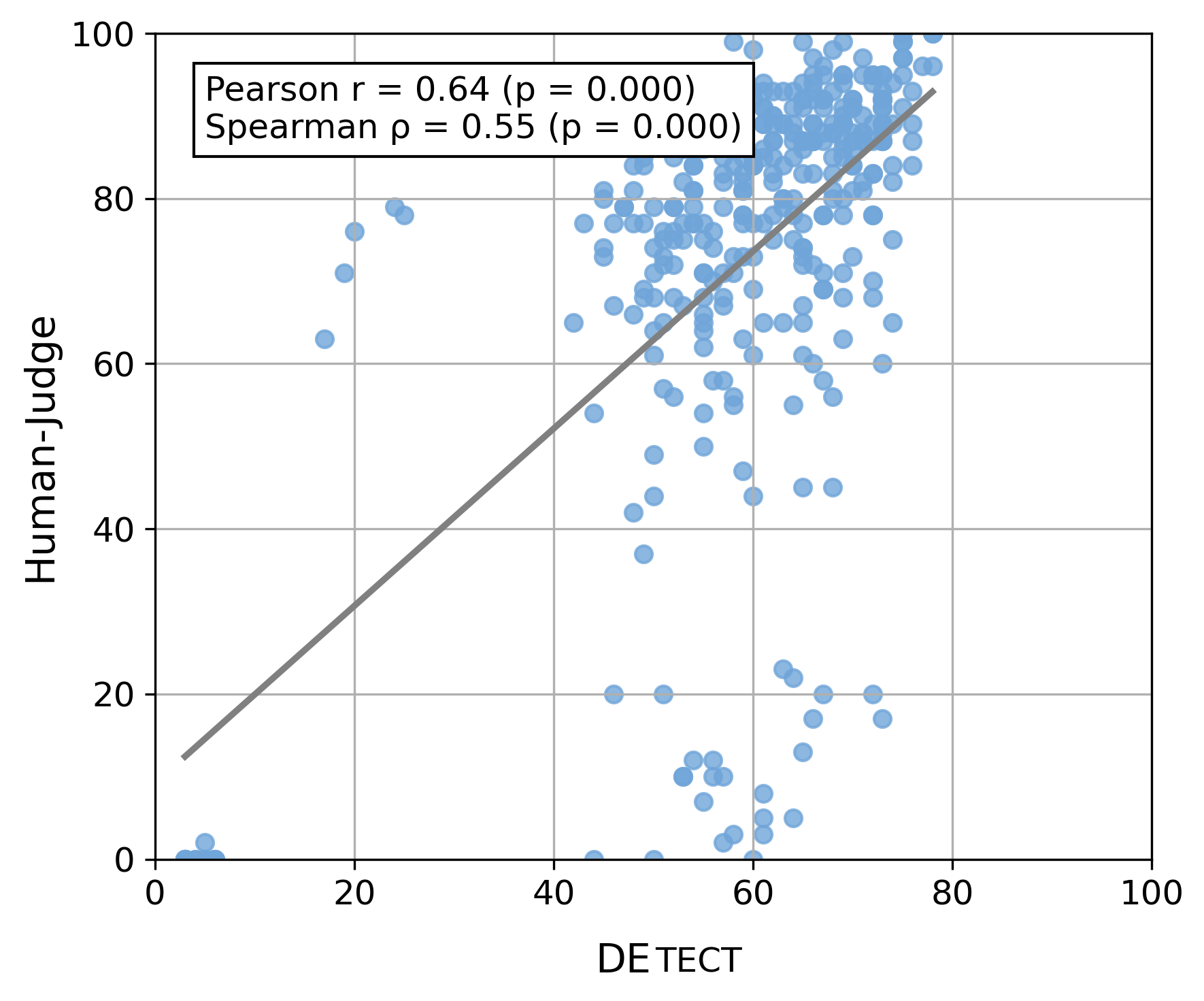}
        \caption{Total Score}
    \end{subfigure}
    \caption{Correlations between \textit{Human-Judge}, \textit{LLM-Judge} and \textsc{DEtect} for each criterion and total scores. Dashed lines indicate ``rejection zones'' where one model scores below 25.}
    \label{fig:scatter-human-LLM}
\end{figure*}

Table~\ref{tab:detect-metrics-coorelations} and Figure~\ref{fig:scatter-human-LLM} show the per score type correlations between \textsc{DEtect} and other metrics. Each category was re-sampled to obtain 100 samples of size 45. Figure~\ref{fig:grading-tool} shows the annotation tool we customized for this work.

\begin{table}[!htbp]
\centering
\caption{Pearson correlations of \textsc{DEtect} and other metrics with \textit{Human-Judge} scores.}
\label{tab:detect-metrics-coorelations}

\begin{subtable}[t]{0.45\textwidth}
\centering
\caption{Simplicity score correlations.}
\label{tab:simplicity}
\renewcommand{\arraystretch}{0.9}
\resizebox{0.95\columnwidth}{!}{
\begin{tabular}{lccc}
\toprule
\multirow{2}{*}{\textbf{Metric}} & \multicolumn{3}{c}{\textbf{\textit{Simplicity}}} \\
\cmidrule{2-4}
 & \textit{Paraphrase} & \textit{Delete} & \textit{Split} \\
\midrule
BLEU & 0.225  & 0.194 & 0.415 \\
SARI & 0.072 & 0.223 & 0.357 \\
BERTScore & 0.370 & 0.395 & 0.554 \\
\textit{LLM-Judge} & 0.265 & 0.322  & 0.395 \\
\textsc{DEtect} & 0.311  & 0.307  & 0.490 \\
\bottomrule
\end{tabular}
}
\end{subtable}

\vspace{0.5em}

\begin{subtable}[t]{0.45\textwidth}
\centering
\caption{Meaning preservation score correlations.}
\label{tab:meaning}
\renewcommand{\arraystretch}{0.9}
\resizebox{0.95\columnwidth}{!}{
\begin{tabular}{lccc}
\toprule
\multirow{2}{*}{\textbf{Metric}} & \multicolumn{3}{c}{\textbf{\textit{Meaning Preservation}}} \\
\cmidrule{2-4}
 & \textit{Paraphrase} & \textit{Delete} & \textit{Split} \\
\midrule
BLEU & 0.240  & 0.337  & 0.471 \\
SARI & -0.017  & 0.256  & 0.276 \\
BERTScore & 0.310  & 0.469 & 0.680 \\
\textit{LLM-Judge} & 0.636  & 0.791  & 0.795 \\
\textsc{DEtect} & 0.464  & 0.654  & 0.744 \\
\bottomrule
\end{tabular}
}
\end{subtable}

\vspace{0.5em}

\begin{subtable}[t]{0.45\textwidth}
\centering
\caption{Fluency score correlations.}
\label{tab:fluency}
\renewcommand{\arraystretch}{0.9}
\resizebox{0.95\columnwidth}{!}{
\begin{tabular}{lccc}
\toprule
\multirow{2}{*}{\textbf{Metric}} & \multicolumn{3}{c}{\textbf{\textit{Fluency}}} \\
\cmidrule{2-4}
 & \textit{Paraphrase} & \textit{Delete} & \textit{Split} \\
\midrule
BLEU & 0.160  & 0.157 & 0.217 \\
SARI & 0.044  & 0.164  & 0.316 \\
BERTScore & 0.196  & 0.307  & 0.294 \\
\textit{LLM-Judge} & 0.301  & 0.511  & 0.471 \\
\textsc{DEtect} & 0.215  & 0.289  & 0.429 \\
\bottomrule
\end{tabular}
}
\end{subtable}
\end{table}

\begin{figure*}[!htb]
    \centering
    \vspace{4em}
    \begin{subfigure}[t]{0.32\textwidth}
        \centering
        \includegraphics[width=\linewidth]{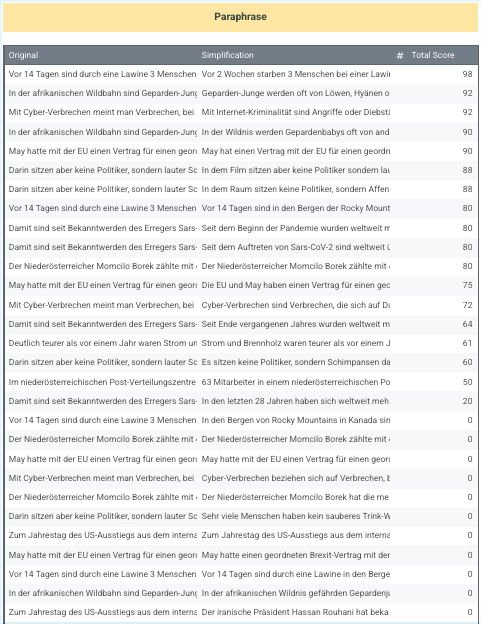}
        \caption{\textit{Paraphrase} samples annotation.}
    \end{subfigure}
    \hfill
    \begin{subfigure}[t]{0.32\textwidth}
        \centering
        \includegraphics[width=\linewidth]{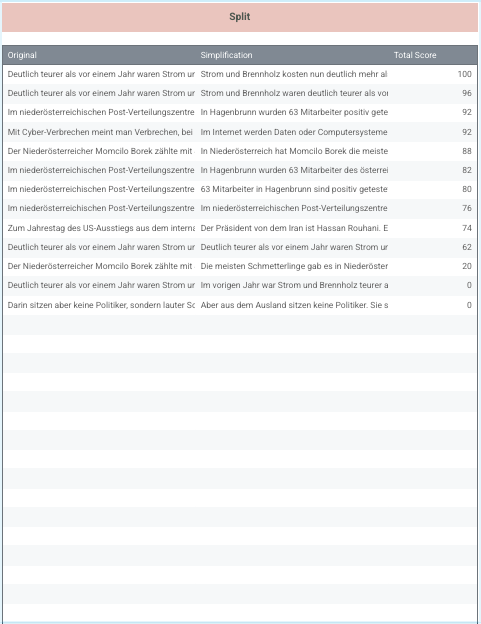}
        \caption{\textit{Split} samples annotation.}
    \end{subfigure}
    \hfill
    \begin{subfigure}[t]{0.32\textwidth}
        \centering
        \includegraphics[width=\linewidth]{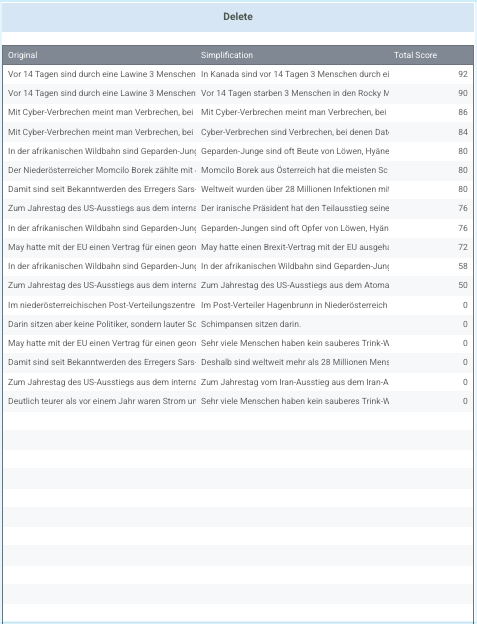}
        \caption{\textit{Delete} samples annotation.}
    \end{subfigure}
\end{figure*}

\begin{figure*}[!htb] 
    \ContinuedFloat
    \centering
    \setcounter{subfigure}{3} 
    \begin{subfigure}[t]{\textwidth}
        \centering
        \includegraphics[width=1\paperwidth, angle=90, keepaspectratio]{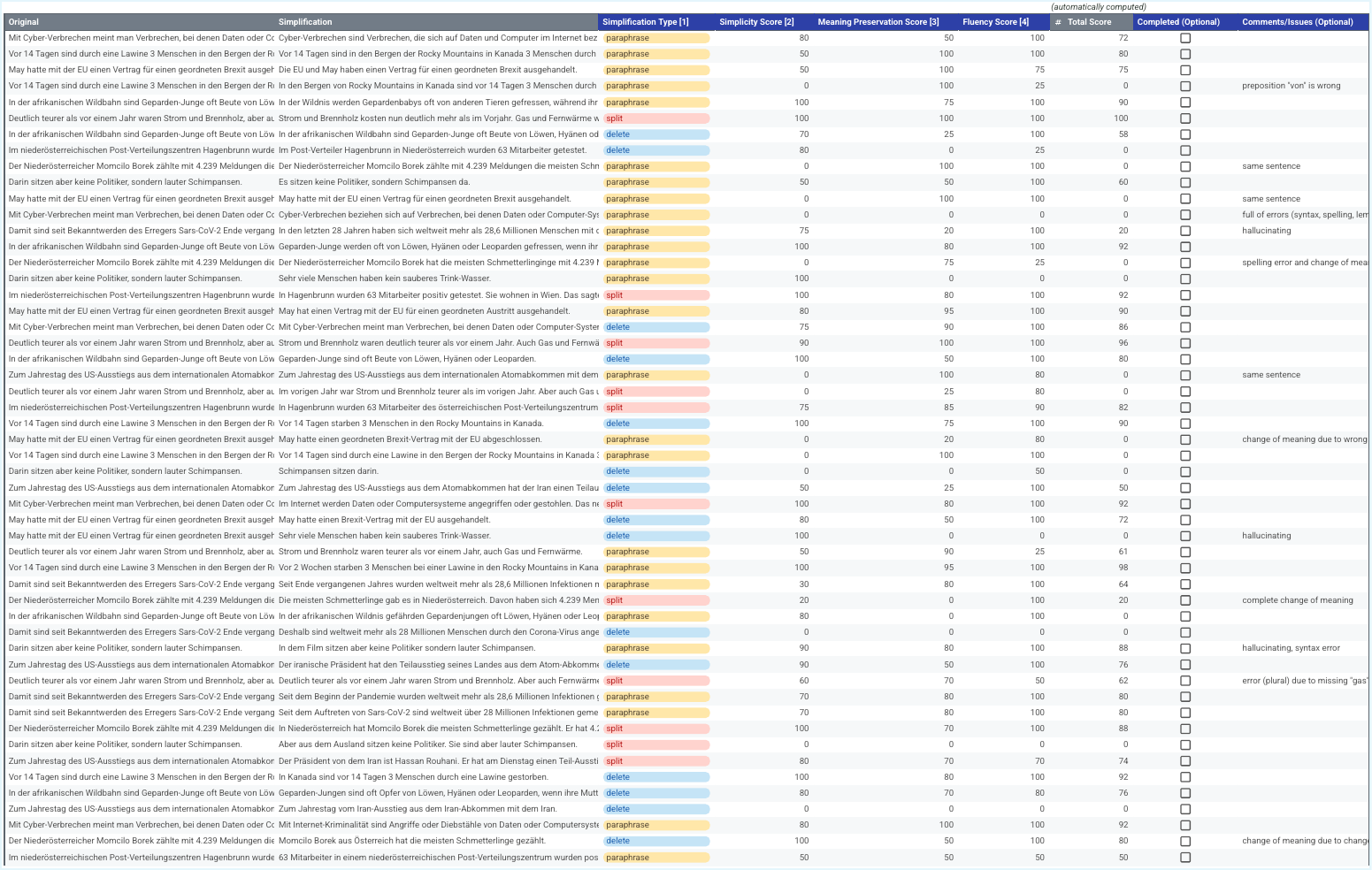}
        \caption{Overall grading tool interface.}
    \end{subfigure}
  \caption{Ranking of \textit{Human-Judge} quality scores. Following the \textsc{Rank \& Rate} protocol by \citet{maddela2023lens}, we used Google Sheet to collect \textit{Human-Judge} scores. Before kicking-off the grading, we hosted instruction sessions with the human experts to show them how to work with the tool.}
  \label{fig:grading-tool}
\end{figure*}


\end{document}